\documentclass[11pt]{article}

\usepackage{microtype} 
\usepackage{booktabs}  
\usepackage{url}  

\usepackage{amsmath}
\usepackage{amsthm}

\usepackage[final]{automl}

\usepackage[utf8]{inputenc} 
\usepackage[T1]{fontenc}    
\usepackage{url}            
\usepackage{booktabs}       
\usepackage{amsfonts}       
\usepackage{nicefrac}       
\usepackage{microtype}      

\usepackage{siunitx}
\usepackage{tabularx}
\usepackage{graphicx}
\usepackage[nolist]{acronym}
\usepackage{hyperref}
\usepackage{appendix}
\usepackage[position=b]{subcaption}
\usepackage{grffile}
\usepackage{authblk}
\usepackage{url}
\usepackage{todonotes}
\setuptodonotes{inline}
\usepackage{amsmath}

\DeclareMathOperator*{\argmin}{arg\,min}
\usepackage[capitalise,noabbrev]{cleveref}

\usepackage{placeins}
\usepackage{xspace}

\usepackage{floatrow}
\newfloatcommand{capbtabbox}{table}[][\FBwidth]

\usepackage{algorithm}
\usepackage{algpseudocode}

\usepackage{wrapfig}

\newcommand{\weiw}{$\alpha$\xspace}
\newcommand{\WEI}{\text{WEI}}
\definecolor{violet}{HTML}{D9347E}

\usepackage{xcolor}

\definecolor{my_magenta}{rgb}{0.796875,0.46875,0.734375}
\definecolor{my_green}{rgb}{0.0078125,0.6171875,0.44921875}
\definecolor{my_gold}{rgb}{0.8671875,0.55859375,0.01953125}
\definecolor{my_gray}{rgb}{0.5,0.5,0.5}
\definecolor{my_blue}{rgb}{0.4,0.4,1.}
\definecolor{my_green2}{HTML}{63a76b}
\definecolor{blue}{HTML}{526fae}

\newcommand{\revt}[1]{\textcolor{black}{#1}}
\newenvironment{rev}{\color{black}}{\ignorespacesafterend}
\newcommand{\revtwot}[1]{\textcolor{black}{#1}}

\usepackage{natbib}
\title{Self-Adjusting Weighted Expected Improvement\\for Bayesian Optimization}

\author[1]{\nameemail{Carolin Benjamins}{c.benjamins@ai.uni-hannover.de}}
\author[2,3]{Elena Raponi}
\author[3]{Anja Jankovic}
\author[3]{Carola Doerr}
\author[1]{Marius Lindauer}

\affil[1]{Institute of AI, Leibniz University Hannover, Germany}
\affil[2]{TUM School of Engineering and Design, TU München, Germany}
\affil[3]{Sorbonne Universit\'e, CNRS, LIP6, Paris, France}

\hypersetup{%
  pdfauthor={Carolin Benjamins}, 
  pdftitle={Self-Adjusting Weighted Expected Improvement for Bayesian Optimization},
  pdfsubject={Bayesian Optimization},
  pdfkeywords={}
}

\begin{document}

\begin{acronym}
\acro{DoE}{\emph{design of experiment}}
\acro{BO}{Bayesian optimization}
\acro{AF}{\emph{acquisition function}}
\acro{EI}{Expected Improvement}
\acro{WEI}{Weighted Expected Improvement}
\acro{SAWEI}{Self-Adjusting Weighted Expected Improvement}
\acro{PI}{Probability of Improvement}
\acro{UCB}{Upper Confidence Bound}
\acro{LCB}{Lower Confidence Bound}
\acro{ELA}{exploratory landscape analysis}
\acro{GP}{Gaussian Process}
\acro{TTEI}{Top-Two Expected Improvement}
\acro{TS}{Thompson Sampling}
\acro{DAC}{Dynamic Algorithm Configuration}
\acro{AC}{Algorithm Configuration}
\acro{CMA-ES}{CMA-ES}
\acro{AS}{algorithm selection}
\acro{PIAS}{per-instance algorithm selection}
\acro{PIAC}{per-instance algorithm configuration}
\acro{AFS}{Acquisition Function Selector}
\acro{VBS}{\emph{virtual best solver}}
\acro{RF}{random forest}
\acro{UBR}{Upper Bound Regret}
\acro{IQM}{interquartile mean}
\end{acronym}

\maketitle

\begin{abstract}
Bayesian Optimization (BO) is a class of surrogate-based, sample-efficient algorithms for optimizing black-box problems with small evaluation budgets.
The BO pipeline itself is highly configurable with many different design choices regarding the initial design, surrogate model, and acquisition function (AF).
Unfortunately, our understanding of how to select suitable components for a problem at hand is very limited.
In this work, we focus on the definition of the AF, whose main purpose is to balance the trade-off between exploring regions with high uncertainty and those with high promise for good solutions.
We propose Self-Adjusting Weighted Expected Improvement (SAWEI), where we let the exploration-exploitation trade-off self-adjust in a data-driven manner, based on a convergence criterion for BO.
On the noise-free black-box BBOB functions of the COCO benchmarking platform, our method exhibits a favorable anytime performance compared to handcrafted baselines and serves as a robust default choice for any problem structure.
The suitability of our method also transfers to HPOBench.
With SAWEI, we are a step closer to on-the-fly, data-driven, and robust BO designs that automatically adjust their sampling behavior to the problem at hand. 
\end{abstract}

\section{Introduction}

Black-box problems are challenging to optimize because we do not have direct access to the underlying structure of the problem landscape.
To optimize them, we can sequentially evaluate different points $x$ and use the obtained objective values $f(x)$ to choose which point(s) to evaluate next, but we do not have \emph{a priori} information where to find the most promising regions or how to best trade off exploration of the search space with exploitation of regions that appear to be very promising. 
Formally, in black-box optimization we want to find the minimum $x^*$ of a given function $f$, $x^* \in \argmin_{x\in \mathcal{X}} f(x)$, 
without having access to the function itself other than through the queries. Typical black-box problems occur in engineering or hyperparameter optimization (HPO), where the quality of potential solutions is evaluated via numeric simulations or training machine learning models.

Balancing exploration with exploitation is particularly challenging when we have a low number of available function evaluations in relation to the size of the search space $\mathcal{X}$. A popular approach to address such settings is \ac{BO}~\citep{mockus-bo89a,garnett_bayesoptbook_2023}, often promoted as sample-efficient for expensive black-box optimization. 
The main idea of \ac{BO} is to use a probabilistic surrogate model (e.g., a Gaussian Process), iteratively refining an approximation of the problem landscape that guides the optimization process.
\ac{BO} starts with an \emph{initial design} or \ac{DoE}, obtained from sampling strategies, e.g., random sampling, low-discrepancy sequences such as Sobol', or Latin Hypercube design~\citep{brochu-arxiv10a}.
With these initial points, the surrogate model is built to approximate the unknown objective function and capture the uncertainty of the true \revt{function value} on unobserved points.
The \ac{AF} (a.k.a. \emph{infill criterion}) is a utility function to trade off exploration of underexplored areas and exploitation of presumably promising ones.
The point with the highest acquisition function value is queried. Afterwards, the surrogate model is adjusted with the new observation, and the optimum is updated if the new point improves the target value of the best-so-far observation.
These steps are repeated for a given overall optimization budget.

Besides accurate probabilistic surrogate models and \revt{the type and size of initial design}~\citep{lindauer-dso19a,bossek-gecco20a,cowenrivers-arxiv21a}, the exploration-exploitation trade-off is crucial for successful and efficient optimization. 
Since the landscape of the black-box optimization problem is unknown, it is a priori unclear which \ac{AF} should be chosen for the optimization problem at hand.
Even worse, since each problem has its unique landscape, we need different exploration-exploitation trade-offs~\citep{benjamins-meta22a, benjamins-corr22a}.

Because there are different choices of \ac{AF}s, e.g., \acf{PI}~\citep{kushner-jfe64a}, \acf{EI}~\citep{mockus-tgo78a}, \ac{UCB}~\citep{forrester_engineering_2008}, \ac{TS}~\citep{thompson-biomet33a}, Entropy Search~\citep{hennig-jmlr12a} and Knowledge Gradient~\citep{frazier-jc09a}, selecting a suitable one for the problem at hand with insights on the landscape remains challenging. 
Furthermore, in the past, the choice of an \ac{AF} has been considered \emph{static} over the \ac{BO} process. 
Prior works suggest that mixed \ac{AF}-strategies~\citep{hoffman-uai11a,kandasamy-jmlr20a} or even very simple schedules switching from \ac{EI} to \ac{PI} can improve anytime performance of \ac{BO}; however, for each problem different schedules, incl. static ones, perform best~\citep{benjamins-corr22a}.

Performance can be improved by selecting an \ac{AF}-schedule with a meta-learned selector based on the \ac{ELA} features~\citep{mersmann-gecco11a} of the initial design which factors in the problem at hand ~\citep{benjamins-meta22a}.
Nevertheless, this approach has its limitations.
First, it requires a large and expensive initial design compared to the overall budget in order to compute the \ac{ELA} features, and the ideal size of it is unknown~\citep{belkhir-lion16a}.
Second, the selector is trained for a specific budget, and it is unclear how it transfers to other dimensions, optimization budgets, or initial designs.

In this work, we instead aim for a \textit{self-adjusting yet simple} approach to adapt the exploration-exploitation trade-off in a data-driven way throughout the optimization process. 
For this, we propose to adaptively set the weight \weiw of \ac{WEI}~\citep{sobester-jgo05a} in an online parameter control fashion~\citep{karafotias-tec15a,doerr-toec20a}. 
Depending on how we parametrize \ac{WEI}, we can be more explorative, recover \ac{EI}, or lean towards a modulated, exploitative \ac{PI}.
The crucial questions to answer here are \begin{enumerate*}[label={(\roman*)}]\item \textit{When} should we adjust \weiw? and \item \textit{How} should we adjust \weiw? \end{enumerate*}

We propose a new method, dubbed \acf{SAWEI}.
Inspired by a termination criterion for \ac{BO}~\citep{makarova-automl22}, we adjust the weight \weiw whenever \ac{BO} tends to converge, indicated by the \acf{UBR}.
We adjust \weiw opposite to the dominant search attitude, either towards exploration or exploitation.
The key mechanism behind \ac{SAWEI} is illustrated in~\cref{fig:SAWEI}.
We demonstrate the effectiveness of our method \ac{SAWEI} on the BBOB functions of the COCO benchmark~\citep{hansen-oms20a} and on tabular benchmarks from HPOBench~\citep{eggensperger-neuripsdbt21a} against baselines of established \ac{AF}s and previously proposed handcrafted \ac{AF}-schedules for \weiw.

\section{Related Work}
\begin{rev}
One line of works directly focuses on improving \ac{AF}s~\citep{qin-nips17a,balandat-neurips20a,volpp-iclr20a}.
To overcome the fact that EI can sometimes be too exploitative, \cite{qin-nips17a} uniformly sample one of the two most promising points instead of always choosing the most promising one according to~EI.
\cite{balandat-neurips20a} offer efficient implementation of Monte-Carlo \ac{AF}s (no closed-form solution available) as well as a one-shot formulation of the Knowledge Gradient.
A different approach is to meta-learn a neural \ac{AF} via Reinforcement Learning to achieve better sample-efficiency on downstream tasks~\citep{volpp-iclr20a}.
\end{rev}
A different line of work is concerned with combining different \ac{AF}s, e.g.,~by building a portfolio of \ac{AF}s (\ac{EI}, \ac{PI}, \ac{UCB} with different hyperparameter settings) and then using an online multi-armed bandit strategy to assign probabilities of which \ac{AF} to use at which step, called GP-Hedge or Portfolio Allocation~\citep{hoffman-uai11a}.
Their work indicates that the performance of \revt{Portfolio Allocation} highly varies with the number of arms and their respective hyperparameter settings.
Similarly to Portfolio Allocation, \citet{kandasamy-jmlr20a} update weights of their portfolio (\ac{UCB}, \ac{EI}, \ac{TS}~\citep{thompson-biomet33a}, \ac{TTEI}~\citep{qin-nips17a}) in an online manner.
They do not include \ac{PI} as they observe it exhibits inferior performance compared to other single static \ac{AF}s.
In addition, robust versions of \ac{EI}, \ac{PI}, and \ac{UCB} can be combined to a multi-objective \ac{AF} combining the strengths of the individual ones~\citep{cowenrivers-arxiv21a}.
In this work, we take a step back and ask ourselves what we could achieve by employing a simplistic approach of self-adjusting the exploration-exploitation trade-off of \ac{WEI}.

It has also been shown in other optimization-related areas that dynamic choices are beneficial in terms of performance, e.g.,~in evolutionary computation~\citep{karafotias-tec15a,doerr-toec20a}, planning~\citep{speck-icaps21a} and deep learning~\citep{adriaensen-arxiv22a}. 
Recently, the introduction of \ac{DAC}~\citep{biedenkapp-ecai20a} underlines the potential of employing dynamic schedules (as opposed to selecting algorithm components \emph{on the fly}, as is usually done in evolutionary computation~\citep{hansen-ec03a}).

Related to that, also setting the weight \weiw of \ac{WEI} has been investigated.
\citet{sobester-jgo05a} propose to cycle through $\alpha \in \{0.1, 0.3, 0.5, 0.7, 0.9 \}$ to pulse from exploring to exploiting.
This idea is based on the suggestion to cycle through global-local balances during the search~\citep{gutmann-jgo01a}.
However, this heuristic is oblivious to the current state of the search.
Another line of work proposes to simply query \ac{WEI} $n$ times with $n$ different values of \weiw in parallel~\citep{liu-asme18a} with the drawback of potentially uninformative function evaluations.
The weights for the exploration and exploration terms in \ac{WEI} can also be set via rewards obtained by calculating the accuracy of the surrogate model~\citep{xiao-compel12a, xiao-ieee13a}.
However, the definition of the rewards is lacking and their method needs to be reset from time to time for the case when the exploration term causes to repeatedly propose the same configuration.

\section{Self-Adjusting Weighted EI}
\begin{figure}[tbp]
    \centering
      \includegraphics[width=0.8\textwidth]{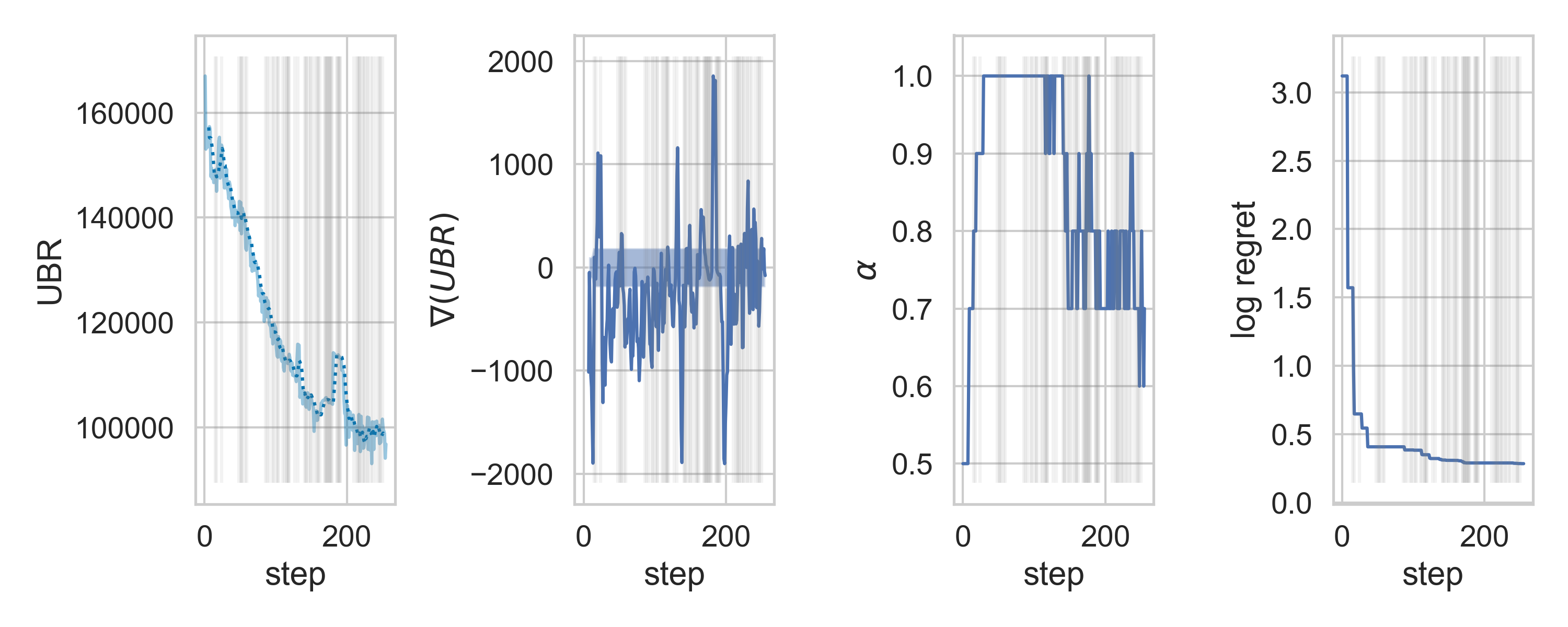}
      \caption{With \ac{SAWEI} we self-adjust the exploration-exploitation trade-off parameter \weiw based on the \acf{UBR} (left). Whenever the gradient of \ac{UBR} (2nd left) becomes \revt{approximately} \num{0} (marked by vertical lines), we adjust \weiw (2nd right), further reducing the log regret (right). BBOB F20, 8d.}
      \label{fig:SAWEI}
\end{figure}
In our method, the \acf{SAWEI}, we adaptively set the weight \weiw$\in[0,1]$ of the \acf{WEI} to steer the exploration-exploitation trade-off.
\ac{WEI}~\citep{sobester-jgo05a} is defined as:
\newcommand{\vecx}{\textbf{x}}
\begin{equation}
\small
WEI(\vecx; \textcolor{violet}{\alpha}) = \textcolor{violet}{\alpha} \, \underbrace{z(\vecx) \hat{s}(\vecx) \Phi \left[z(\vecx)\right]}_{\small\text{exploitation-driven}}
+ \,
\textcolor{violet}{(1 - \alpha)} \, \underbrace{\hat{s}(\vecx) \phi \left[z(\vecx)\right]}_{\small\text{exploration-driven}} 
\label{eq:WEI}
\end{equation}
with $z(\vecx) = (f_{\text{min}} - \hat{y}(\vecx)) / \hat{s}(\vecx)$, $f_{\text{min}}$ being the lowest observed function value, $\hat{y}(\vecx)$ and $\hat{s}(\vecx)$ the predicted mean and standard deviation from the surrogate model, and $\phi$ and $\Phi$ being the PDF and CDF of a Gaussian distribution, respectively.
The \weiw coefficient weighs the exploration and exploitation terms.
For example, $\alpha=0.5$ recovers standard EI~\citep{mockus-tgo78a} and $\alpha=1$ has a similar behavior as $PI(\vecx) = \Phi \left[z(\vecx)\right]$~\citep{kushner-jfe64a}.
\revtwot{With $\alpha=0$ we only utilize the exploration term, but this does not equal pure exploration or complete randomness.}

\paragraph{When To Adjust}
In order to be able to set \weiw adaptively, we need an indicator of the progress of the optimization.
Recently,~\cite{makarova-automl22} proposed a termination criterion to stop \ac{BO} for hyperparameter optimization.
If the \acf{UBR} falls under a certain threshold, they terminate.
\ac{UBR} estimates the true regret at iteration $k$ by:
\begin{equation}
     \text{UBR}(G_k;\mathcal{X}) = r_k := \min_{\vecx \in G_k} \text{UCB}_k (\vecx) - \min_{\vecx \in \mathcal{X}} \text{LCB}_k (\vecx)
\label{eq:ubr}
\end{equation}
with $G_k$ being the history of all evaluated points, $\mathcal{X}$ being the entire search space, and LCB and UCB being the lower and upper confidence bound, e.g., $UCB(x) = \mu_t(x) + \beta_t \sigma_t(x)$ and $LCB(x) = \mu_t(x) - \beta_t \sigma_t(x)$, respectively. 
The first term of \ac{UBR} estimates the worst-case function value of the best-observed point, a.k.a. the incumbent, and the second term is the lowest function value across the whole search space.
This means the smaller the gap between both terms becomes, the closer we are to the asymptotic function value \textit{under the current settings of the optimizer}.
\revt{We empirically show that the UBR indeed changes after we change the acquisition function during the optimization in~\cref{sec:ubr_anaysis}, supporting our intuition.}
The \ac{UBR} does not directly operate on function values, but UCB and LCB are computed on the surrogate model instead.
Instead of using \ac{UBR} to stop the optimization process, it serves as an indicator for us when to adjust components\revt{, i.e., update the value of \weiw}.

\noindent
\textbf{Our rule is:} When the gradient of \ac{UBR} over the last $n$ steps becomes close to \num{0}, we adjust the exploration-exploitation attitude with \weiw.
The sensitivity to the gradient is controlled by our hyperparameter $\epsilon$.

\paragraph{How to Adjust}
\label{sec:how_to_adjust}
The remaining question is \textit{how} to adjust \weiw, by how much and into which direction.
We propose a rather simple, yet effective additive change by $\Delta_\alpha$.
\revt{Our intuition is to set \weiw opposite to the current \textit{search attitude}, since the current search attitude led to convergence of the optimization.
The term search attitude describes the current search behavior, whether the acquisition function is more explorative or more exploitative.}
We set $\Delta_\alpha = 0.1$ to allow for gradual changes.
We determine the sign of $\Delta_\alpha$ by the recent \revt{search} attitude: depending on whether the exploration\revt{-term $a_{\text{explore}}$} or exploitation-term \revt{$a_{\text{explore}}$} of \cref{eq:WEI} is larger for the last selected point \revt{$\vecx_{\text{next}}$}, the current search attitude was either steered more for exploring or exploiting, respectively.
\revt{The terms are the summands of \ac{WEI} and defined as follows:
\begin{align}
    \label{eq:search_attitude}
    a_{\text{explore}}(\vecx_{\text{next}}) &= \hat{s}(\vecx_{\text{next}}) \phi \left[z(\vecx_{\text{next}})\right]\\
    a_{\text{exploit}}(\vecx_{\text{next}}) &= z(\vecx_{\text{next}}) \hat{s}(\vecx_{\text{next}}) \Phi \left[z(\vecx_{\text{next}})\right]
\end{align}
We use $a_{\text{exploit}} = \Phi \left[z(\vecx_{\text{next}})\right]$, omitting $z(\vecx_{\text{next}}) \hat{s}(\vecx_{\text{next}})$, which is equal to PI.\footnote{Empirically, both methods perform almost equivalent for BBOB but not for HPOBench, see~\cref{sec:search_attitude}. We conjecture that original $a_{\text{exploit}}$ is less exploitative than the original PI. Since we look for a strong (global) signal on how exploitative a point was, we opted for PI instead of the WEI term.}
Please note that we only do this for determining the search attitude.
Now if the exploration term is bigger than the exploitation term, i.e.,  $a_{\text{explore}} > a_{\text{exploit}}$, the current search attitude is exploration.}
We inspect the attitude and adjust \weiw in the \textit{opposite} direction, to provide a chance for more exploration or exploitation in contrast to the currently dominating attitude.

\paragraph{SAWEI in a Nutshell}
We illustrate and summarize our method \ac{SAWEI} in~\cref{fig:SAWEI} and in~\cref{alg:sawei}.
Our goal is to adjust the exploration-exploitation trade-off based on the current search attitude whenever the \acf{UBR} converges.
\ac{SAWEI} enhances the standard \ac{BO} pipeline by calculating the \ac{UBR} in each iteration and by tracking the search attitude via the exploration term and the exploitation term of \ac{WEI}.
First, we define and evaluate the initial design and train our surrogate model (Line~1).
Then, as long as we have function evaluations left (Line~2), we query the acquisition function (here \acf{WEI}) for the next point to be evaluated (Line~3).
Meanwhile, we track the search attitude with the exploration and exploitation terms of \ac{WEI} (Line~4\revt{, see~\cref{eq:search_attitude}}).
The function is evaluated as usual with the proposed point and we update our history and our surrogate model (Lines~5-7).
Now we calculate the \ac{UBR} estimating the gap to the true regret based on the history of evaluated points and the search space (Line~8).
We smooth the history of \ac{UBR} with moving \ac{IQM} \revt{(\SI{25}{\percent}-\SI{75}{\percent} quartiles) with a window size of \num{7}} (Lines~9-10, \texttt{smooth\_with\_iqm}).
Based on this smoothed version, we check whether \ac{UBR} has converged, i.e., the gradient of \ac{UBR} is close to 0 (Line~11).
\revt{In more detail, we signal time to adjust when the last absolute gradient is close to \num{0} with an absolute tolerance of $\epsilon$ times the last observed maximum of the absolute gradient.}
If it is the case, we adjust the weight $\alpha$ of \ac{WEI} based on the search attitude (Line~12).
The search attitude is calculated with the exploration and exploitation terms of \ac{WEI}.

\begin{algorithm}[h]
\caption{Bayesian Optimization with \acf{SAWEI}}\label{alg:sawei}

\begin{algorithmic}[1]
\Require Initial weight of $\WEI$ $\alpha = 0.5$, history of evaluated points $G = \emptyset$, history of regret estimates/UBR $R$, surrogate model $\mathcal{M}$, function to optimize $f$
\State Evaluate initial design and train surrogate model $\mathcal{M}$
\While{Optimization Budget Not Exhausted}
\State $x_{\text{next}} \gets \WEI(\mathcal{M})$  \Comment{Propose next configuration to evaluate}
\State $a_{\text{explore}}, a_{\text{exploit}} \gets \WEI(\revt{x_{\text{next}}})$ \Comment{Get summands of $\WEI\revt{(x_{\text{next}})}$ before $\mathcal{M}$ is trained}
\State $y \gets f(x_{\text{next}})$  \Comment{Evaluate function}
\State $G \gets G \cup \{x_{\text{next}} \}$  \Comment{Update history}
\State Train surrogate model $\mathcal{M}$
\State $r \gets \text{UBR}(G, \mathcal{X})$    \Comment{Upper Bound Regret (UBR) estimate, \cref{eq:ubr}}
\State $R \gets R\texttt{.append}(r)$
\State $\bar{R} \gets $ \texttt{smooth\_with\_iqm}($R$)  \Comment{Smooth rugged signal with moving IQM}
\textcolor{violet}{
\If{$\nabla \bar{R} \approx 0$}  \Comment{Check if UBR converged}
    \State $\alpha \gets \text{\texttt{adjust}}(\alpha, a_{\text{explore}}, a_{\text{exploit}})$ \Comment{Adjust exploration-exploitation based on attitude}
\EndIf
}
\EndWhile
\end{algorithmic}
\end{algorithm}

\section{Experiments}
In our experiments, we empirically evaluate our method \ac{SAWEI} on different benchmarks and compare it to baselines from the literature and handcrafted ones.
We benchmark the algorithms on the BBOB functions from the COCO problem suite~\citep{hansen-oms20a} and on HPOBench~\citep{eggensperger-neuripsdbt21a}.
Our implementations are built upon the \ac{BO} tool SMAC3 (v2.0.0b1)~\citep{lindauer-jmlr22a}.
\revt{We use a standard GP as configured in SMAC's BlackBoxFacade and SMAC optimizes the acquisition function with a combination of local and random search which also applies to minimizing LCB in~\cref{eq:ubr} for calculating the \ac{UBR}. 
We set $\beta_t = 2 \log (d  t^2 / \beta), \beta = 1$ for UCB/LCB as done in SMAC following the original UCB~\citep{srinivas-icml10a}.}
The code is available at \url{https://github.com/automl/SAWEI}.
The exact setting for our method is $\epsilon=0.1$ and adding or subtracting $\Delta\alpha=0.1$.
We set our convergence check horizon to $n=1$, i.e., we check whether the last gradient is close to \num{0}.
We validate our hand-crafted settings an ablation study in~\cref{sec:ablation}.

Our evaluation protocol repeats the optimization \num{10} times with different random seeds and calculates the \acf{IQM} across seeds to robustly estimate the regret per function.
For each schedule, we then determine the rank for each of the \num{24} BBOB functions and compute the global rank across functions.
For the rank table, we aggregate the ranks across the single tasks per schedule with the \ac{IQM}.
In the plots over optimization steps, we show the mean and $\SI{95}{\percent}$ confidence interval across all the functions.

\paragraph{BBOB} For the \num{24} noiseless, synthetic BBOB functions~\citep{hansen-oms20a} we set the dimensionality to \num{8}, the budget of the initial design to \num{24} function evaluations (FEs), and the budget for the surrogate-based optimization to \num{256} FEs.
We optimize the first three instances of each function. In BBOB, the instances are obtained by scaling, shifting, and rotating the base function (hence preserving the problem structure but changing the embedding). 

\paragraph{HPOBench} We evaluate all methods on the tabular machine learning benchmarks from HPOBench~\citep{eggensperger-neuripsdbt21a}.
To this end, we randomly selected \revt{eight} tasks from the OpenML dataset~\citep{casalicchio-17a,feurer-jmlr21a} and optimize a Random Forest, MLP, SVM, Logistic Regression, and XGBoost. 
We allow an initial design of \num{15} FEs and a BO-based optimization budget of \num{100} FEs.
For each FE, we average the metric over the five available seeds.

\paragraph{Baselines}
We compare our data-driven, self-adjusting method \ac{SAWEI} to \begin{enumerate*}[label={(\roman*)}]
\item the well-established best practice of simply using a single \ac{AF} \revt{(EI, PI, and LCB)} and \item hand-designed schedules of $\alpha$
\end{enumerate*},  see \cref{tab:baselines}.
We start with static schedules of $\alpha \in \{0, 0.5, 1\}$, \revt{either more exploring, \ac{EI}, or more exploiting}.
Further, we define a schedule from \ac{EI} \revt{($\WEI(\alpha=0.5)$)} to modulated \ac{PI} \revt{($\WEI(\alpha=1)$),} and vice versa\revt{,} as a step function with \num{5} steps.
In addition, we compare to hard switches from \ac{EI} to \ac{PI}~\citep{benjamins-meta22a} as well as the Gutmann-Sobester pulse cycling through $\alpha$~\citep{gutmann-jgo01a,sobester-jgo05a}.
We also include Portfolio Allocation~\citep{hoffman-uai11a} and use their portfolio of nine acquisition functions consisting of different parametrizations of \ac{UCB}, \ac{PI}, and \ac{EI}.

\begin{table}[t]
    \centering
    \caption{Baselines.}
    \label{tab:baselines}
    \footnotesize
    \begin{tabular}{cc}
         \toprule
         \revt{$\WEI(\alpha=0)$ (}Explore\revt{)} & $\alpha = 0.0$ \\
         \revt{$\WEI(\alpha=0.5)$ (}\ac{EI}\revt{)} & $\alpha = 0.5$ \\
         \revt{$\WEI(\alpha=1)$ (modulated }\ac{PI}\revt{)} & $\alpha = 1.0$ \\
         \midrule
         \revt{$\WEI(\alpha=0.5)$} $\rightarrow$ \revt{$\WEI(\alpha=1)$} (\revt{Steps}) & \num{5} steps \\
         \revt{$\WEI(\alpha=1)$} $\rightarrow$ \revt{$\WEI(\alpha=0.5)$} (\revt{Steps}) & \num{5} steps \\
         \midrule
         \ac{EI} $\rightarrow$ \ac{PI} & switch after $\SI{25}{\percent}$ \\
         \ac{EI} $\rightarrow$ \ac{PI} & switch after $\SI{50}{\percent}$ \\
         \ac{EI} $\rightarrow$ \ac{PI} & switch after $\SI{75}{\percent}$ \\
         \midrule
         Gutmann-Sobester Pulse~\citep{gutmann-jgo01a,sobester-jgo05a} & Cycle $\alpha \in [0.1, 0.3, 0.5, 0.7, 0.9]$ \\ 
         \midrule
         Portfolio Allocation~\citep{hoffman-uai11a} & - \\
         \bottomrule
    \end{tabular}
\end{table}

\begin{figure}[h]
    \centering
    \includegraphics[width=0.65\textwidth,keepaspectratio,height=5cm]{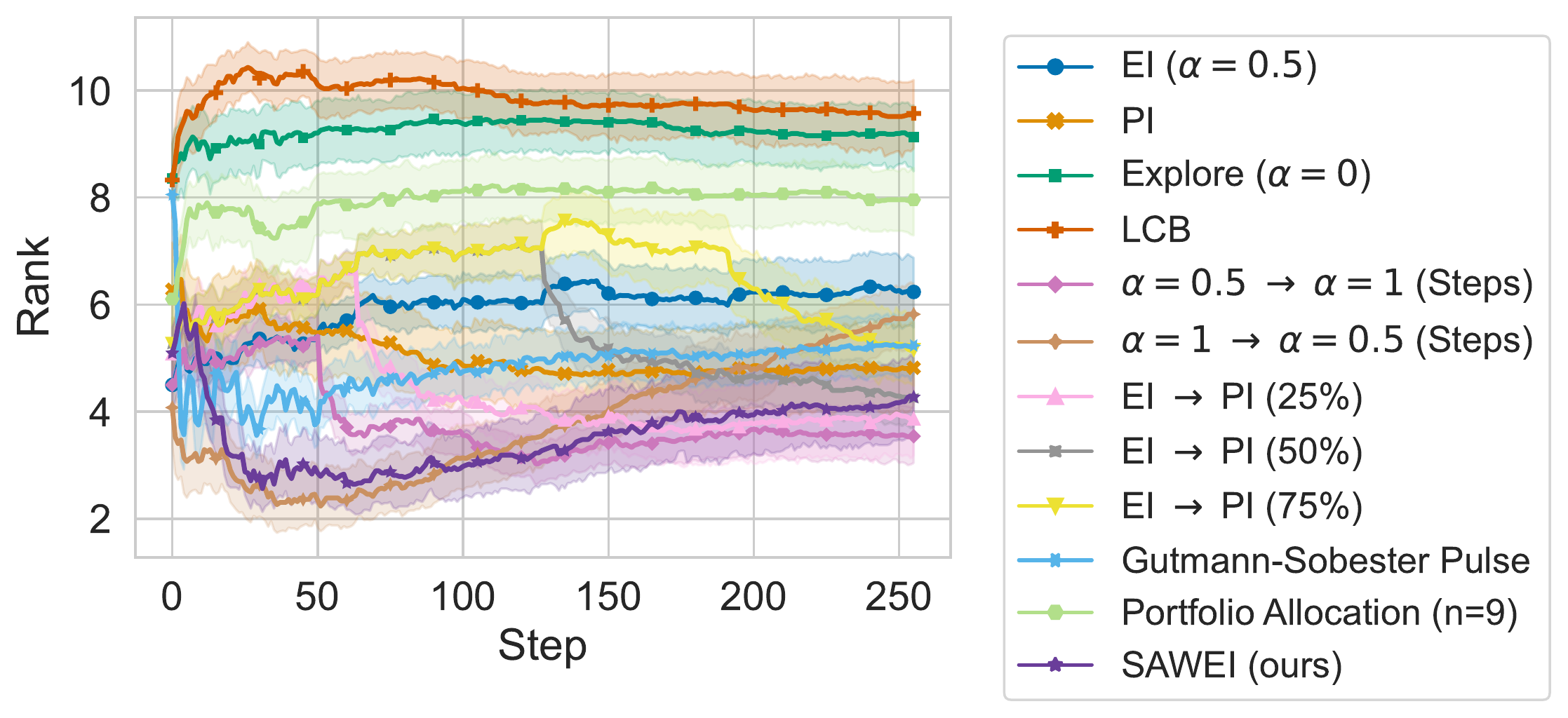}
    \caption{Ranks per Step on BBOB}
    \label{fig:bbob_ranks_per_step}
\end{figure}

\subsection{Results}
\begin{figure}[htb]
    \centering
    \begin{subfigure}[t]{0.75\textwidth}
         \centering
        \includegraphics[width=\textwidth]{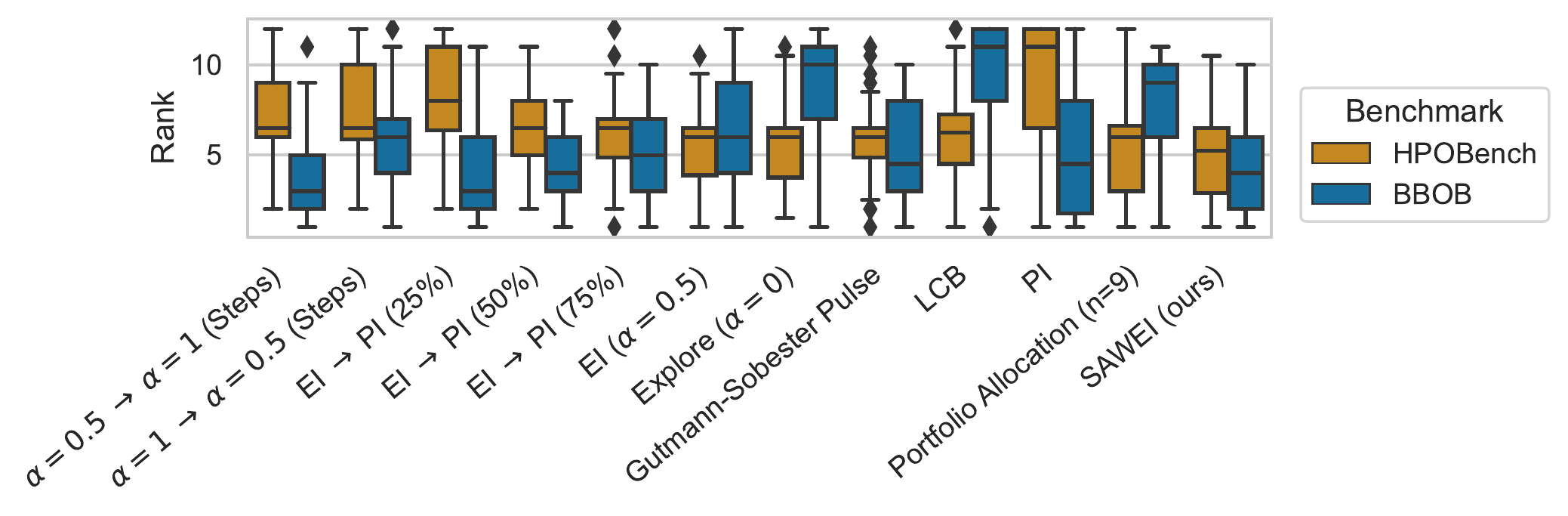}
        \caption{Our method \ac{SAWEI} has the most favorable rank distribution \textit{across domains}. That is, even though on different benchmarks other schedules perform on par, their suitability highly differs depending on the benchmark. \revt{Ranks are computed on final performance.}}
        \label{fig:rank_comparison}
    \end{subfigure}
    \hfill
    \begin{subfigure}[t]{0.2\textwidth}
         \centering
        \includegraphics[width=\textwidth]{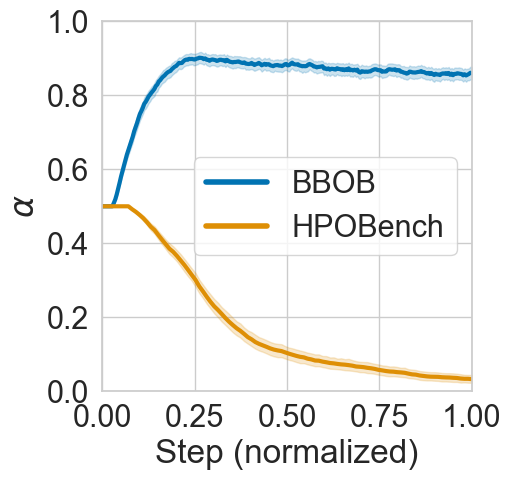}
        \caption{
        \revt{
        \weiw traversed by \ac{SAWEI} which adapts to the different benchmarks.}}
        \label{fig:alpha_bbob_vs_hpobench}
    \end{subfigure}
    \caption{Comparison of BBOB and HPOBench}
\end{figure}

\paragraph{BBOB}
Our method \ac{SAWEI} ranks among the first based on final performance (cf.~\cref{fig:rank_comparison}), which is very similar to dynamic baselines going from EI ($\alpha = 0.5$) to the modulated PI ($\alpha = 1$). 
One drawback of the hand-designed schedules is that the optimization budget needs to be defined beforehand, whereas our method is self-adjusting and is oblivious of the total budget.
Surprisingly, the modulated PI is comparatively strong and performs better than EI, suggesting that the BBOB landscapes require a higher percentage of exploitation.
\ac{SAWEI} also exhibits a favorable anytime performance, making it a consistent and robust default choice, see~\cref{fig:bbob_ranks_per_step}.
Schedules dominating \ac{SAWEI} only do so for a portion of the optimization, hence they are not consistent.
Confirming results from~\cite{benjamins-corr22a}, the effect of switching from EI to PI can be clearly seen as a boost in the ranks.
On BBOB, the general\revt{ly} well-performing schedules involve PI which our method can easily mimic.
\ac{SAWEI} finds a suitable transition from exploring to exploiting per-run.

\begin{figure}[htb]
    \centering
    \begin{subfigure}[b]{0.3\textwidth}
         \centering
         \includegraphics[width=\textwidth]{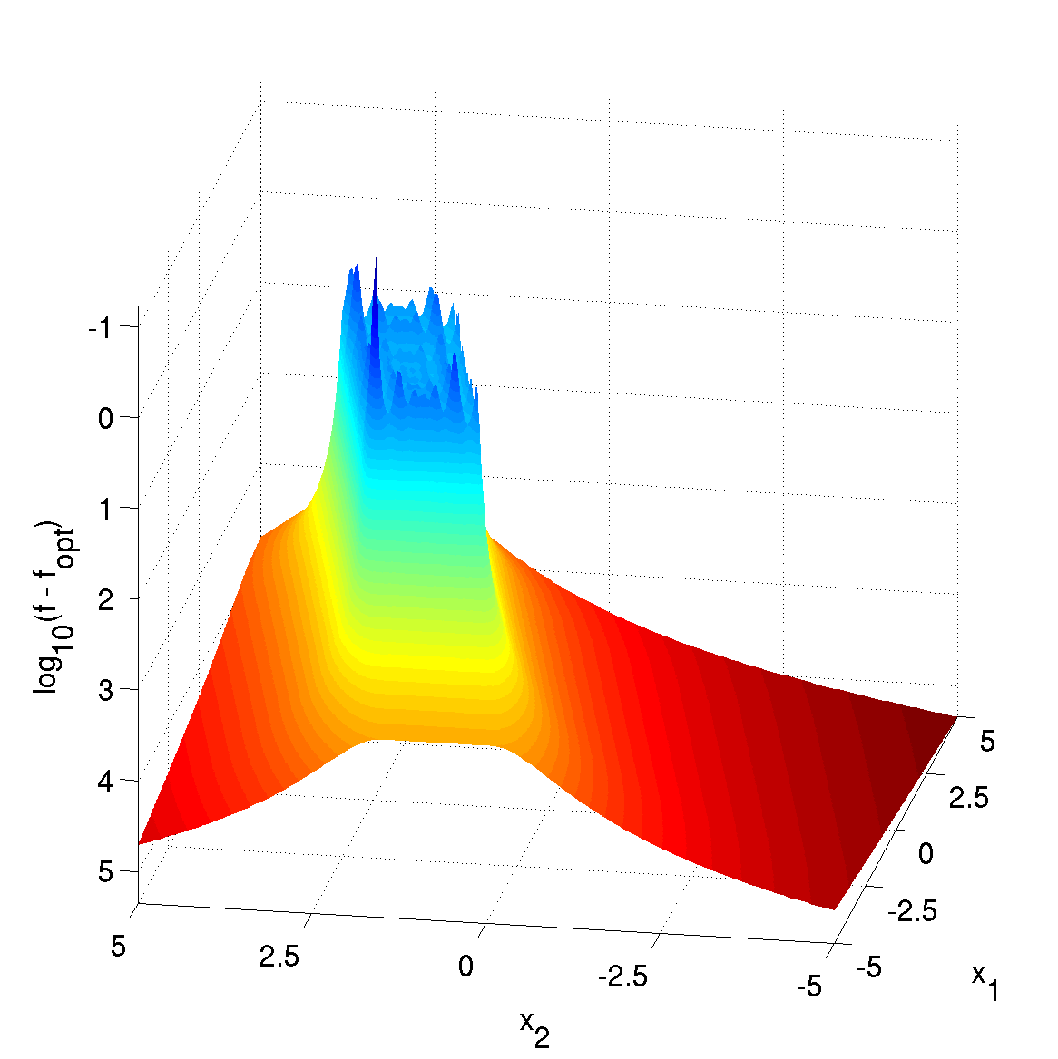}
         \caption{BBOB F20: Schwefel Function~\citep{hansen-oms20a} (Image source~\citep{finck-report09})}
         \label{fig:f20}
    \end{subfigure}
    \hfill
    \begin{subfigure}[b]{0.67\textwidth}
         \centering
         \includegraphics[width=\textwidth]{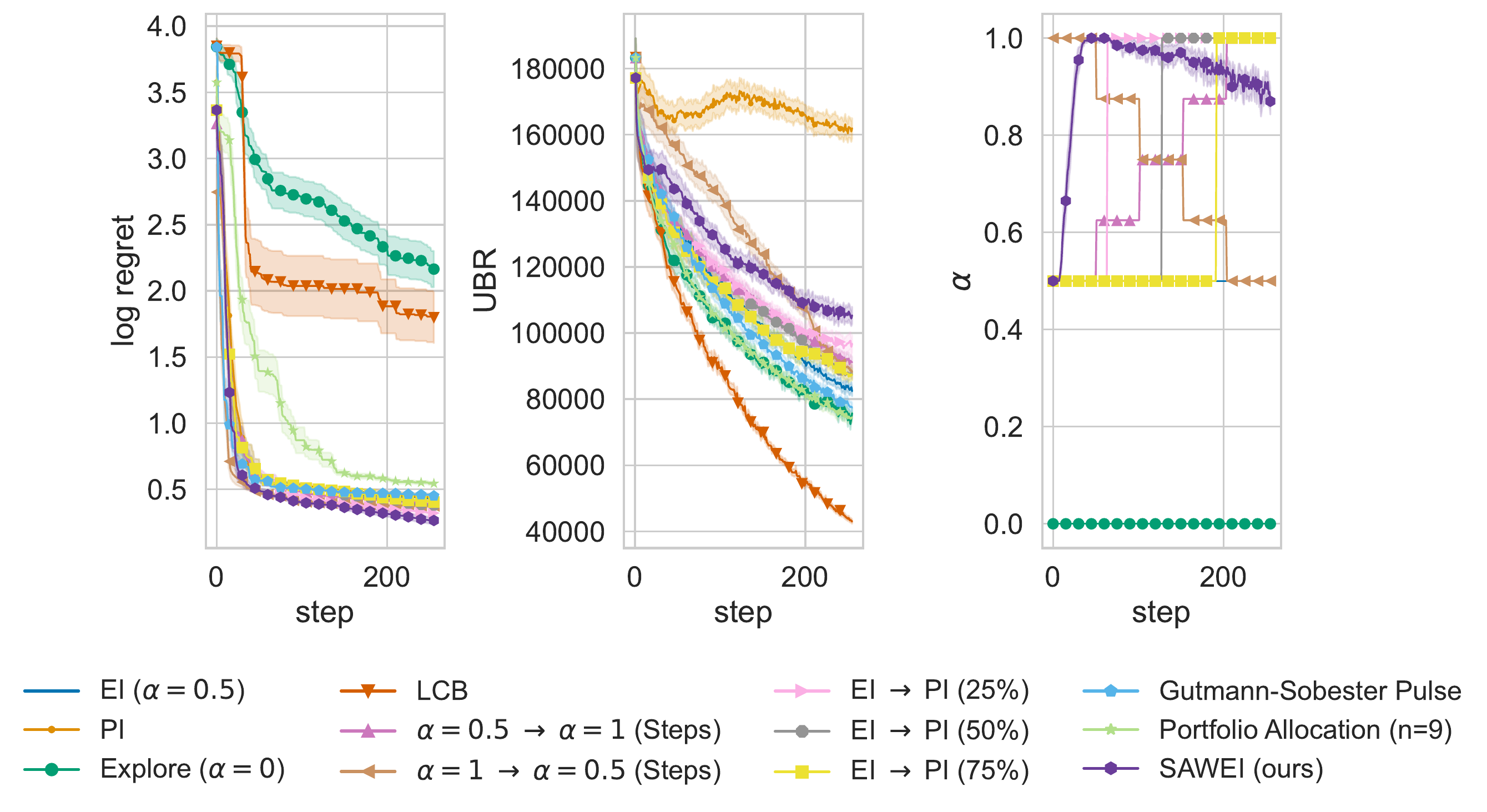}
         \caption{F20: Log regret, \ac{UBR} and $\alpha$-schedules}
         \label{fig:alpha_f20}
    \end{subfigure}\\
    \begin{subfigure}[b]{0.3\textwidth}
         \centering
         \includegraphics[width=\textwidth]{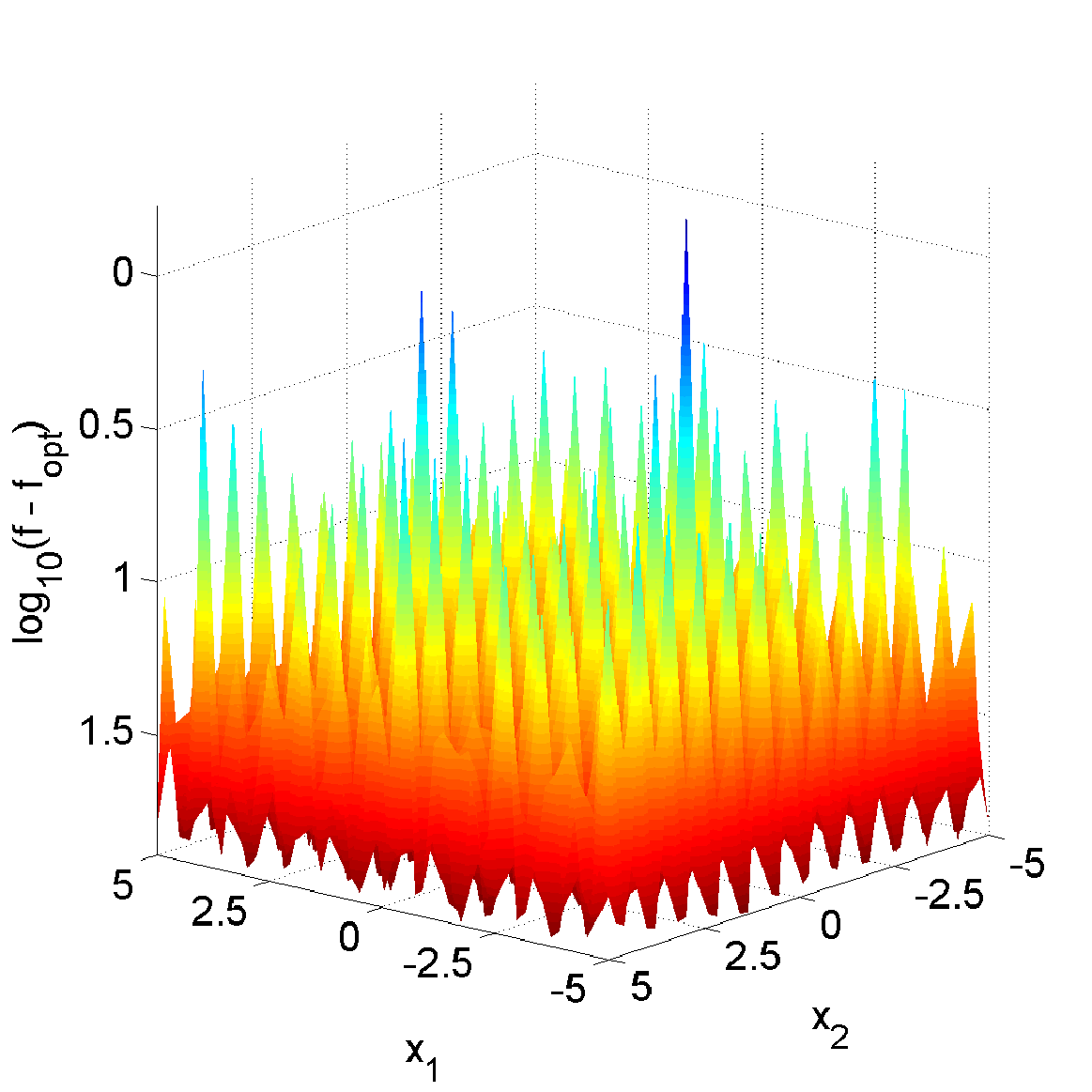}
         \caption{BBOB F23: Katsuura Function~\citep{hansen-oms20a} (Image source~\citep{finck-report09})}
         \label{fig:f23}
    \end{subfigure}
    \hfill
    \begin{subfigure}[b]{0.67\textwidth}
         \centering
         \includegraphics[width=\textwidth]{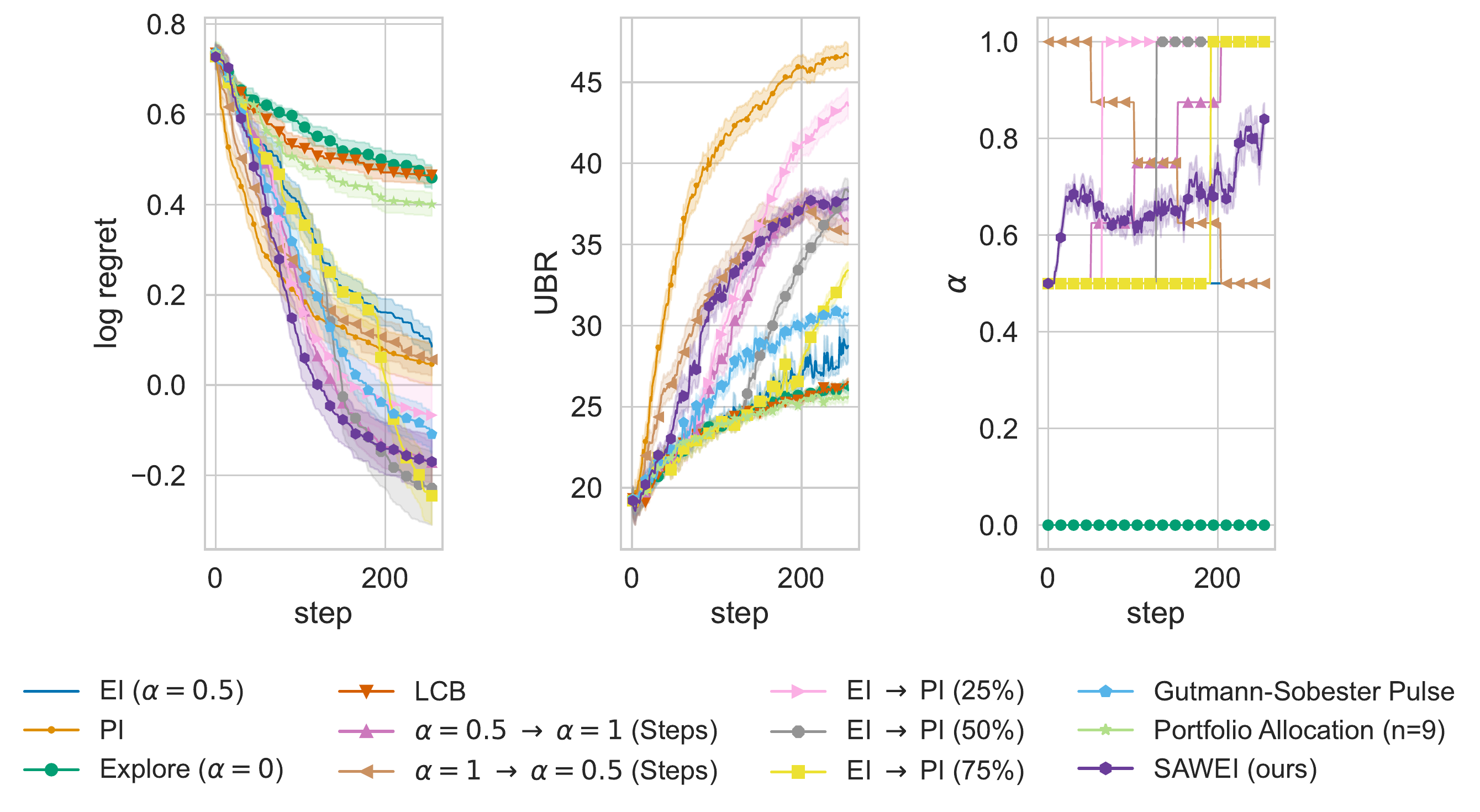}
         \caption{F23: Log regret, \ac{UBR} and $\alpha$-schedules}
         \label{fig:alpha_f23}
    \end{subfigure}
    \caption{BBOB Functions F20 and F23 with different performance and behavior indicators}
    \label{fig:bbob_alpha}
\end{figure}

In general, the tendency of the $\alpha$-schedules traversed by \ac{SAWEI} is moving from exploration to exploitation.
Often, we can observe a decrease, a change to more exploration again, after some iterations.
On \revt{one} BBOB function, the multi-modal Schwefel function (F20) with weak global structure (\cref{fig:f20}), \ac{SAWEI} manages to efficiently transform from EI ($\alpha = 0.5$, higher explorative attitude) to modulated PI ($\alpha = 1$) with an exploitative attitude.
At the end of the optimization, when the basin was already discovered, \ac{SAWEI} decreases $\alpha$ to more exploration to explore the surroundings.
We can also clearly observe the effect of the hand-designed switching \revt{(EI $\rightarrow$ PI (x $\SI{}{\percent}$)) in the sharp bends downwards in the log regret and upwards in the \ac{UBR}}, although \ac{SAWEI} discovers a more suitable point and can change its attitude again.
On Katsuura\revt{, which is} highly multi-modal \revt{and has} weak global structure \revt{(\cref{fig:f23}),} \ac{SAWEI} \revt{increases} $\alpha$ more slowly to exploitation, presumably because of the highly rugged landscape\revt{, see~\cref{fig:alpha_f23}}.
Also here, \ac{SAWEI} discovers the boost from changing from exploration to exploitation.
If we look closely we can see that the \acl{UBR} jumps up after the switch happened for the switching schedules (EI to PI) which is an indication of the adequacy of \ac{UBR} as a state descriptor.
All schedule plots for each BBOB function, as well as the box plots of the final log regrets can be found in~\cref{sec:app_bbob_results}.

\paragraph{HPOBench}
\begin{figure}[t]
    \centering
    \includegraphics[width=0.65\textwidth,keepaspectratio,height=5cm]{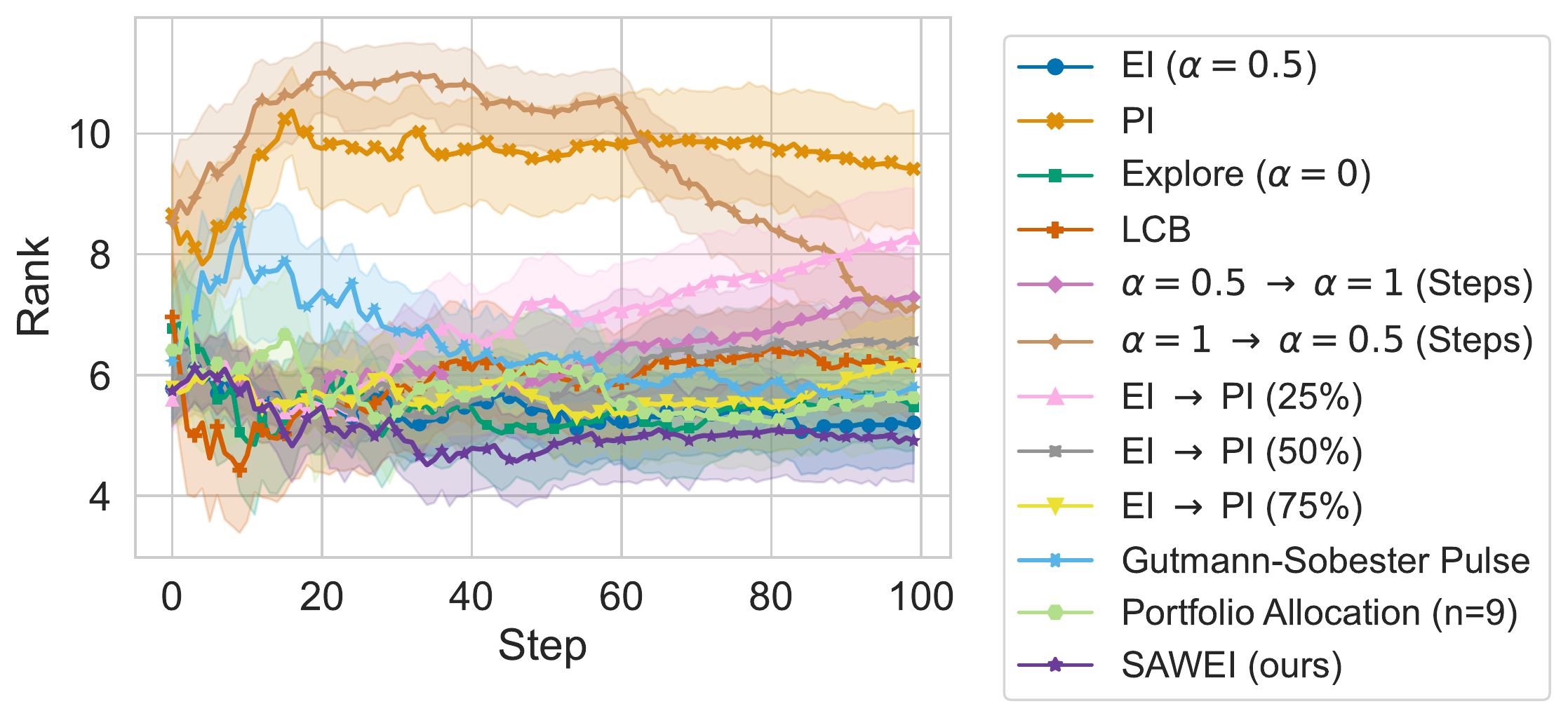}
    \caption{Ranks per Step on HPOBench}
    \label{fig:hpob_ranks_per_step}
\end{figure}
On HPOBench we see that \ac{SAWEI} also has a favorable anytime performance, see~\cref{fig:hpob_ranks_per_step}, and ranks among the first for the final log regret (\cref{fig:rank_comparison}). 
It is on par with Explore ($\alpha = 0$), and they are directly followed by Portfolio Allocation and EI.
The supremacy of the exploratory schedules is quite surprising, given the simplicity commonly attributed to response landscapes in HPO~\citep{pushak-ppsn18b}. We will investigate this further in our future work.
With a closer look at the schedules, we see the general trend to start from EI ($\alpha=0.5$) and go to Explore ($\alpha = 0$) which is the complete opposite of the BBOB behavior.
Boxplots of the final log regret and all plots with log regret, UBR and $\alpha$ over time can be found in~\cref{sec:app_hpob_results}.

\begin{rev}
\paragraph{Comparison of BBOB and HPOBench}
\end{rev}
In summary, we observe that the optimal schedule and search behavior vary on two levels.
First, for a given problem type, the optimal schedule varies across the single tasks.
Second, the search behavior depends on the type of problem, whether we optimize synthetic functions in BBOB or find optimal hyperparameters for machine learning models in HPOBench.
\ac{SAWEI} mimics the strategy fitting best to the problem at hand and exhibits the most favorable rank distribution \textit{across domains}, see~\cref{fig:rank_comparison}.
\begin{rev}
BBOB in general requires more exploitation and HPOBench more exploration which is visible prominently in two ways.
First, PI performs better on BBOB than on HPOBench and EI vice versa.
Second, \ac{SAWEI}'s trajectories of \weiw are contrary on BBOB and HPOBench (see~\cref{fig:alpha_bbob_vs_hpobench}) and thus adjust to the required search attitude.
\end{rev}

\paragraph{Ablation on BBOB}
\label{sec:ablation}
\revt{We perform an ablation study to assess the sensitivitiy of our method to its hyperparameters.}
In particular, we vary $\Delta\alpha \in \{0.05, 0.1, 0.25\}$, i.e., the amount to add or subtract to our current weight $\alpha$.
In addition, we can track the attitude in different ways: either just considering the last step \revt{(\texttt{last})}, or accumulating the terms until the last \revt{point where the best configuration (the incumbent)} change\revt{d (\texttt{inc.~change})} or until the last \revt{adjustment happened (\texttt{last adjust})}.
In the latter cases, $a_{\text{explore}}$ and $a_{\text{exploit}}$ become sums.
This hyperparameter defines the convergence check horizon $n$, which is varied during the run for the latter two options.
Finally, we vary the sensitivity to the gradient of UBR by the width of the tolerance band when compared to 0: $\epsilon \in \{0.05, 0.1, 0.5, 1 \}$.
The bigger $\epsilon$, the more often we switch.
\revt{We evaluate all \num{36} combinations on} all \num{24} BBOB functions with \num{10} seeds and \num{1} instance on \num{8} dimensions and assess the hyperparameter importance with fANOVA~\citep{hutter-icml14a}.
We normalize the log regret for each BBOB function and use this as the performance metric.

\begin{figure}[t]
    \centering
    \includegraphics[width=0.9\textwidth]{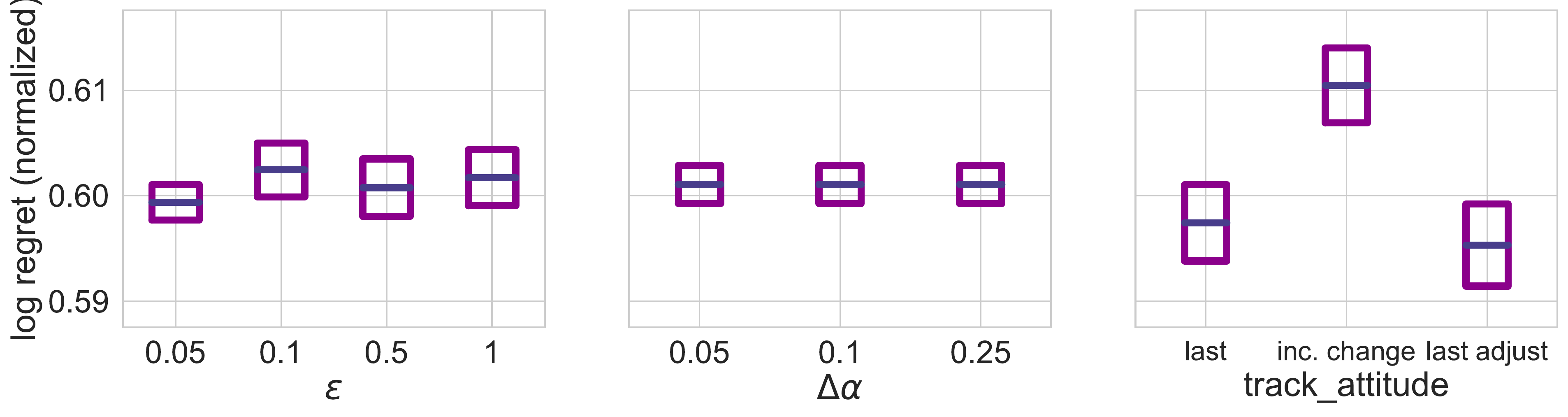}
    \caption{\revt{Marginal performances (mean and standard deviation) of }\ac{SAWEI}\revt{'s hyperparameters}}
    \label{fig:hp_importance}
\end{figure}

We show the marginals of each hyperparameter in~\cref{fig:hp_importance}.
The sensitivity $\epsilon$ to the gradient has a slight tendency to \revt{\num{0.05}} but the overall differences are small, we argue that the exact timing of the signal to adjust is less important.
In addition, setting the granularity of $\Delta \alpha$ is quite robust to the exact setting.
In contrast, tracking the attitude has a tendency to favor checking the exploration/exploration terms until the last \revt{adjust}.
It is likely that on other benchmarks the importances might change and our default of $\epsilon=0.1, \texttt{track\_attitude}=\text{last}, \Delta\alpha=0.1$ proves to be a robust one.

\section{Limitations and Future Work}
Our method \ac{SAWEI} introduces a slight overhead due to the need to optimize \ac{LCB} for computing \ac{UBR} in each iteration.
Everything that follows, namely deciding whether and how to adjust \weiw, is negligible in terms of computational cost.

In our analysis, we did not experiment with the initial value of \weiw, which may not be optimal for every tested function.
Also, our method does not allow jumps or resetting \weiw, which could also be beneficial.
In this context, defining \weiw directly as a function of the exploration/exploitation terms of \ac{WEI} could be a way to allow more flexibility.

One limitation is that so far we have only combined \ac{EI} and  \ac{PI}.
Our approach can easily be extended to \textit{any} linear combination of two acquisition functions.
Moreover, we can combine \ac{SAWEI} with \acf{DAC}~\citep{biedenkapp-ecai20a} to learn policies of \weiw across instances and tasks. 
More generally, we strongly believe that meta-learning and self-adjustment should go hand in hand, another topic to be explored in future work.
Building on the work by~\cite{benjamins-meta22a}, one could consider to warmstart \ac{SAWEI} using meta-models utilizing \ac{ELA} features~\citep{mersmann-gecco11a}.
\revt{Future work, and a current limitation, is the investigation of more domains as the domains show large variations.}
Finally, we believe that also other components of BO \revt{like the surrogate model} could benefit from self-adjusting choices.

\section{Conclusions}
Through a self-adjusting choice of the acquisition function in Bayesian Optimization, we aim to benefit from two main levers: (1) an automated identification of the AF best suitable for the unknown task at hand (e.g., while PI performs better than EI on BBOB, it is the other way around for HPO problems), and (2) an adjustment to the different needs during the optimization process.

Our method \ac{SAWEI} uses the convergence of \acf{UBR} as a criterion \revt{for} \emph{when} to adjust its parametrized acquisition function. 
SAWEI proves to achieve promising performance on two classic benchmark suites, BBOB and HPOBench, outperforming the static EI and PI AFs.
It is hence able to achieve both goals, (1) and (2), listed above.
It furthermore does not only achieve good final ranks, but also exhibits a favorable anytime performance on both suites. 

As a side result of our study, we observe that the general trends in BBOB and HPOBench are orthogonal to each other: while \ac{SAWEI} generally traverses from \ac{EI} (exploration) to a modulated \ac{PI} (exploitation) for BBOB, it moves from \ac{EI} to even more exploration on HPOBench.
This demonstrates the need for flexible, on-the-fly-adjustment of BO components. 

\vspace{1ex}
\textbf{Broader Impact Statement:} 
After careful reflection, the authors have determined that this work presents no notable negative impacts on society or the environment, since it presents a foundational approach without any concrete application at hand.

\begin{acknowledgements}
The authors gratefully acknowledge the computing time provided to them on the high-performance computers Noctua2 at the NHR Center PC2 under the project hpc-prf-intexml. These are funded by the Federal Ministry of Education and Research and the state governments participating on the basis of the resolutions of the GWK for the national high performance computing at universities (www.nhr-verein.de/unsere-partner).
Carolin Benjamins and Marius Lindauer acknowledge funding by the German Research Foundation (DFG) under LI 2801/4-1.
Elena Raponi acknowledges funding by the PRIME programme of the German Academic Exchange Service (DAAD) with funds from the German Federal Ministry of Education and Research (BMBF).
\end{acknowledgements}

\bibliography{bib/custom,bib/shortstrings,bib/lib,bib/shortproc}

%
%
%
%
%

\section{Submission Checklist}

\begin{enumerate}
\item For all authors\dots
  \begin{enumerate}
  \item Do the main claims made in the abstract and introduction accurately
    reflect the paper's contributions and scope?
    \answerYes{}
  \item Did you describe the limitations of your work?
    \answerYes{}
  \item Did you discuss any potential negative societal impacts of your work?
    \answerYes{}
  \item Have you read the ethics author's and review guidelines and ensured that
    your paper conforms to them? \url{https://automl.cc/ethics-accessibility/}
    \answerYes{}
  \end{enumerate}
\item If you are including theoretical results\dots
  \begin{enumerate}
  \item Did you state the full set of assumptions of all theoretical results?
    \answerNA{No theoretical results.}
  \item Did you include complete proofs of all theoretical results?
    \answerNA{}
  \end{enumerate}
\item If you ran experiments\dots
  \begin{enumerate}
  \item Did you include the code, data, and instructions needed to reproduce the
    main experimental results, including all requirements (e.g.,
    \texttt{requirements.txt} with explicit version), an instructive
    \texttt{README} with installation, and execution commands (either in the
    supplemental material or as a \textsc{url})?
    \answerYes{}
  \item Did you include the raw results of running the given instructions on the
    given code and data?
    \answerNo{We will upload the datasets upon acceptance.}
  \item Did you include scripts and commands that can be used to generate the
    figures and tables in your paper based on the raw results of the code, data,
    and instructions given?
    \answerYes{}
  \item Did you ensure sufficient code quality such that your code can be safely
    executed and the code is properly documented?
    \answerYes{}
  \item Did you specify all the training details (e.g., data splits,
    pre-processing, search spaces, fixed hyperparameter settings, and how they
    were chosen)?
    \answerYes{}
  \item Did you ensure that you compared different methods (including your own)
    exactly on the same benchmarks, including the same datasets, search space,
    code for training and hyperparameters for that code?
    \answerYes{}
  \item Did you run ablation studies to assess the impact of different
    components of your approach?
    \answerYes{See main.}
  \item Did you use the same evaluation protocol for the methods being compared?
    \answerYes{}
  \item Did you compare performance over time?
    \answerYes{Rank over time}
  \item Did you perform multiple runs of your experiments and report random seeds?
    \answerYes{10 seeds}
  \item Did you report error bars (e.g., with respect to the random seed after
    running experiments multiple times)?
    \answerYes{}
  \item Did you use tabular or surrogate benchmarks for in-depth evaluations?
    \answerYes{}
  \item Did you include the total amount of compute and the type of resources
    used (e.g., type of \textsc{gpu}s, internal cluster, or cloud provider)?
    \answerYes{}
  \item Did you report how you tuned hyperparameters, and what time and
    resources this required (if they were not automatically tuned by your AutoML
    method, e.g. in a \textsc{nas} approach; and also hyperparameters of your
    own method)?
    \answerYes{}
  \end{enumerate}
\item If you are using existing assets (e.g., code, data, models) or
  curating/releasing new assets\dots
  \begin{enumerate}
  \item If your work uses existing assets, did you cite the creators?
    \answerNA{}
  \item Did you mention the license of the assets?
    \answerNA{}
  \item Did you include any new assets either in the supplemental material or as
    a \textsc{url}?
    \answerNA{}
  \item Did you discuss whether and how consent was obtained from people whose
    data you're using/curating?
    \answerNA{}
  \item Did you discuss whether the data you are using/curating contains
    personally identifiable information or offensive content?
    \answerNA{}
  \end{enumerate}
\item If you used crowdsourcing or conducted research with human subjects\dots
  \begin{enumerate}
  \item Did you include the full text of instructions given to participants and
    screenshots, if applicable?
    \answerNA{}
  \item Did you describe any potential participant risks, with links to
    Institutional Review Board (\textsc{irb}) approvals, if applicable?
    \answerNA{}
  \item Did you include the estimated hourly wage paid to participants and the
    total amount spent on participant compensation?
    \answerNA{}
  \end{enumerate}
\end{enumerate}

\begin{acknowledgements}

\end{acknowledgements}





\appendix

\section{Hardware and Runtime}
All experiments are conducted on a CPU cluster with 990 nodes with AMD Milan 7763 CPUs.
The compute time for the BBOB 8d functions was $\SI{45}{\min}$ each so $\SI{14040}{\hour} = \SI{585}{\day}$ in total on CPU (including ablation).
The compute time for the HPOBench was $\SI{90}{\sec}$ each so $\SI{288}{\hour} = \SI{12}{\day}$ in total on CPU.

\begin{rev}
\section{Search Attitude}
\label{sec:search_attitude}

\begin{figure}
    \centering
    \begin{subfigure}[t]{0.6\textwidth}
        \centering
        \includegraphics[width=\textwidth]{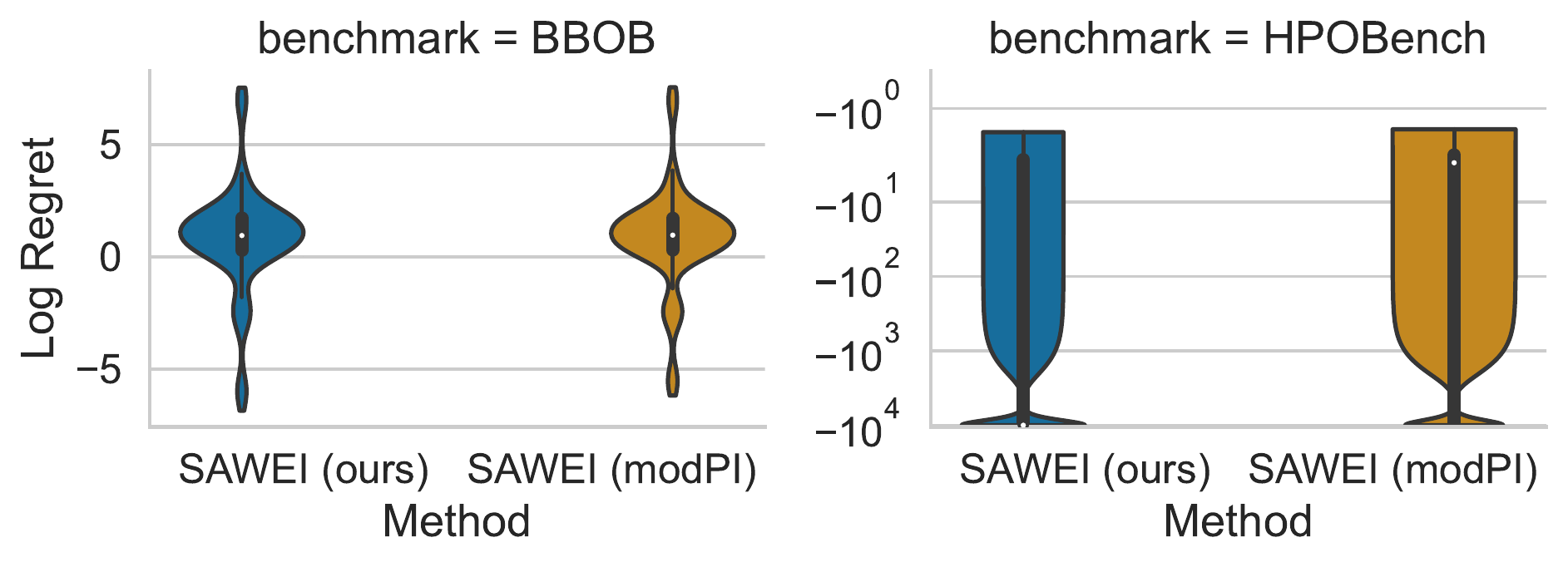}
        \caption{Log Regret}
        \label{fig:log_regret_modPI}
    \end{subfigure}\hfill
    \begin{subfigure}[t]{0.3\textwidth}
        \centering
        \includegraphics[width=\textwidth]{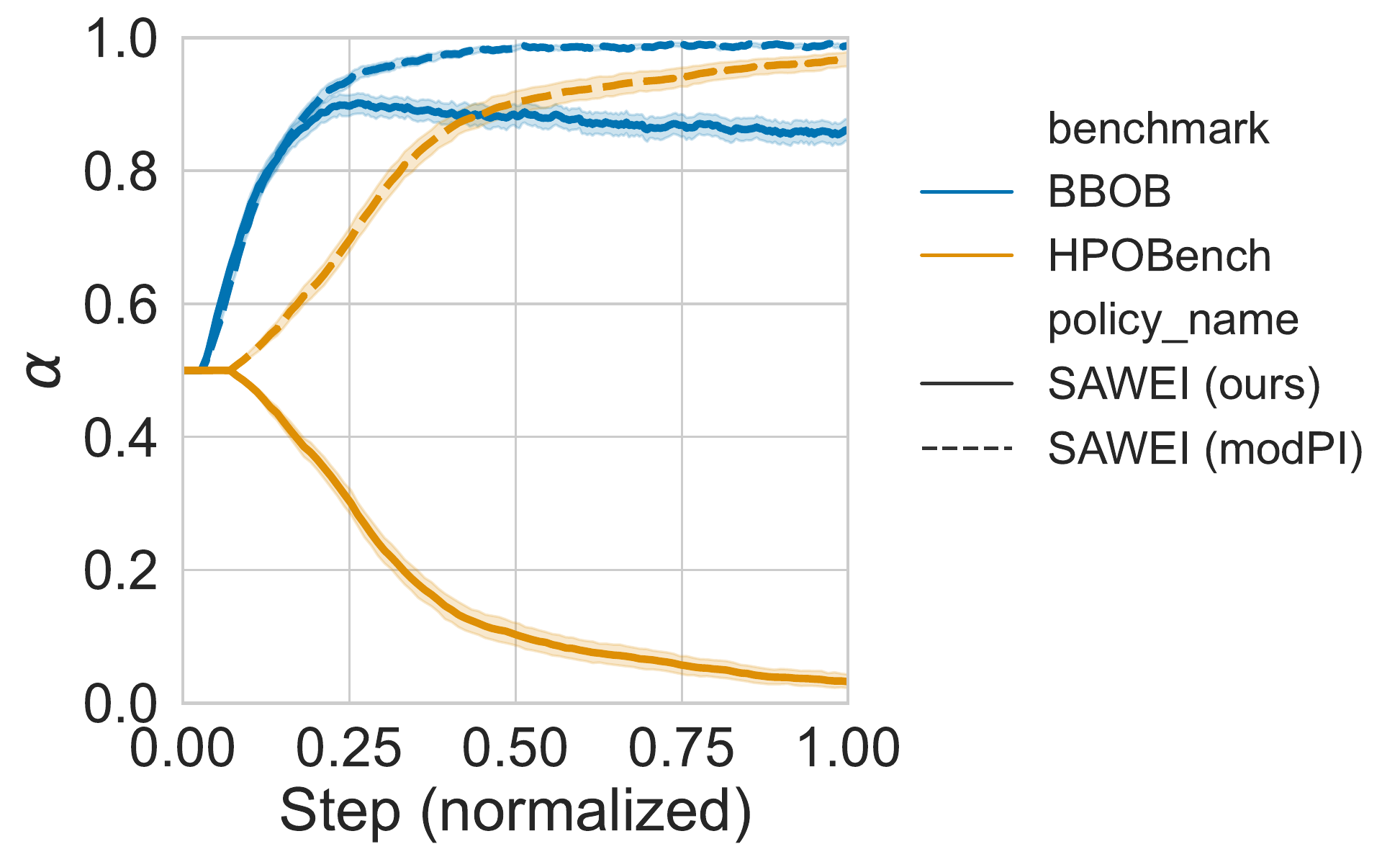}
        \caption{\weiw Trajectory}
        \label{fig:alpha_modPI}
    \end{subfigure}
    \caption{Search Attitude Variation}
    \label{fig:log_regret_mod_vs_pure_PI}
\end{figure}

We determine the search attitude based on the exploration-term of WEI and PI ($\Phi \left[z(\vecx_{\text{next}})\right]$, \cref{sec:how_to_adjust}). 
Originally we compared the exploration-term with the exploitation-term of WEI, the latter being a modified version of PI ($z(\vecx_{\text{next}}) \hat{s}(\vecx_{\text{next}}) \Phi \left[z(\vecx_{\text{next}})\right]$).
We evaluate both versions on BBOB (all 24 functions, 8d, 3 instances, 10 seeds, like in main) and HPOBench (5 models on 8 tasks, 10 seeds, like in main).
In Figure~\cref{fig:log_regret_modPI} on BBOB, we see that the current version (SAWEI (ours)) achieves slightly lower log regret than the one using the modified PI term (SAWEI (modPI)) but otherwise the distributions seem very similar.
On HPOBench, the log regret of SAWEI (modPI) is drastically worse than for SAWEI (ours).
Please note that we denote the optimum log regret of $\log (0)$ by \num{-10000}.
This can be explained by the traversed \weiw, see Figure~\ref{fig:alpha_modPI}.
SAWEI (modPI) adjusts \weiw to exploitation where exploration is required.
In addition, SAWEI (modPI) is not able to reduce \weiw again for BBOB.

\FloatBarrier
\newpage
\section{UBR Intuition}  
\label{sec:ubr_anaysis}
The \acf{UBR} can be used to stop \ac{BO}~\citep{makarova-automl22}.
This means the UBR signalizes whether it is worth to continue optimization.
We add our intuition that this holds for the \textit{current optimizer settings}. This is empirically supported by observing the UBR for the switching policies (EI to PI) where we see sharp bends in the UBR after switching, see~\cref{fig:ubr_switch}.
In our case "current setting" implicitly describes the search attitude whether it is exploring or exploiting.
Therefore we can use the UBR to signal when we should change our search attitude. 

\begin{figure}[h]
    \centering
    \includegraphics[width=0.85\textwidth]{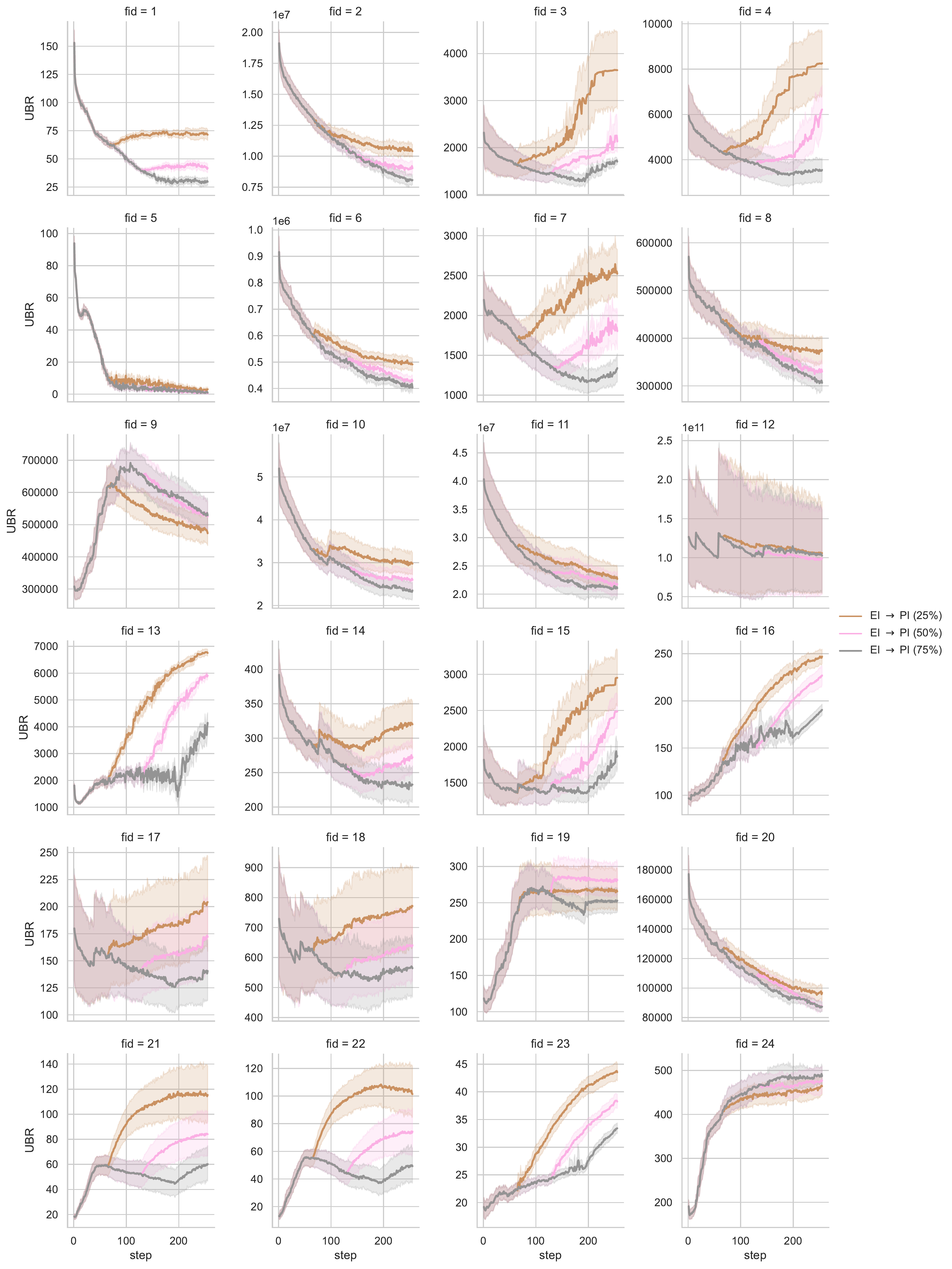}
    \caption{Effect in the \acf{UBR} after changing optimizer settings. Here we switch the acquisition function from EI to PI. BBOB functions, 8d, 10 seeds, 3 instances.}
    \label{fig:ubr_switch}
\end{figure}
\newpage
\end{rev}

\section{BBOB Results}
\label{sec:app_bbob_results}

\begin{figure}[ht]
    \centering
    \includegraphics[width=\textwidth]{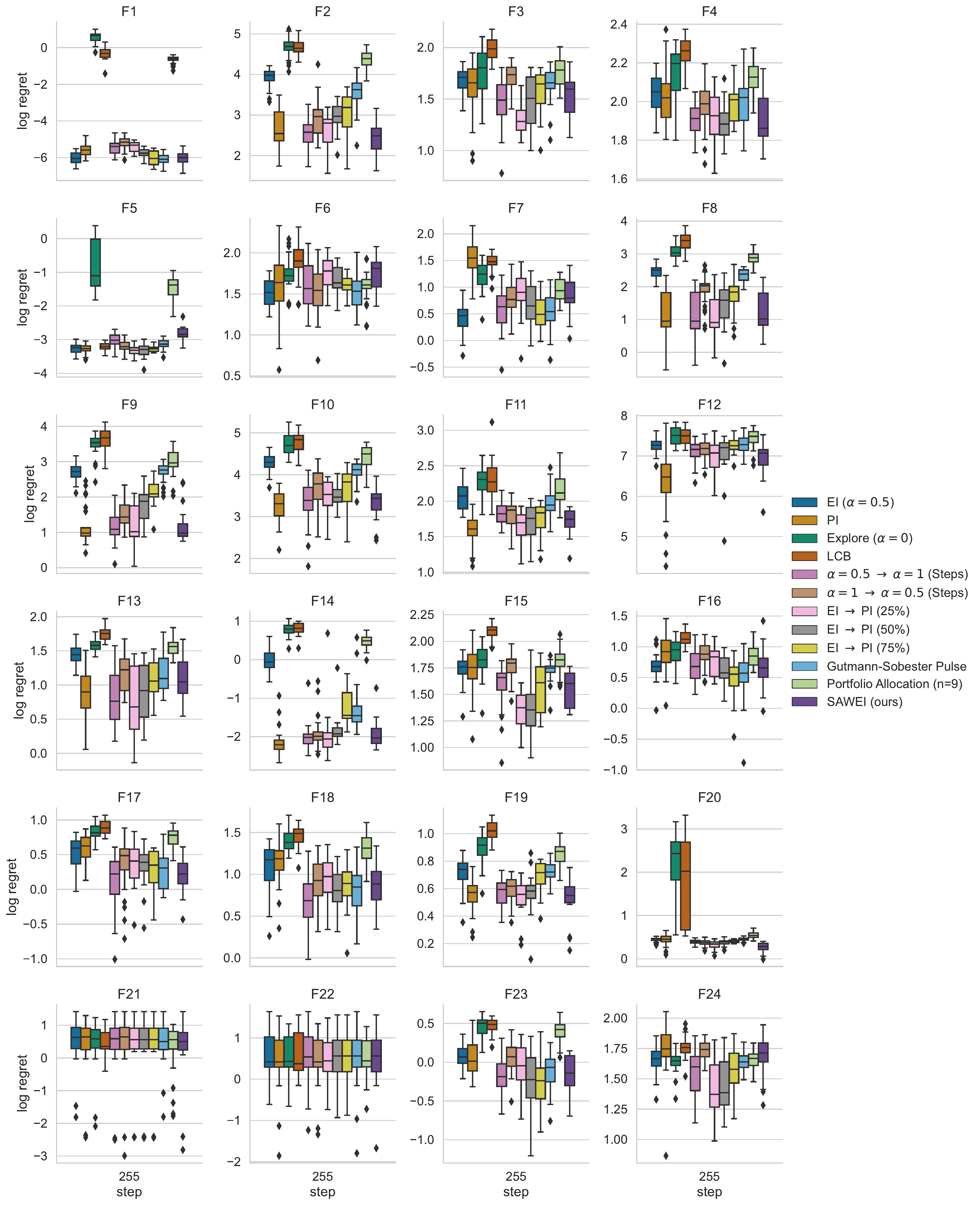}
    \caption{Final log regret on BBOB (8d, 10 seeds, 3 instances)}
    \label{fig:bbob_final_log_regret}
\end{figure}

\label{sec:app_BBOB_schedules}

\begin{figure}[h]
    \centering
    \includegraphics[width=0.85\linewidth]{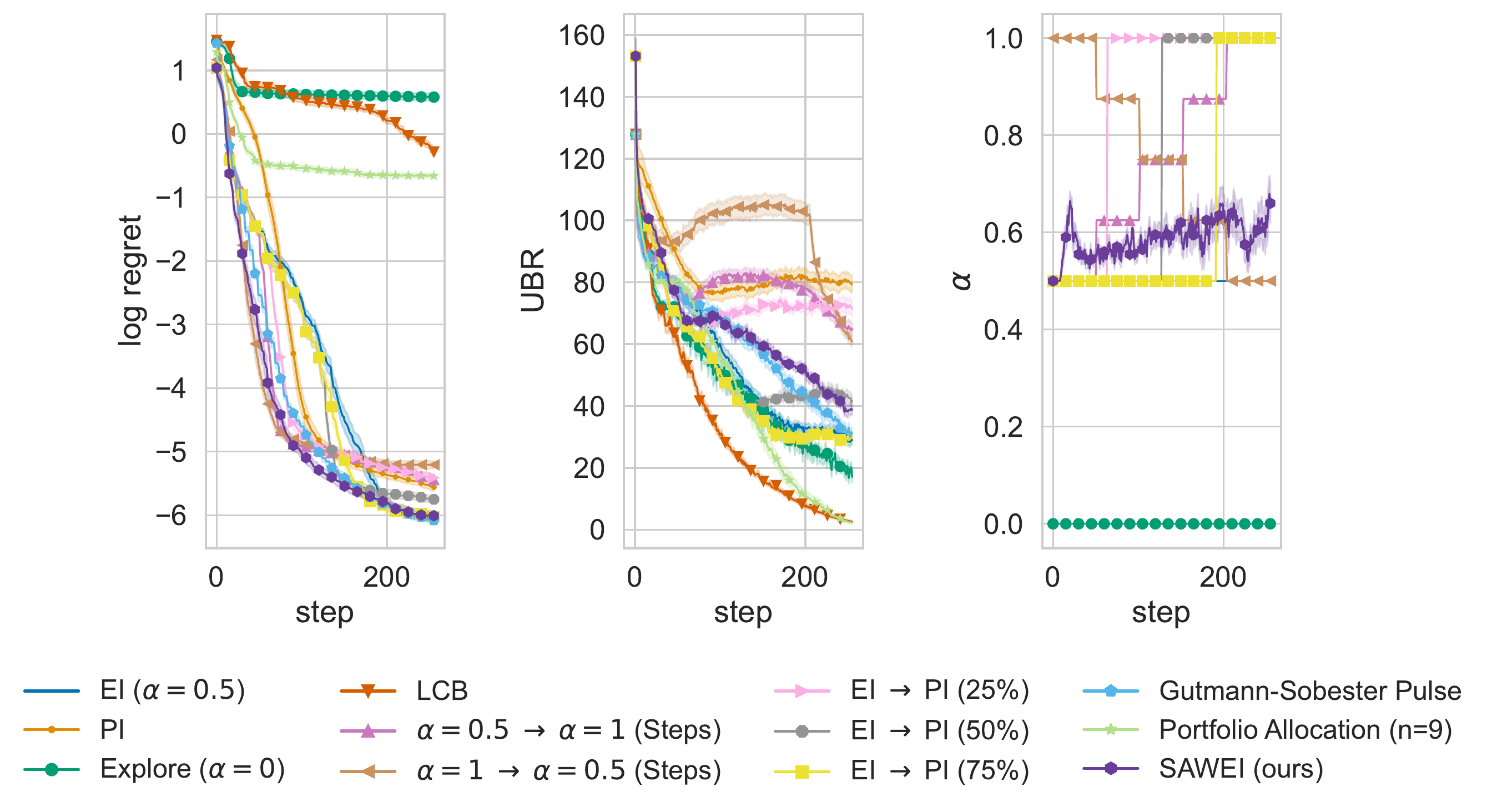}
    \caption{BBOB Function 1}
    \label{fig:figures/BBOB/alpha/001.pdf}
\end{figure}

\begin{figure}[h]
    \centering
    \includegraphics[width=0.85\linewidth]{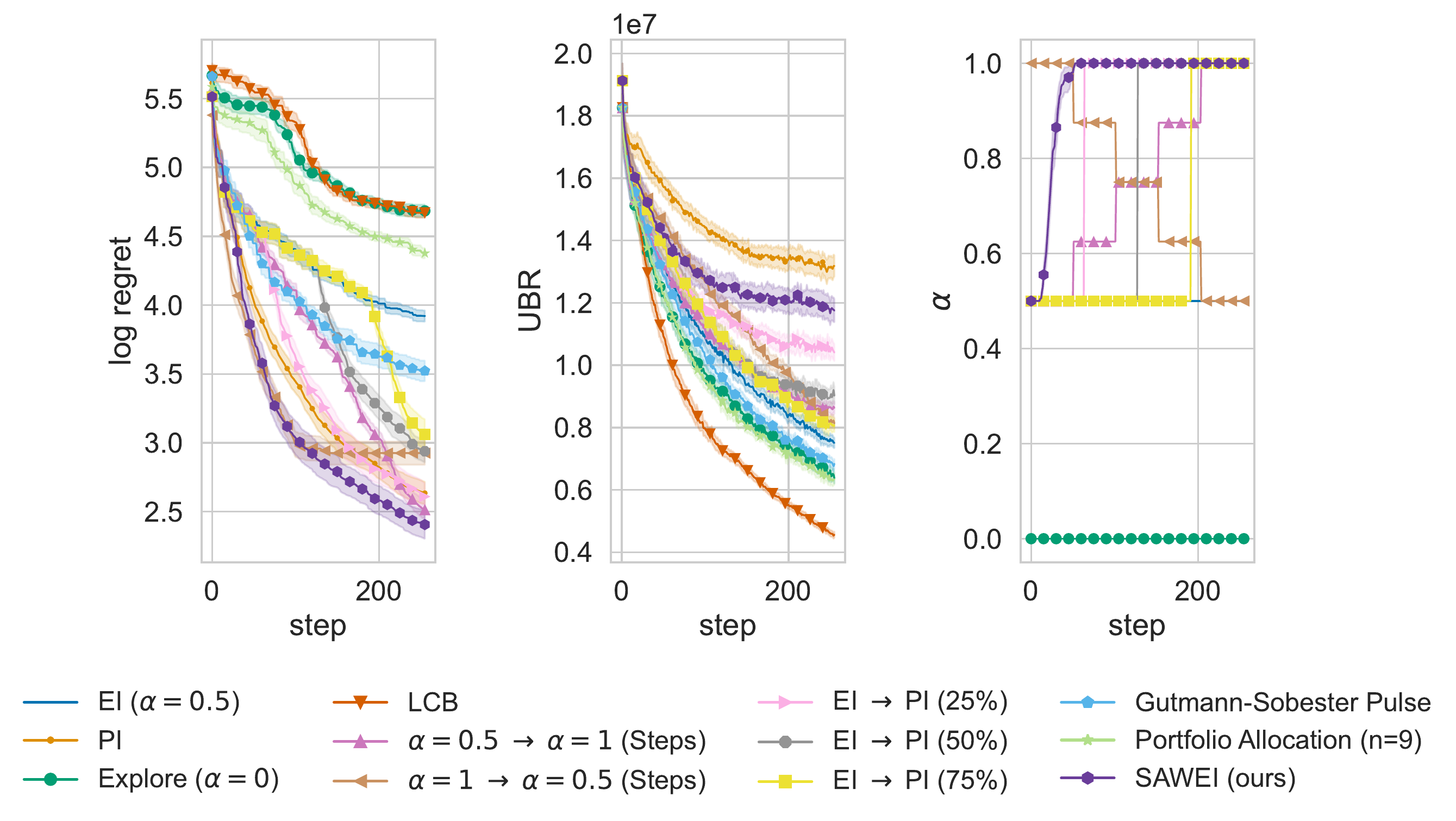}
    \caption{BBOB Function 2}
    \label{fig:figures/BBOB/alpha/002.pdf}
\end{figure}

\begin{figure}[h]
    \centering
    \includegraphics[width=0.85\linewidth]{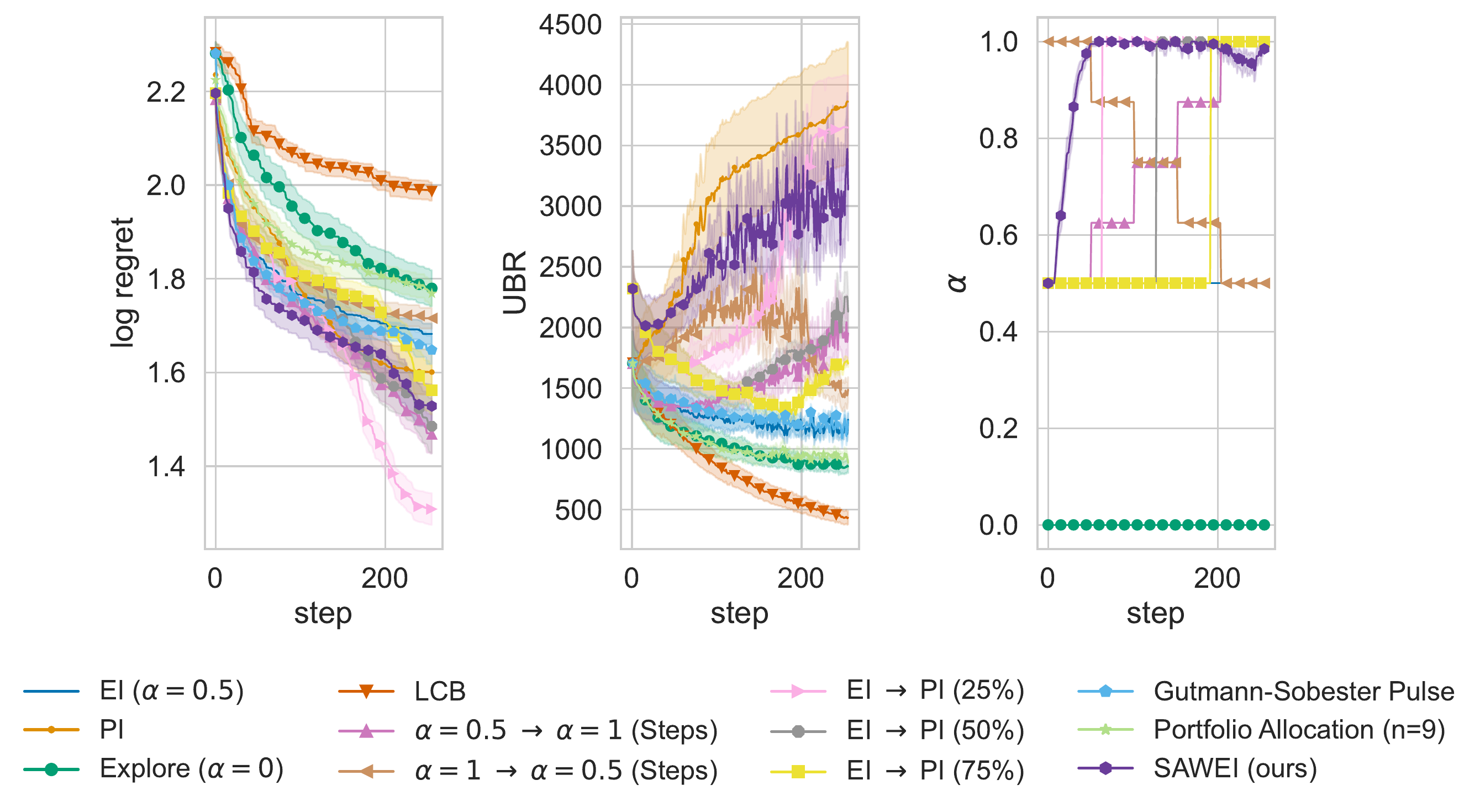}
    \caption{BBOB Function 3}
    \label{fig:figures/BBOB/alpha/003.pdf}
\end{figure}

\begin{figure}[h]
    \centering
    \includegraphics[width=0.85\linewidth]{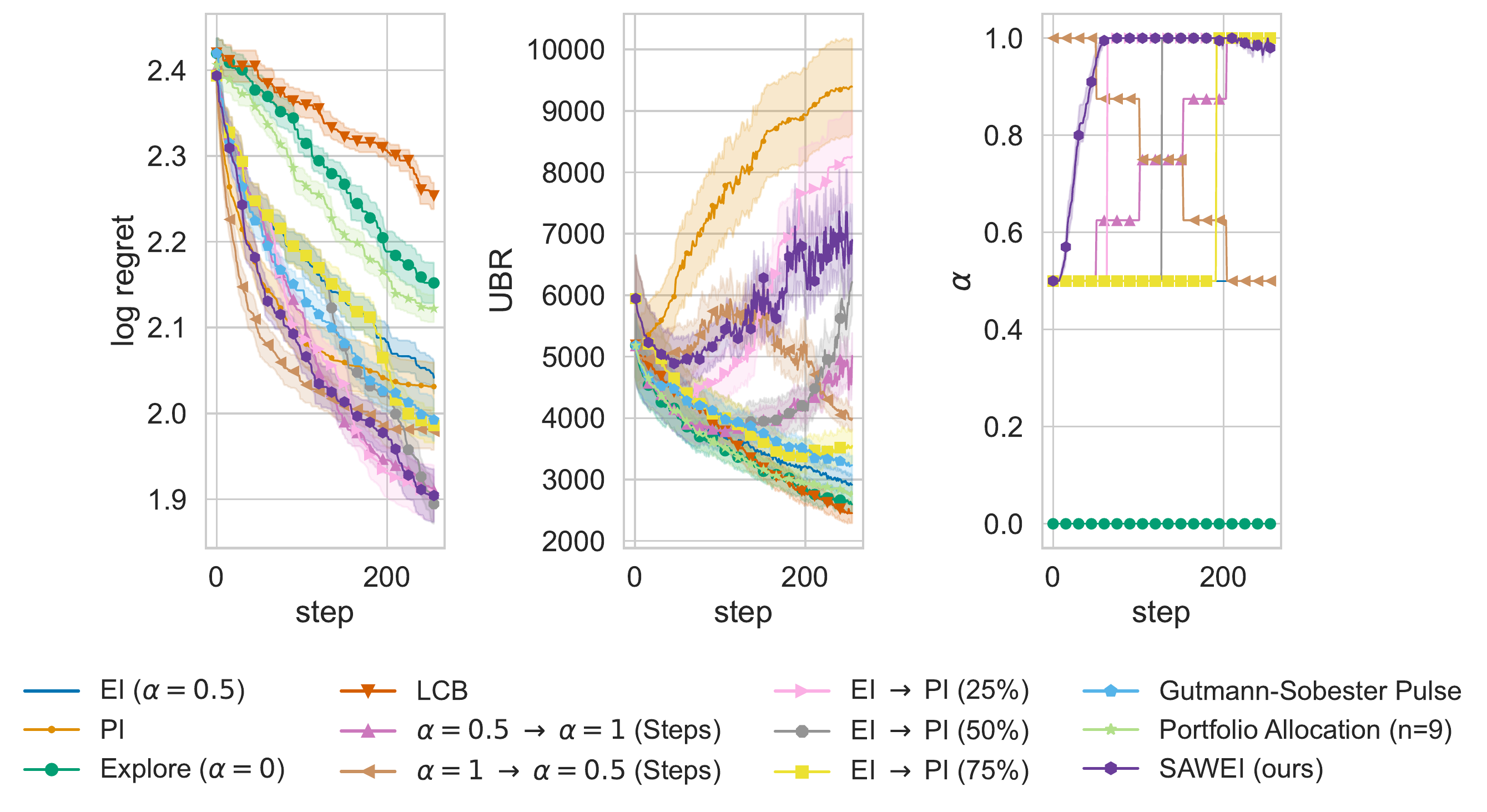}
    \caption{BBOB Function 4}
    \label{fig:figures/BBOB/alpha/004.pdf}
\end{figure}

\begin{figure}[h]
    \centering
    \includegraphics[width=0.85\linewidth]{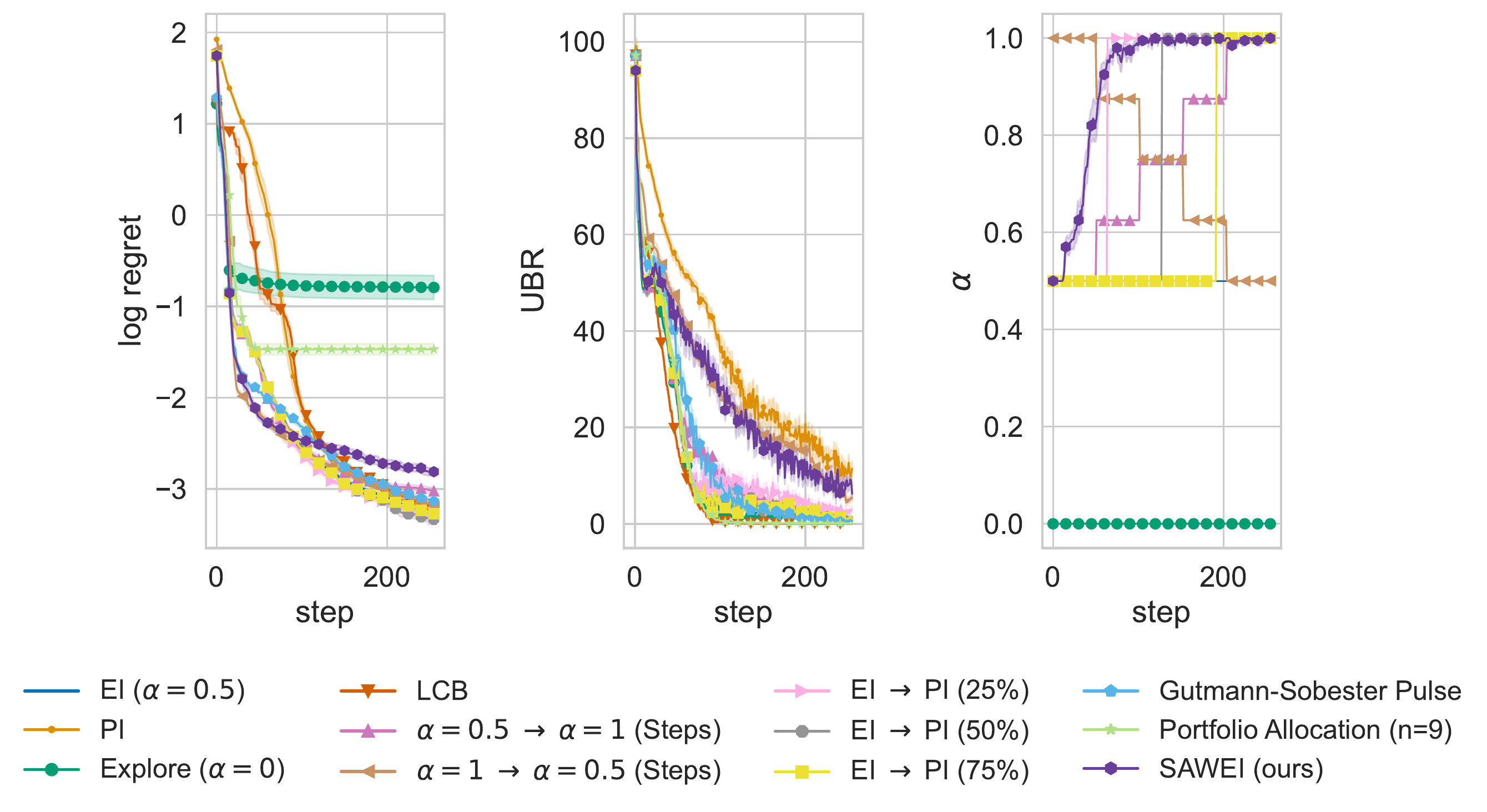}
    \caption{BBOB Function 5}
    \label{fig:figures/BBOB/alpha/005.pdf}
\end{figure}

\begin{figure}[h]
    \centering
    \includegraphics[width=0.85\linewidth]{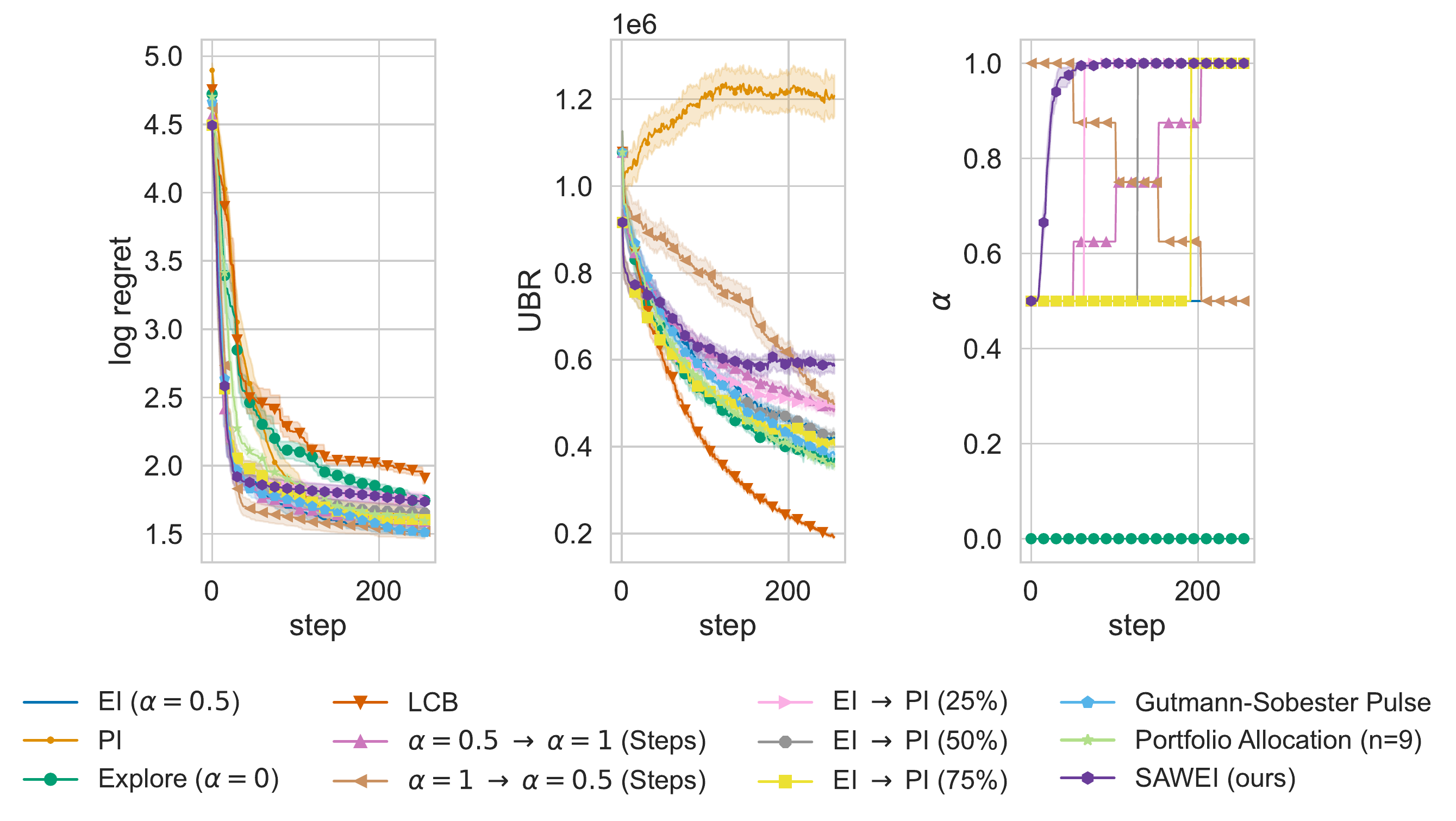}
    \caption{BBOB Function 6}
    \label{fig:figures/BBOB/alpha/006.pdf}
\end{figure}

\begin{figure}[h]
    \centering
    \includegraphics[width=0.85\linewidth]{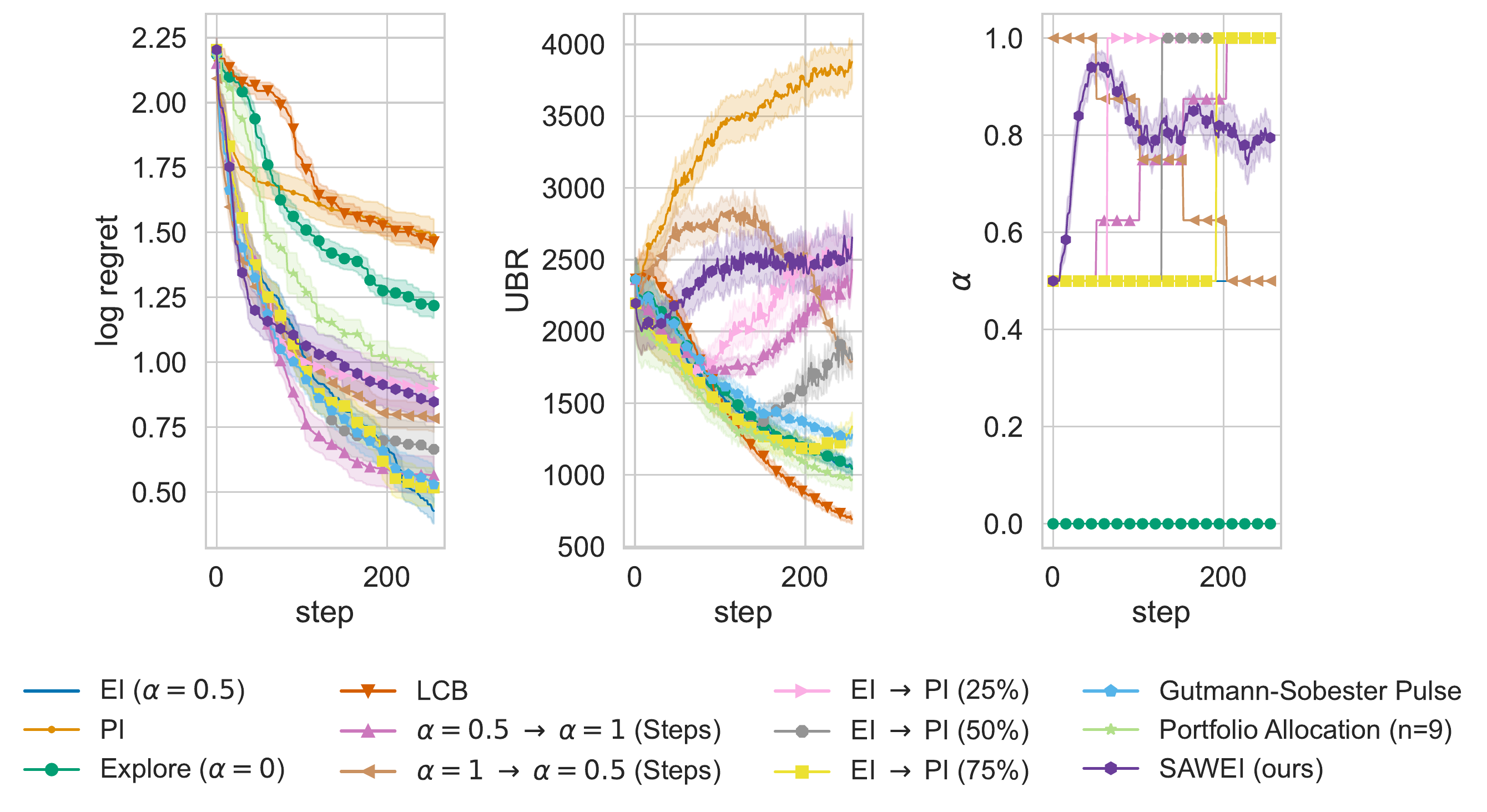}
    \caption{BBOB Function 7}
    \label{fig:figures/BBOB/alpha/007.pdf}
\end{figure}

\begin{figure}[h]
    \centering
    \includegraphics[width=0.85\linewidth]{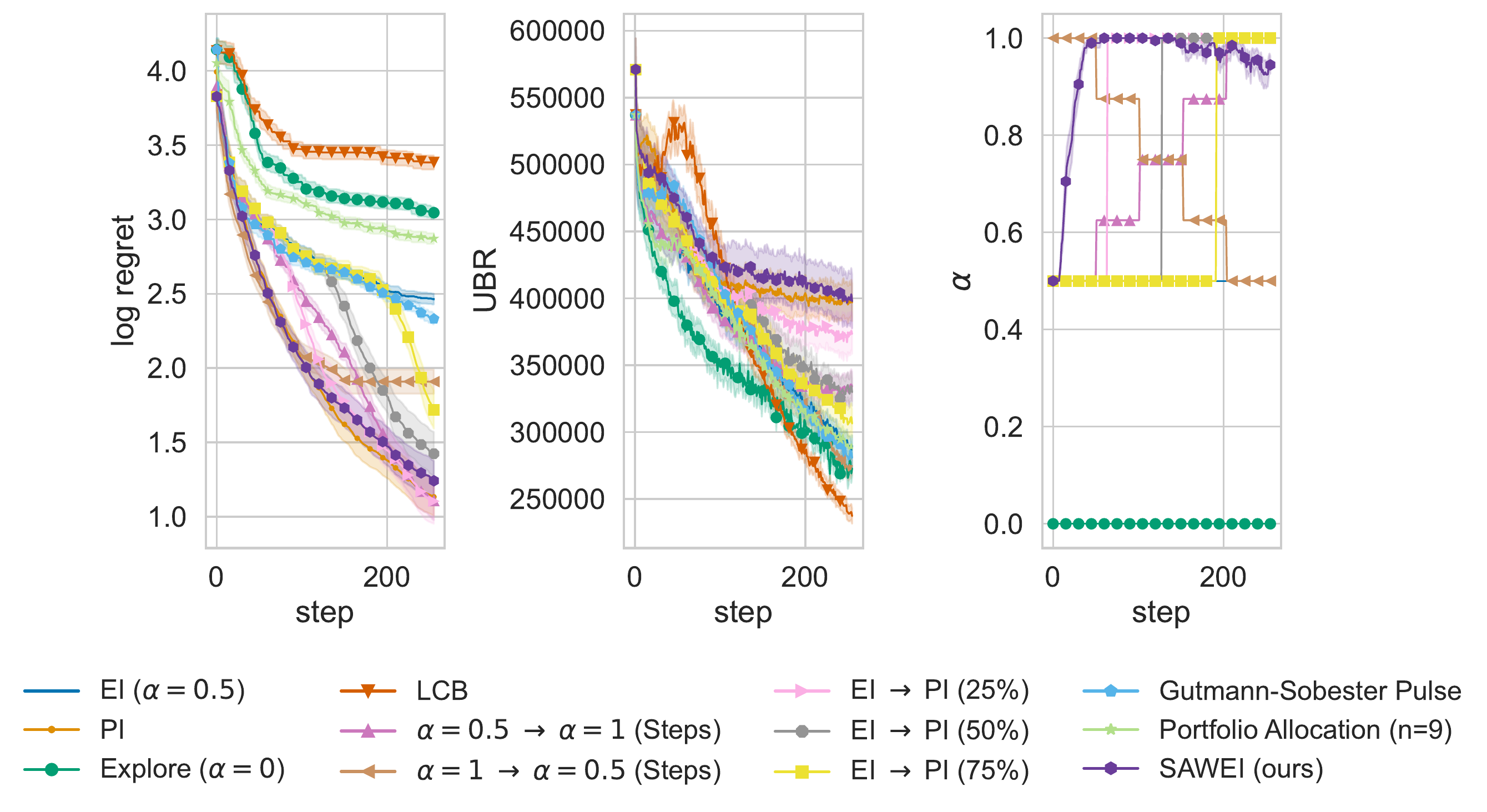}
    \caption{BBOB Function 8}
    \label{fig:figures/BBOB/alpha/008.pdf}
\end{figure}

\begin{figure}[h]
    \centering
    \includegraphics[width=0.85\linewidth]{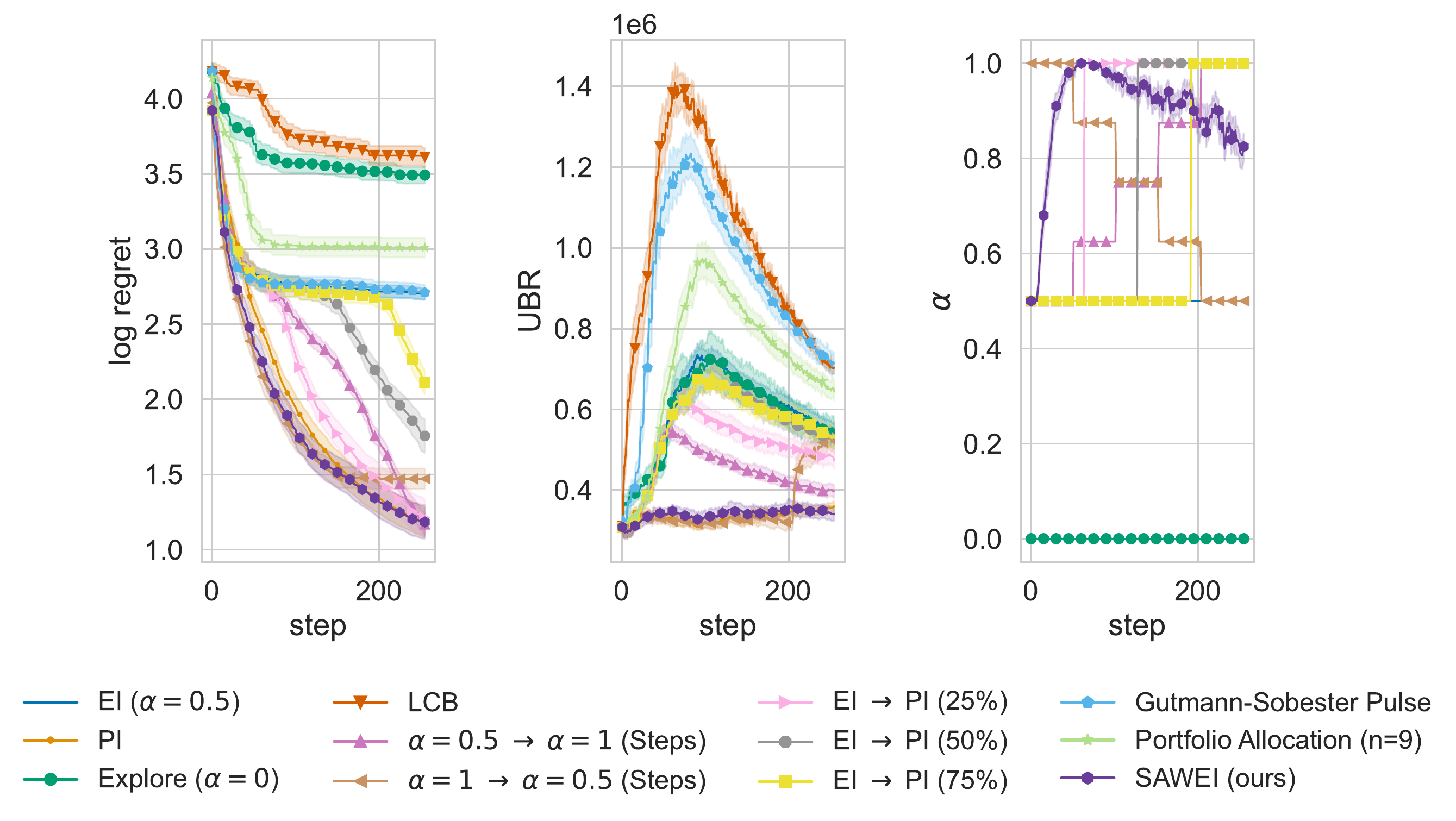}
    \caption{BBOB Function 9}
    \label{fig:figures/BBOB/alpha/009.pdf}
\end{figure}

\begin{figure}[h]
    \centering
    \includegraphics[width=0.85\linewidth]{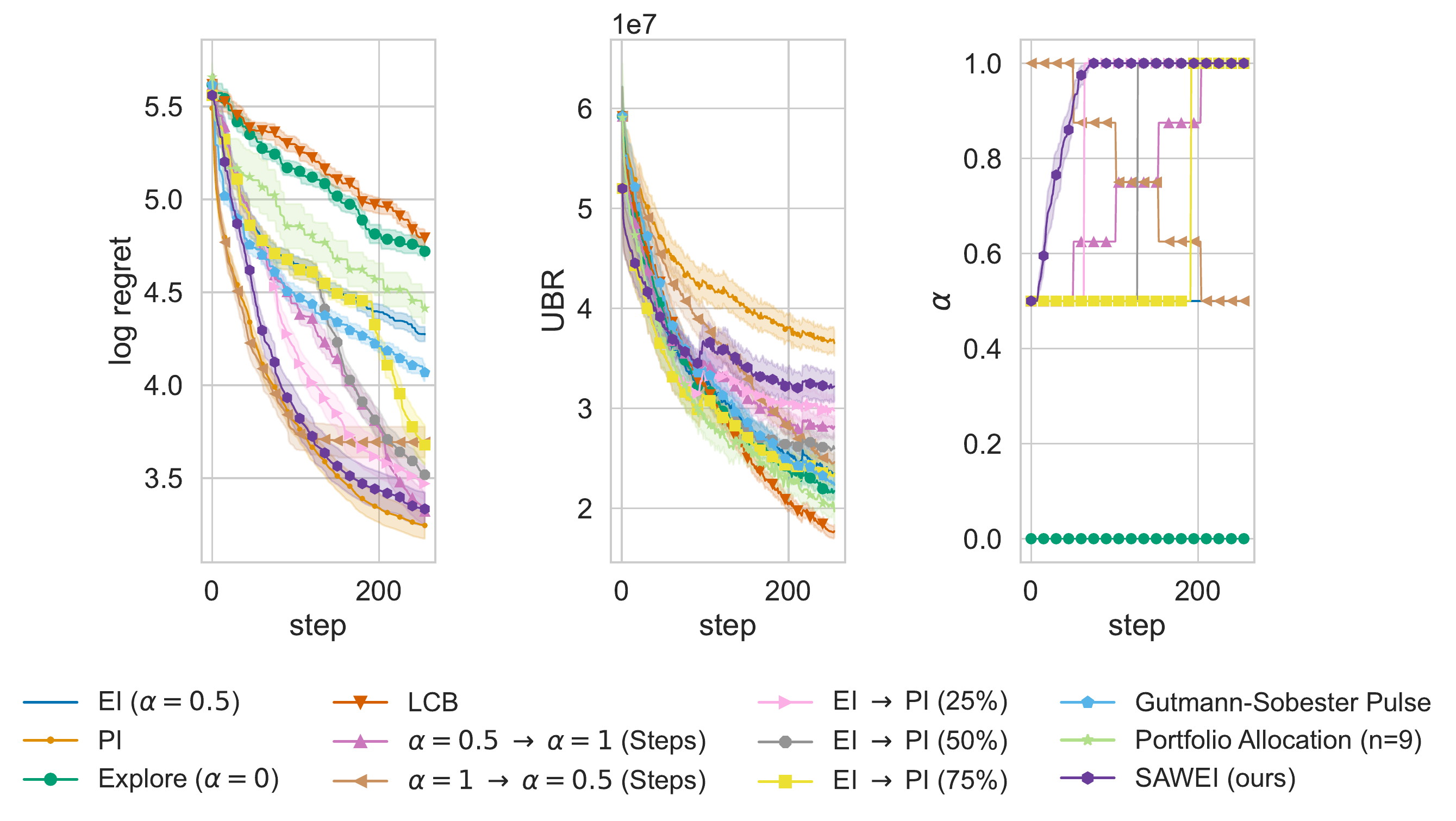}
    \caption{BBOB Function 10}
    \label{fig:figures/BBOB/alpha/010.pdf}
\end{figure}

\begin{figure}[h]
    \centering
    \includegraphics[width=0.85\linewidth]{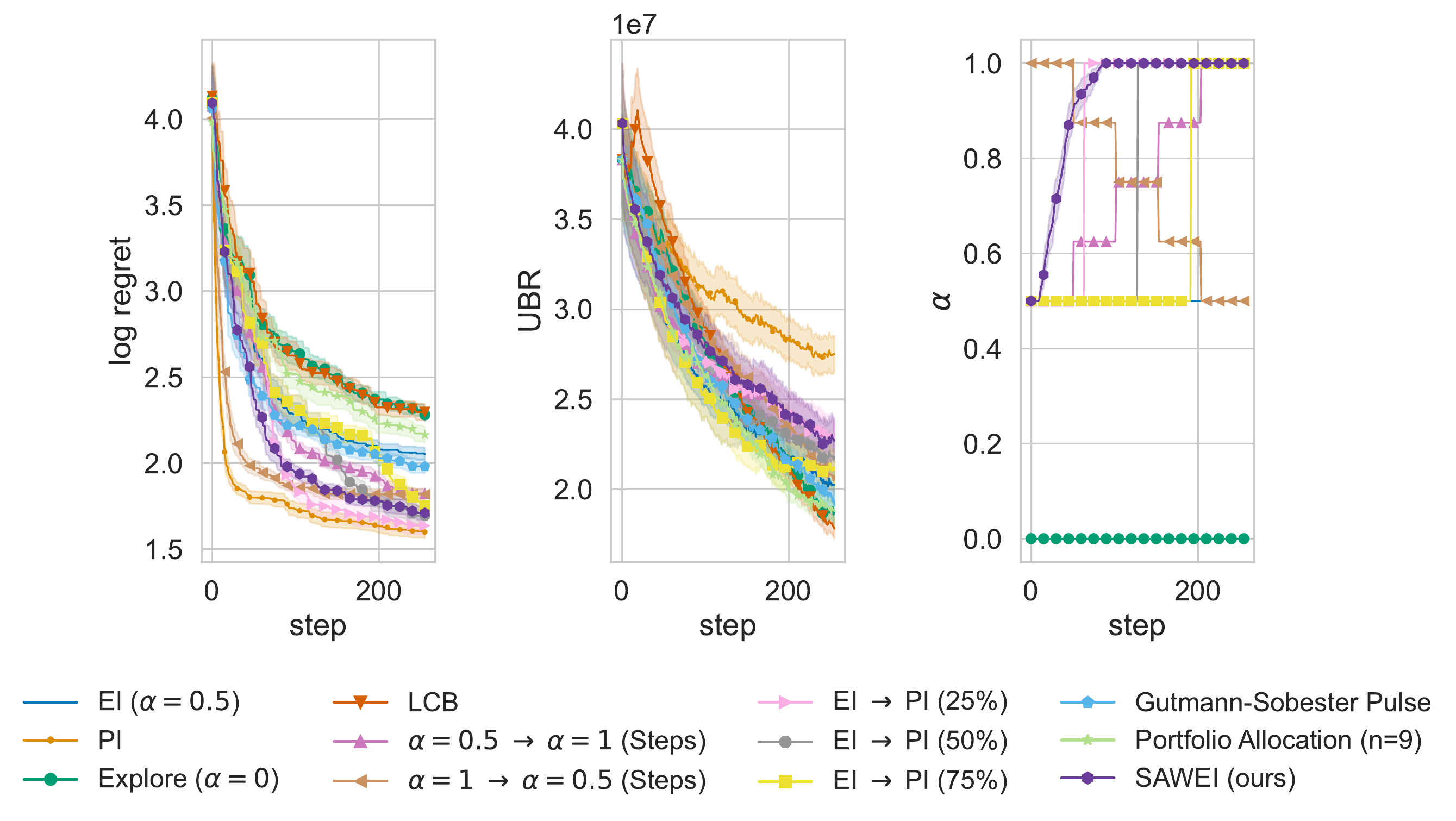}
    \caption{BBOB Function 11}
    \label{fig:figures/BBOB/alpha/011.pdf}
\end{figure}

\begin{figure}[h]
    \centering
    \includegraphics[width=0.85\linewidth]{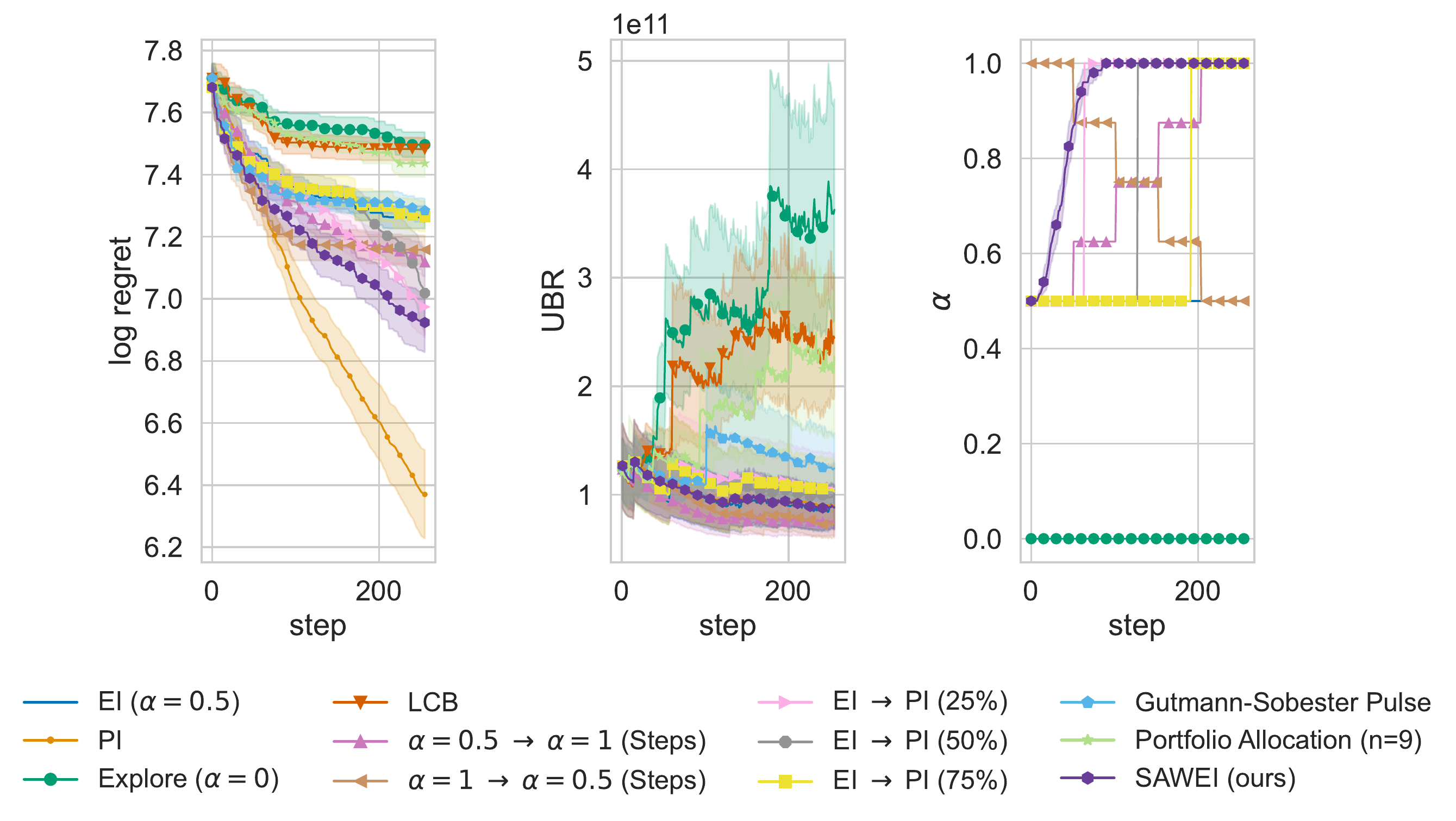}
    \caption{BBOB Function 12}
    \label{fig:figures/BBOB/alpha/012.pdf}
\end{figure}

\begin{figure}[h]
    \centering
    \includegraphics[width=0.85\linewidth]{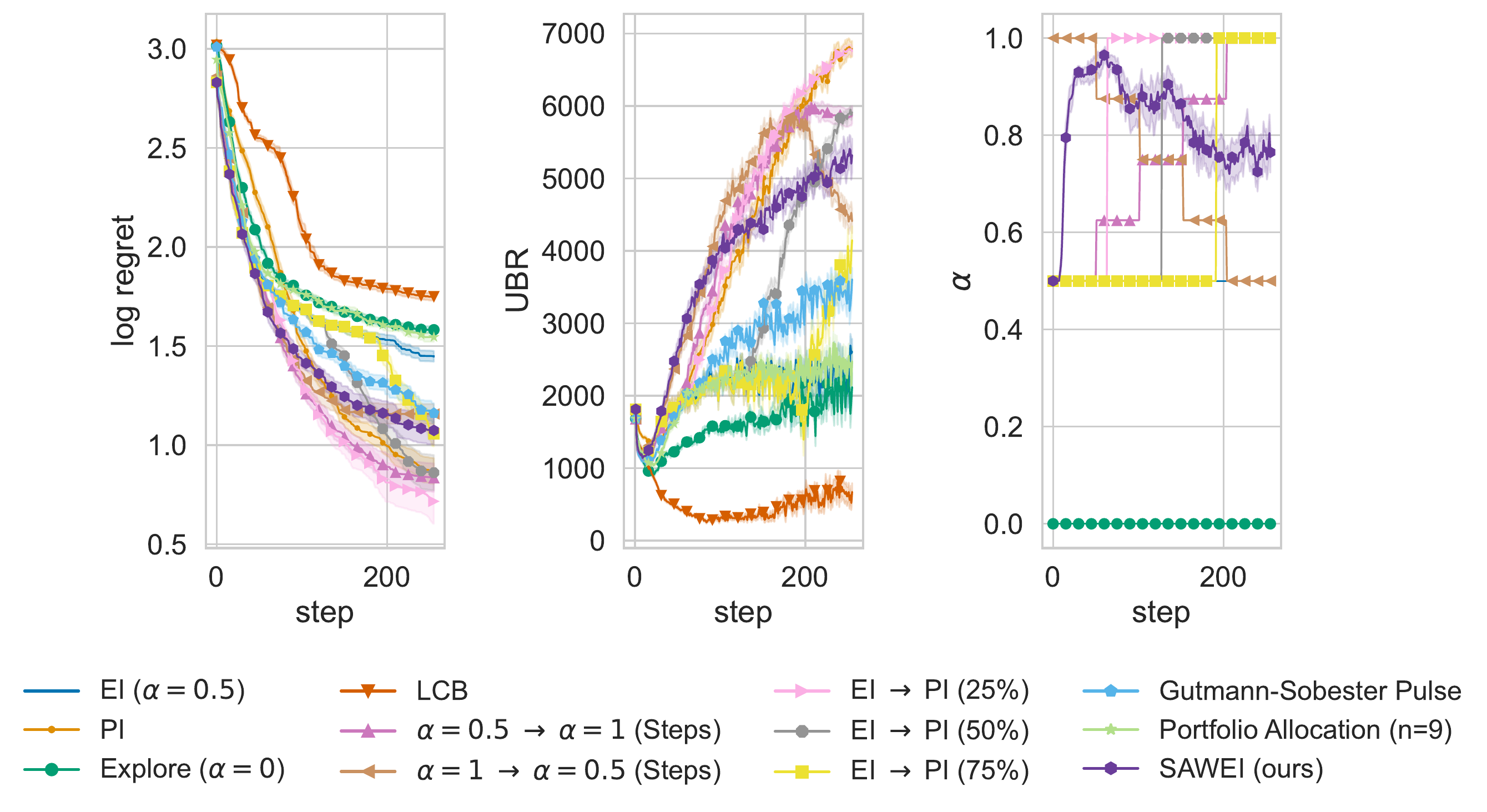}
    \caption{BBOB Function 13}
    \label{fig:figures/BBOB/alpha/013.pdf}
\end{figure}

\begin{figure}[h]
    \centering
    \includegraphics[width=0.85\linewidth]{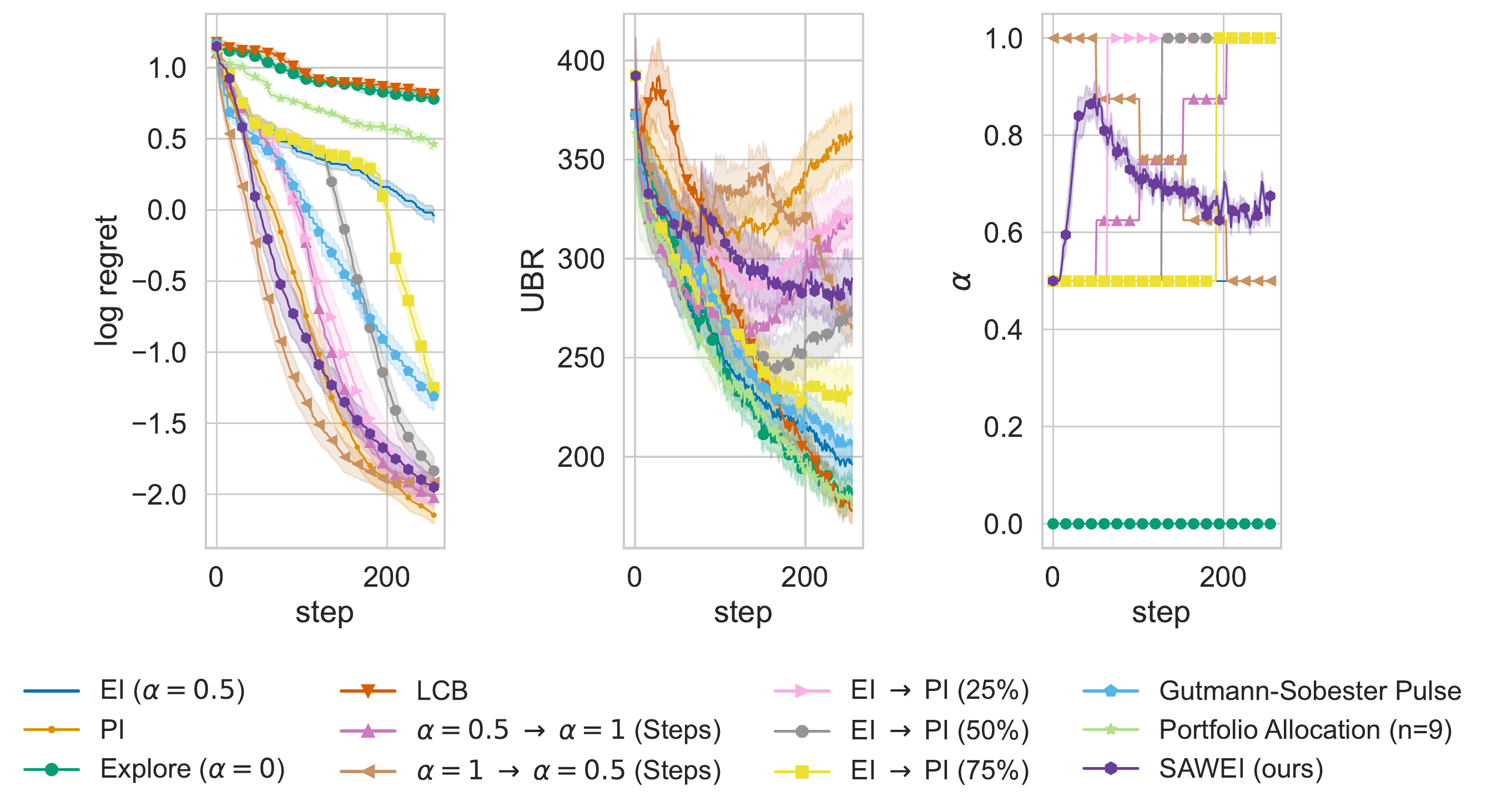}
    \caption{BBOB Function 14}
    \label{fig:figures/BBOB/alpha/014.pdf}
\end{figure}

\begin{figure}[h]
    \centering
    \includegraphics[width=0.85\linewidth]{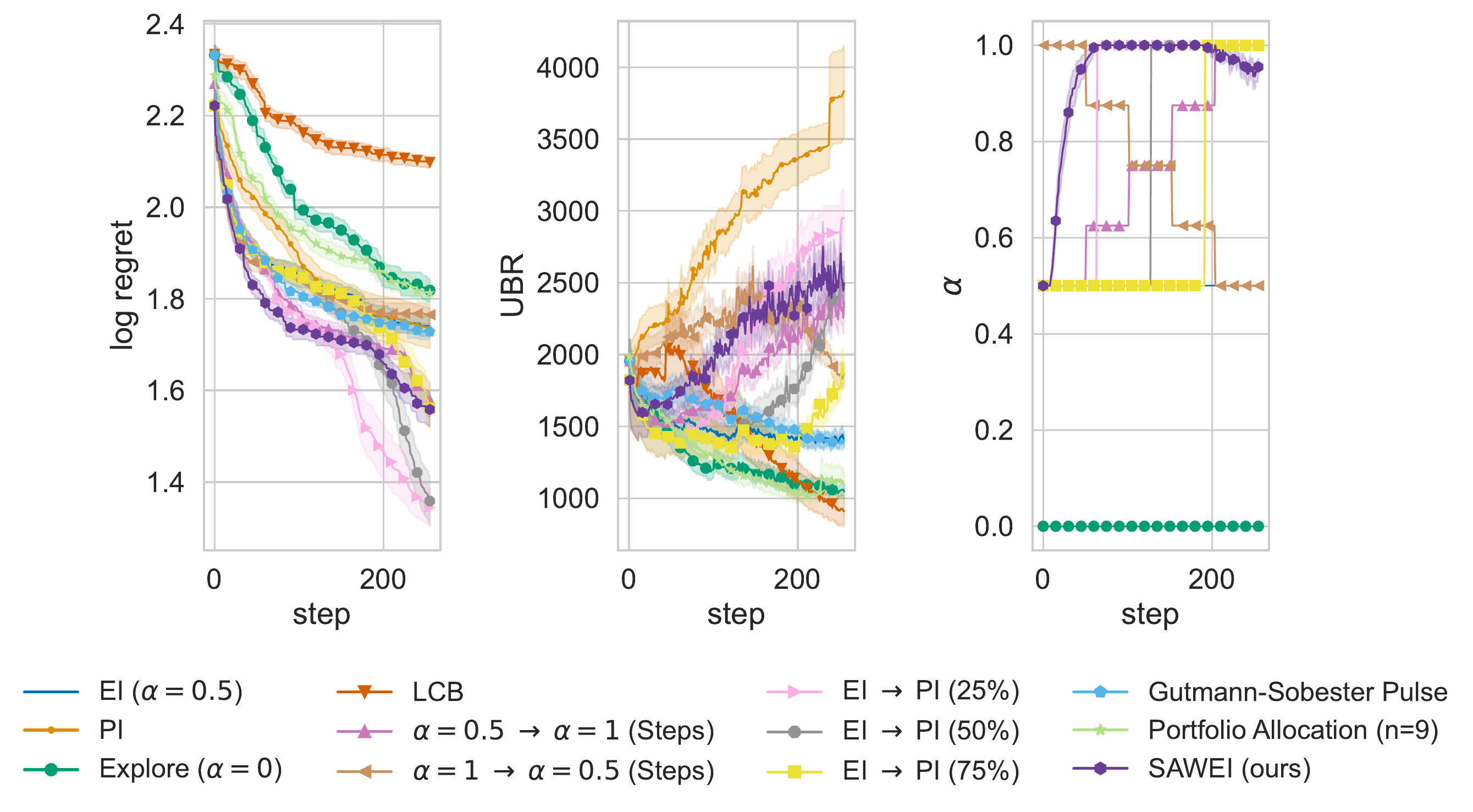}
    \caption{BBOB Function 15}
    \label{fig:figures/BBOB/alpha/015.pdf}
\end{figure}

\begin{figure}[h]
    \centering
    \includegraphics[width=0.85\linewidth]{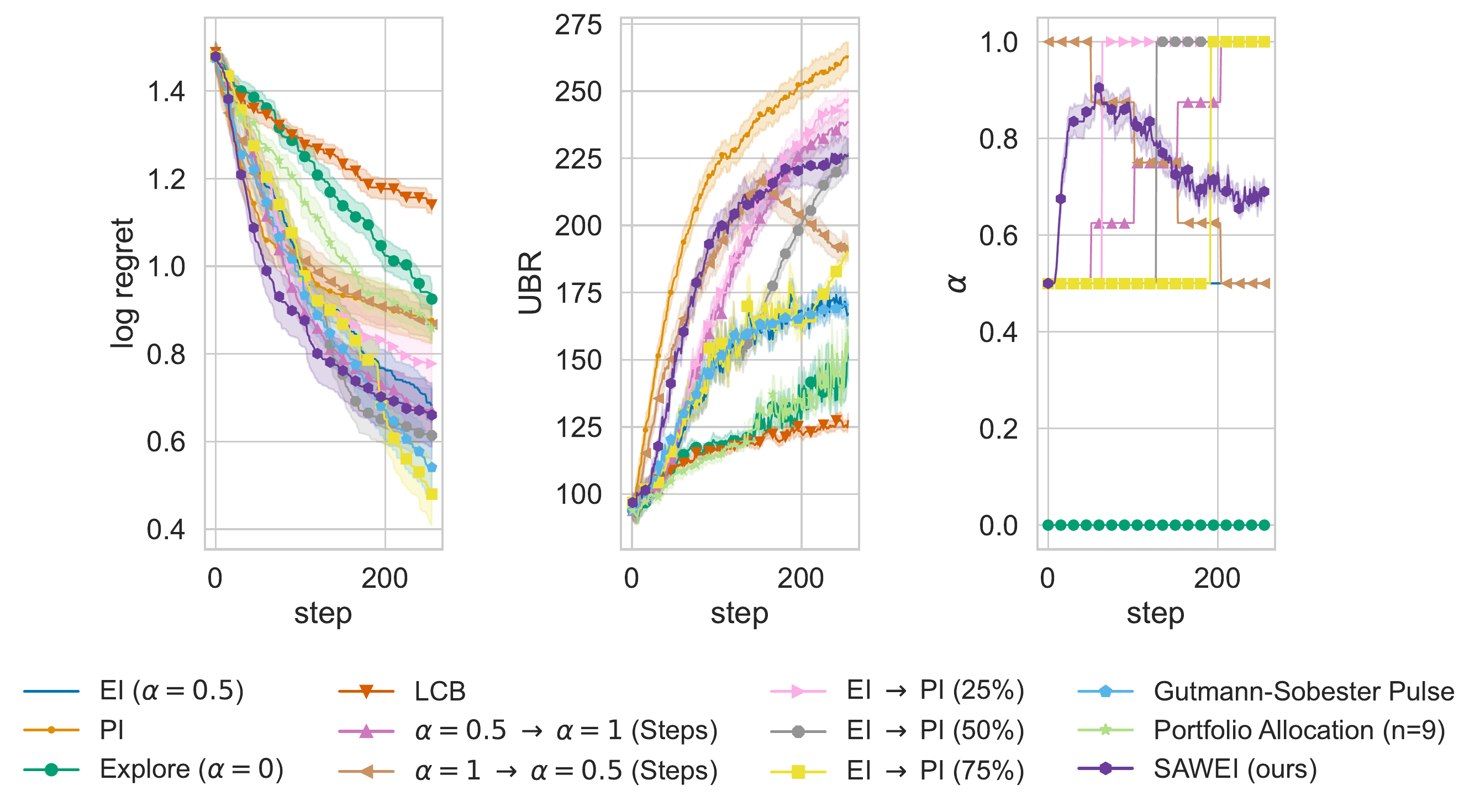}
    \caption{BBOB Function 16}
    \label{fig:figures/BBOB/alpha/016.pdf}
\end{figure}

\begin{figure}[h]
    \centering
    \includegraphics[width=0.85\linewidth]{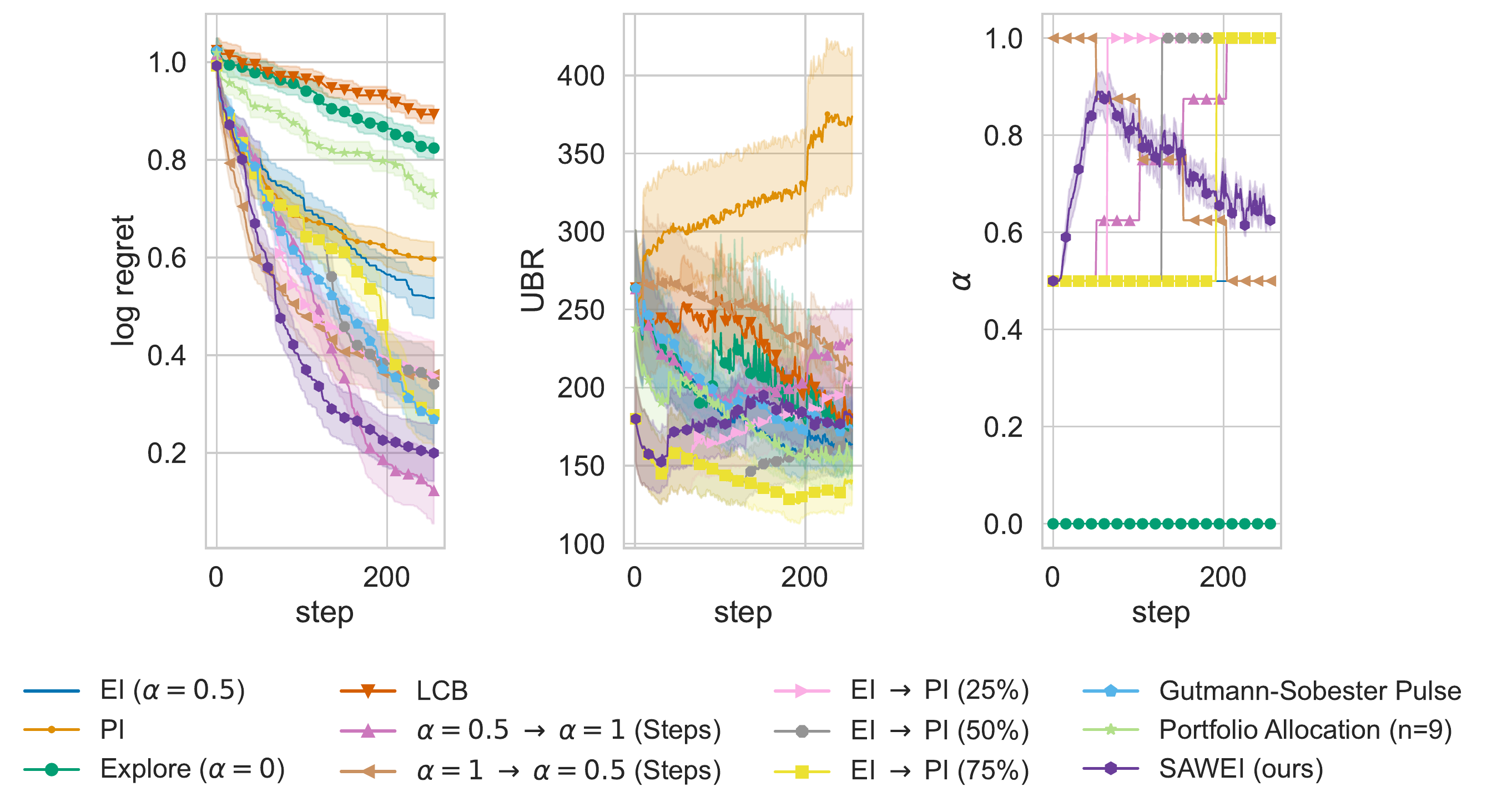}
    \caption{BBOB Function 17}
    \label{fig:figures/BBOB/alpha/017.pdf}
\end{figure}

\begin{figure}[h]
    \centering
    \includegraphics[width=0.85\linewidth]{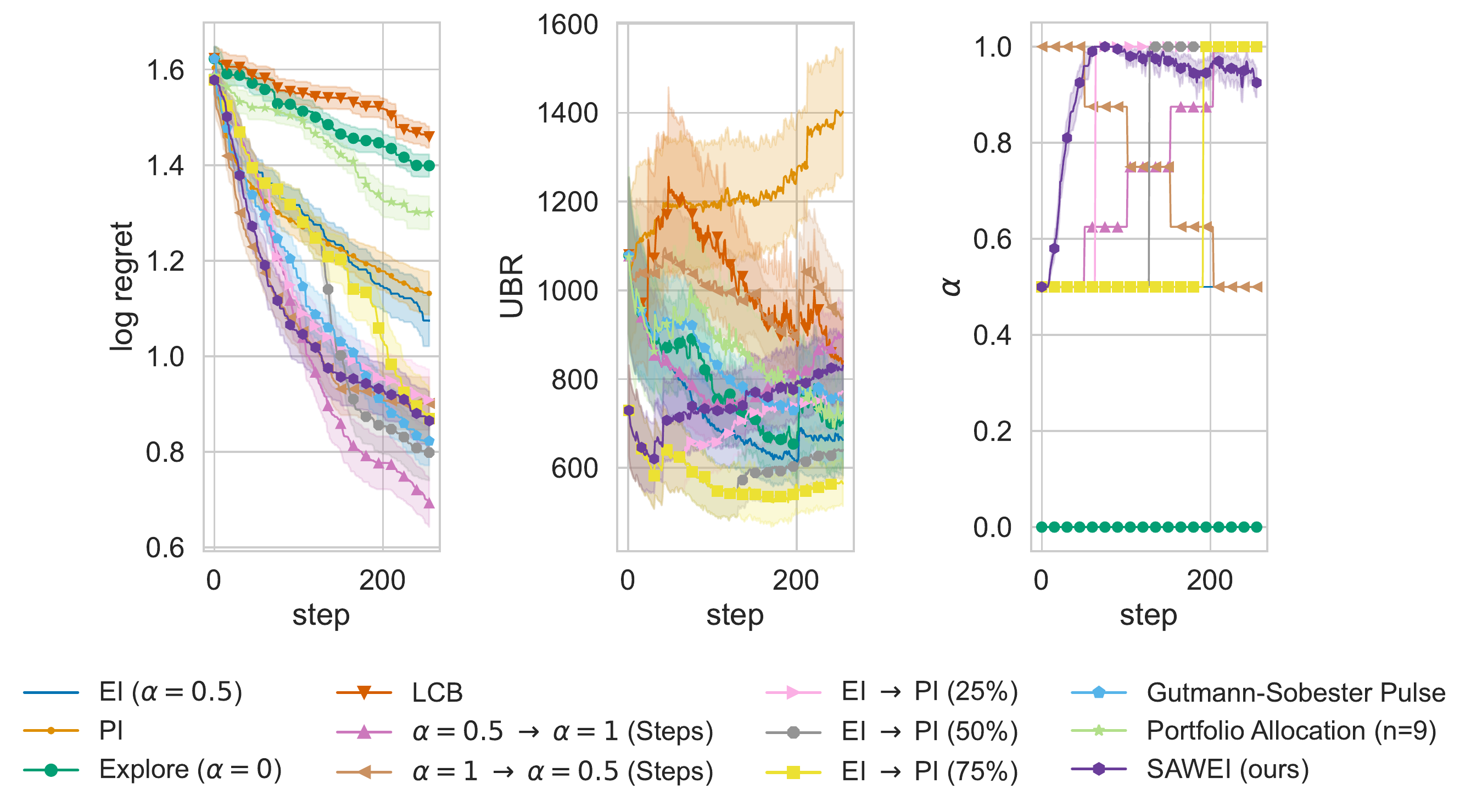}
    \caption{BBOB Function 18}
    \label{fig:figures/BBOB/alpha/018.pdf}
\end{figure}

\begin{figure}[h]
    \centering
    \includegraphics[width=0.85\linewidth]{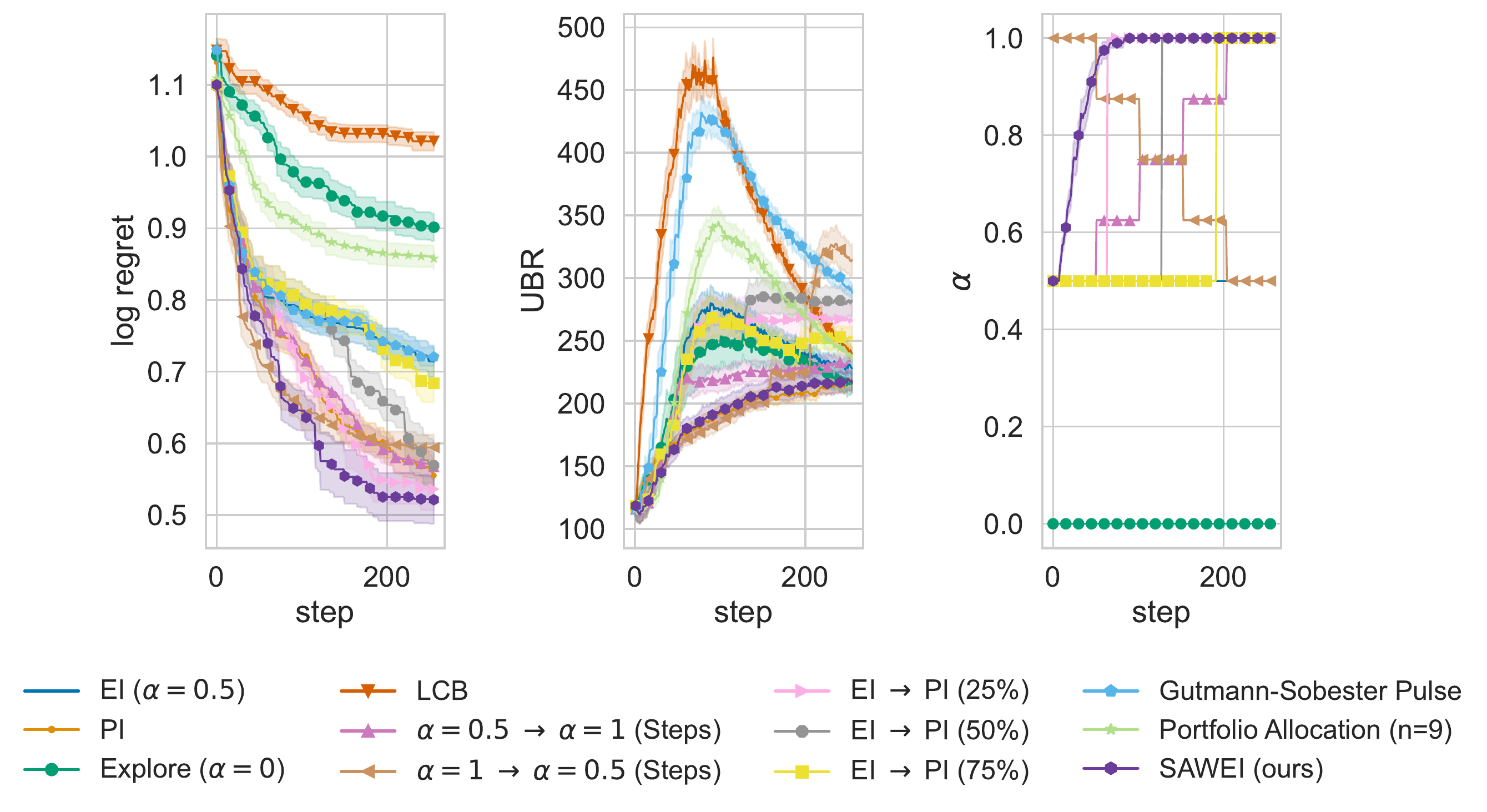}
    \caption{BBOB Function 19}
    \label{fig:figures/BBOB/alpha/019.pdf}
\end{figure}

\begin{figure}[h]
    \centering
    \includegraphics[width=0.85\linewidth]{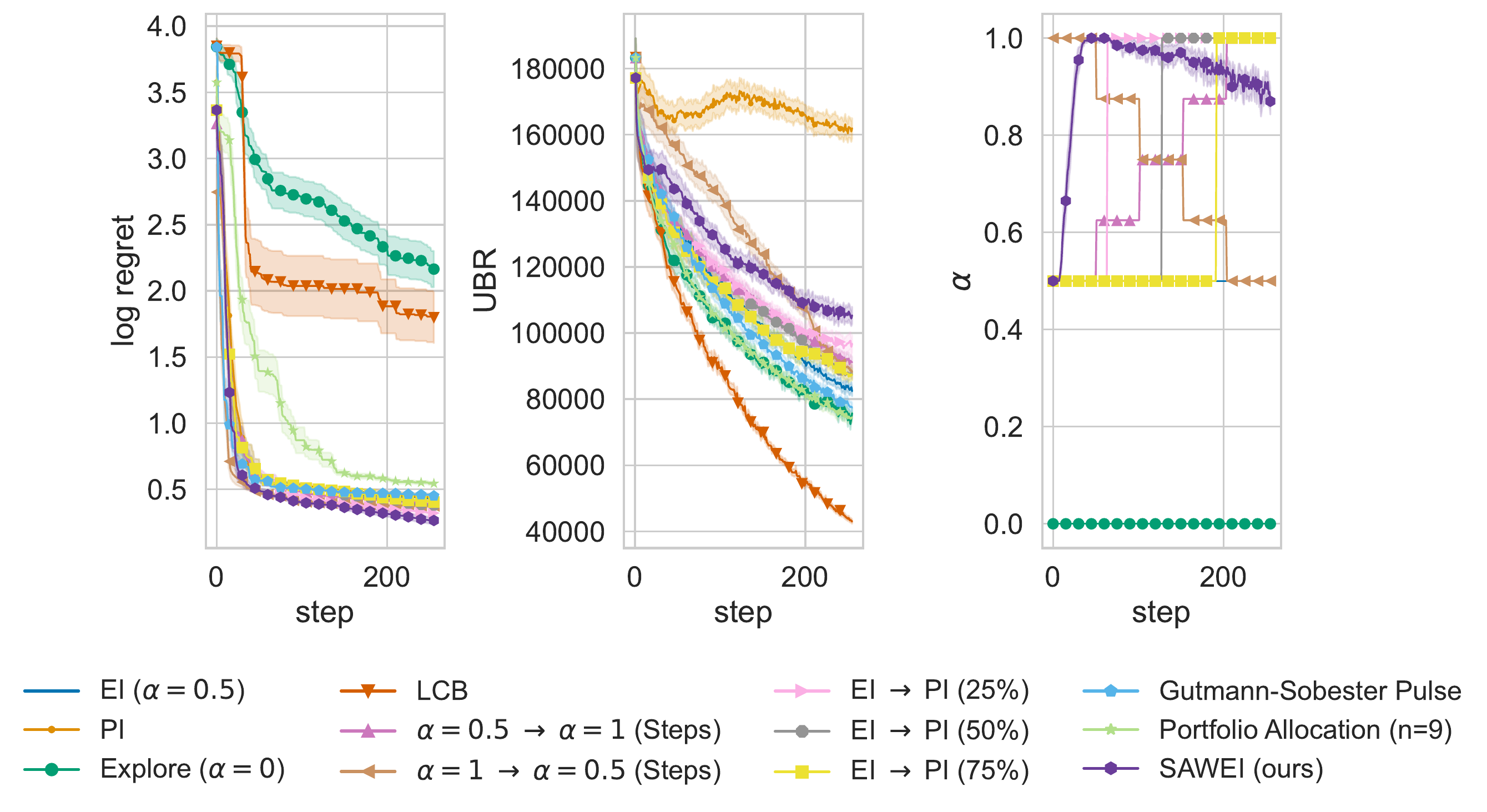}
    \caption{BBOB Function 20}
    \label{fig:figures/BBOB/alpha/020.pdf}
\end{figure}

\begin{figure}[h]
    \centering
    \includegraphics[width=0.85\linewidth]{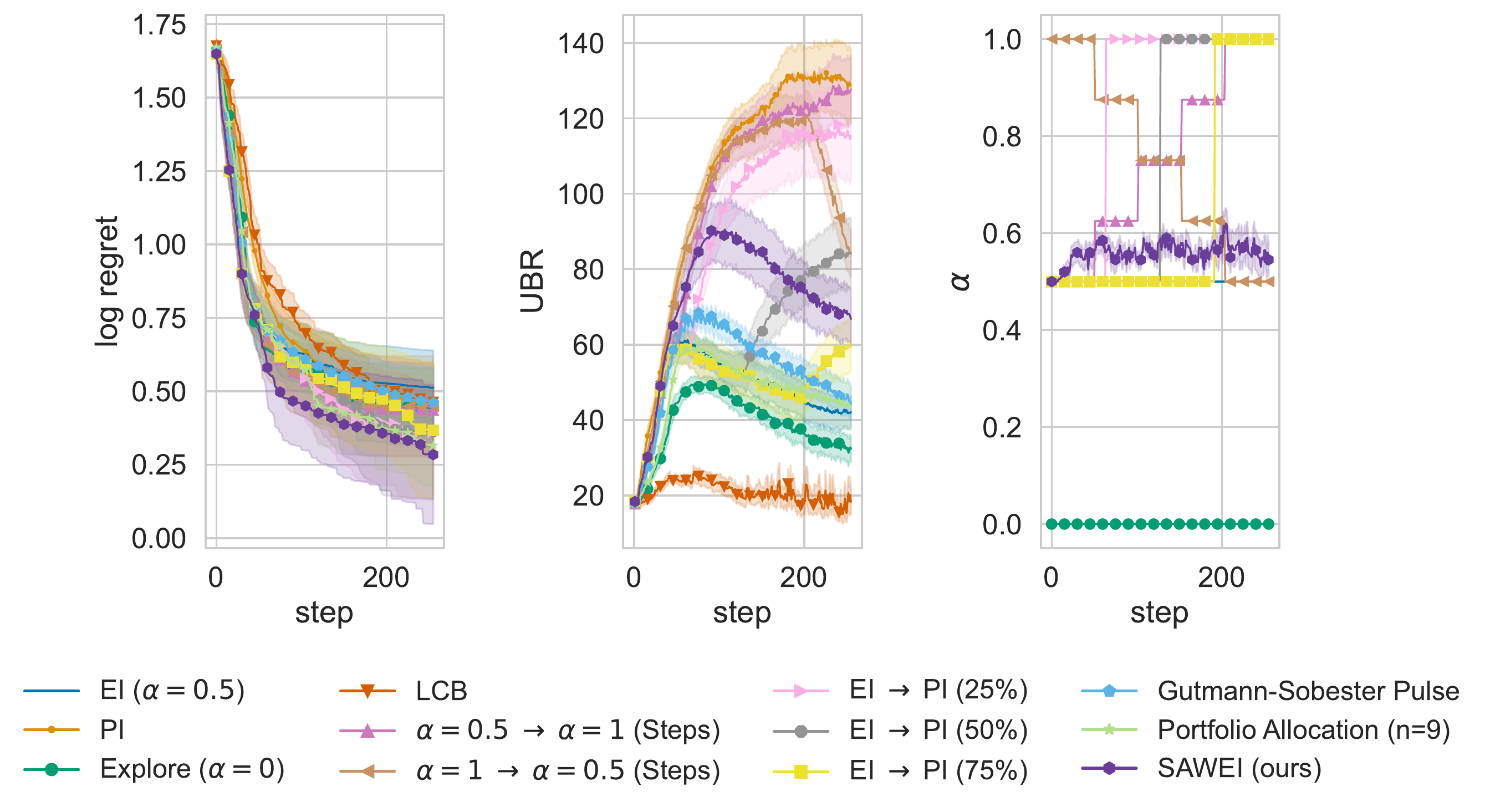}
    \caption{BBOB Function 21}
    \label{fig:figures/BBOB/alpha/021.pdf}
\end{figure}

\begin{figure}[h]
    \centering
    \includegraphics[width=0.85\linewidth]{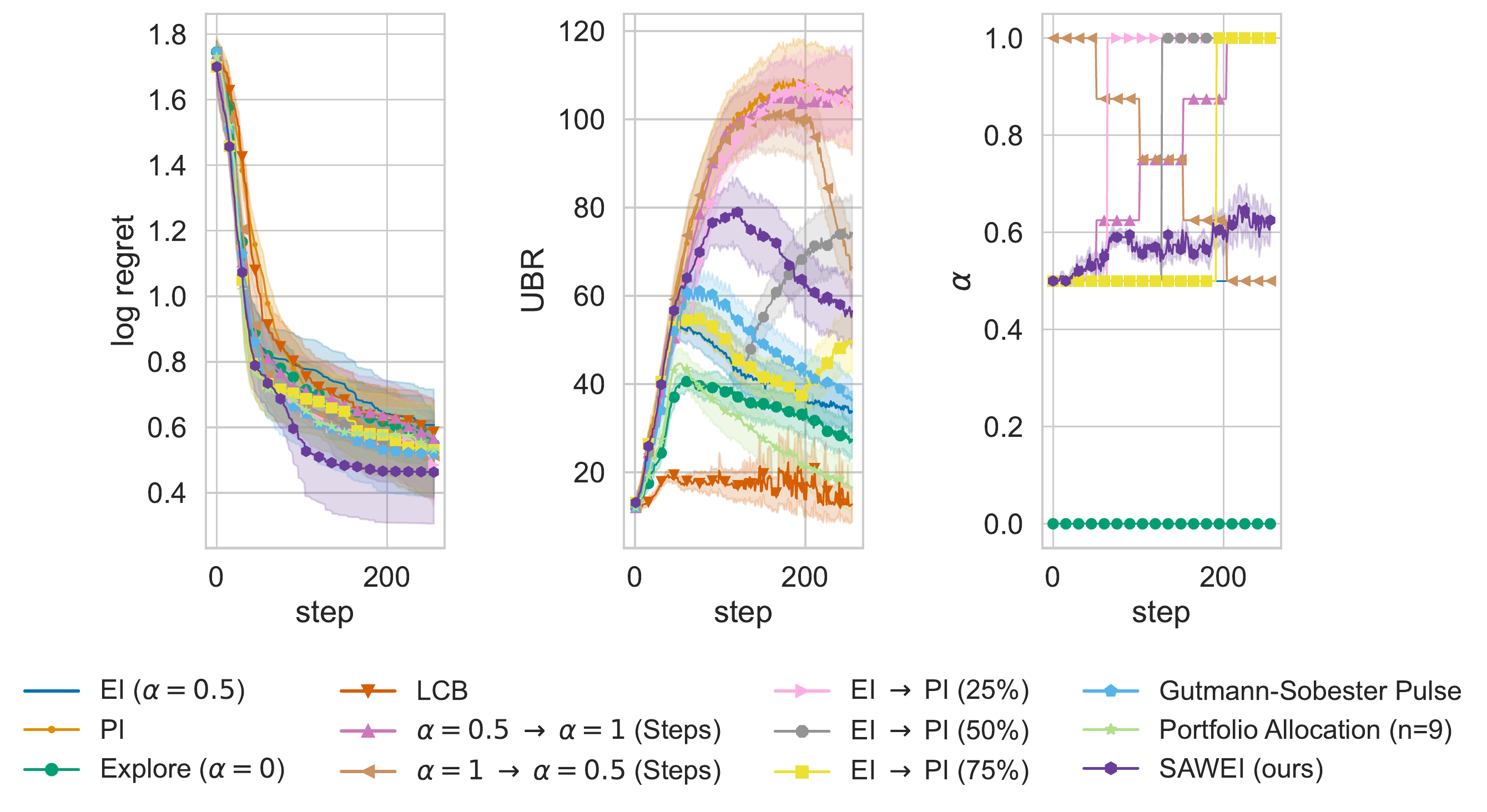}
    \caption{BBOB Function 22}
    \label{fig:figures/BBOB/alpha/022.pdf}
\end{figure}

\begin{figure}[h]
    \centering
    \includegraphics[width=0.85\linewidth]{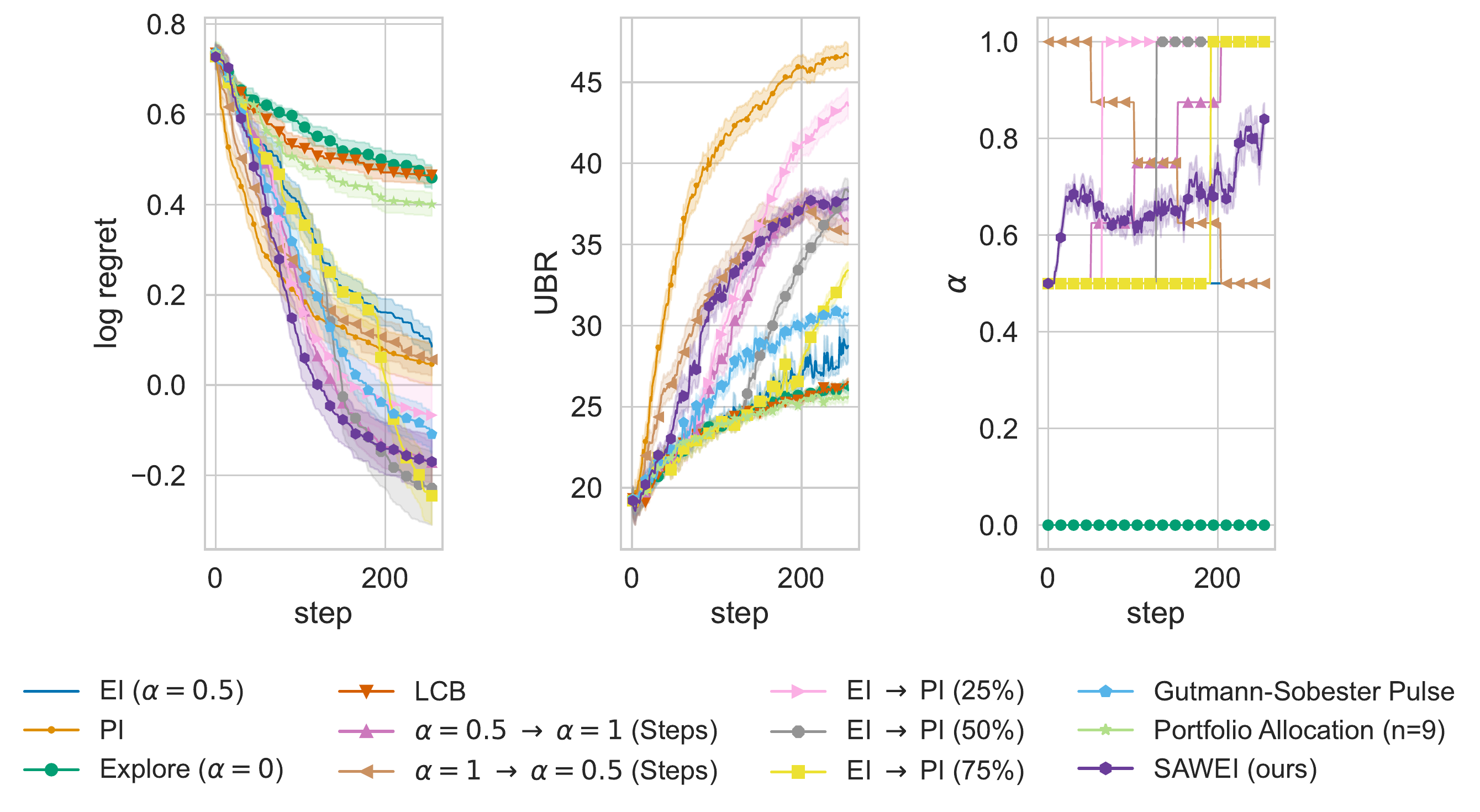}
    \caption{BBOB Function 23}
    \label{fig:figures/BBOB/alpha/023.pdf}
\end{figure}

\begin{figure}[h]
    \centering
    \includegraphics[width=0.85\linewidth]{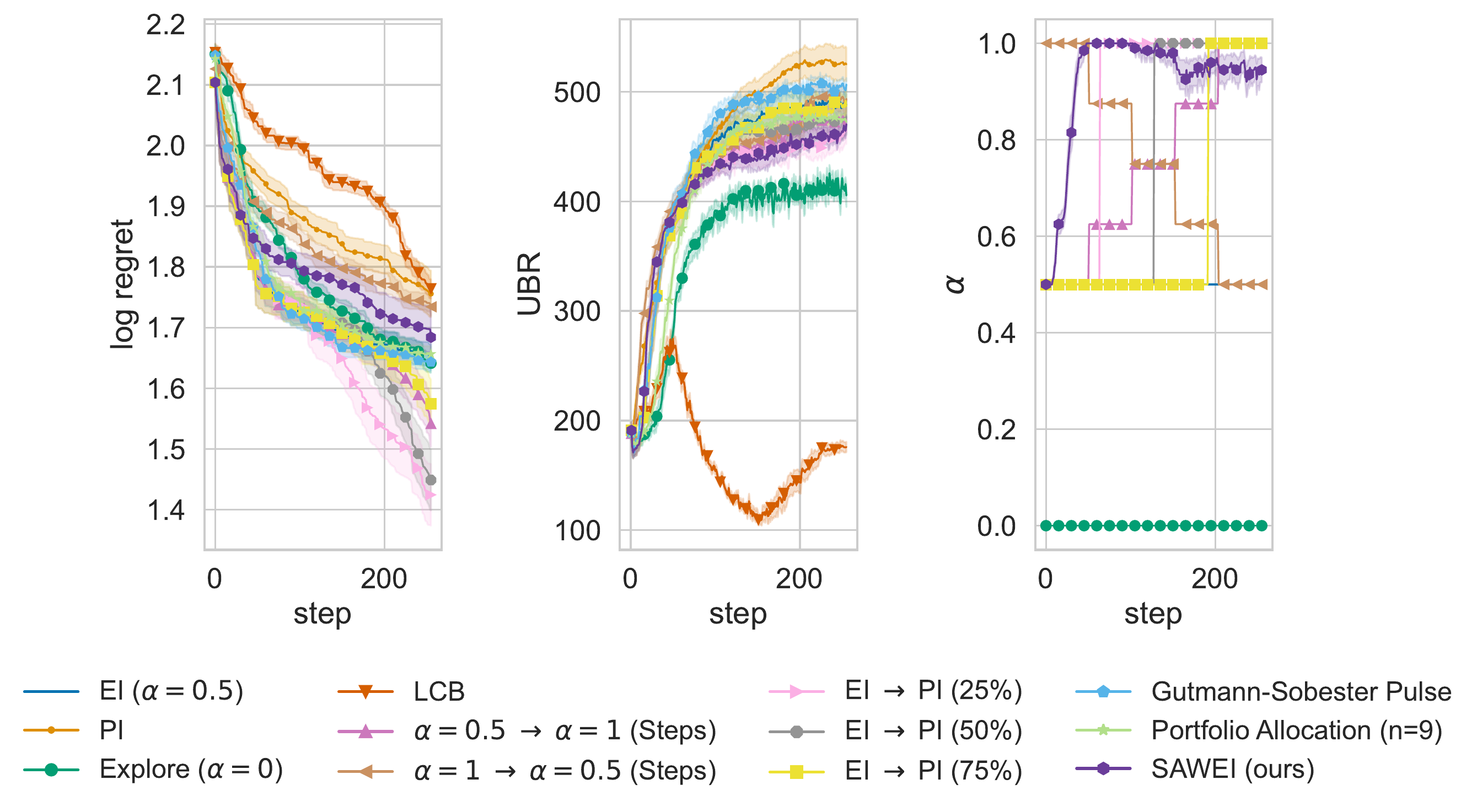}
    \caption{BBOB Function 24}
    \label{fig:figures/BBOB/alpha/024.pdf}
\end{figure}

\section{HPOBench Results}
\label{sec:app_hpob_results}

\begin{figure}[ht]
    \centering
    \includegraphics[width=0.9\textwidth]{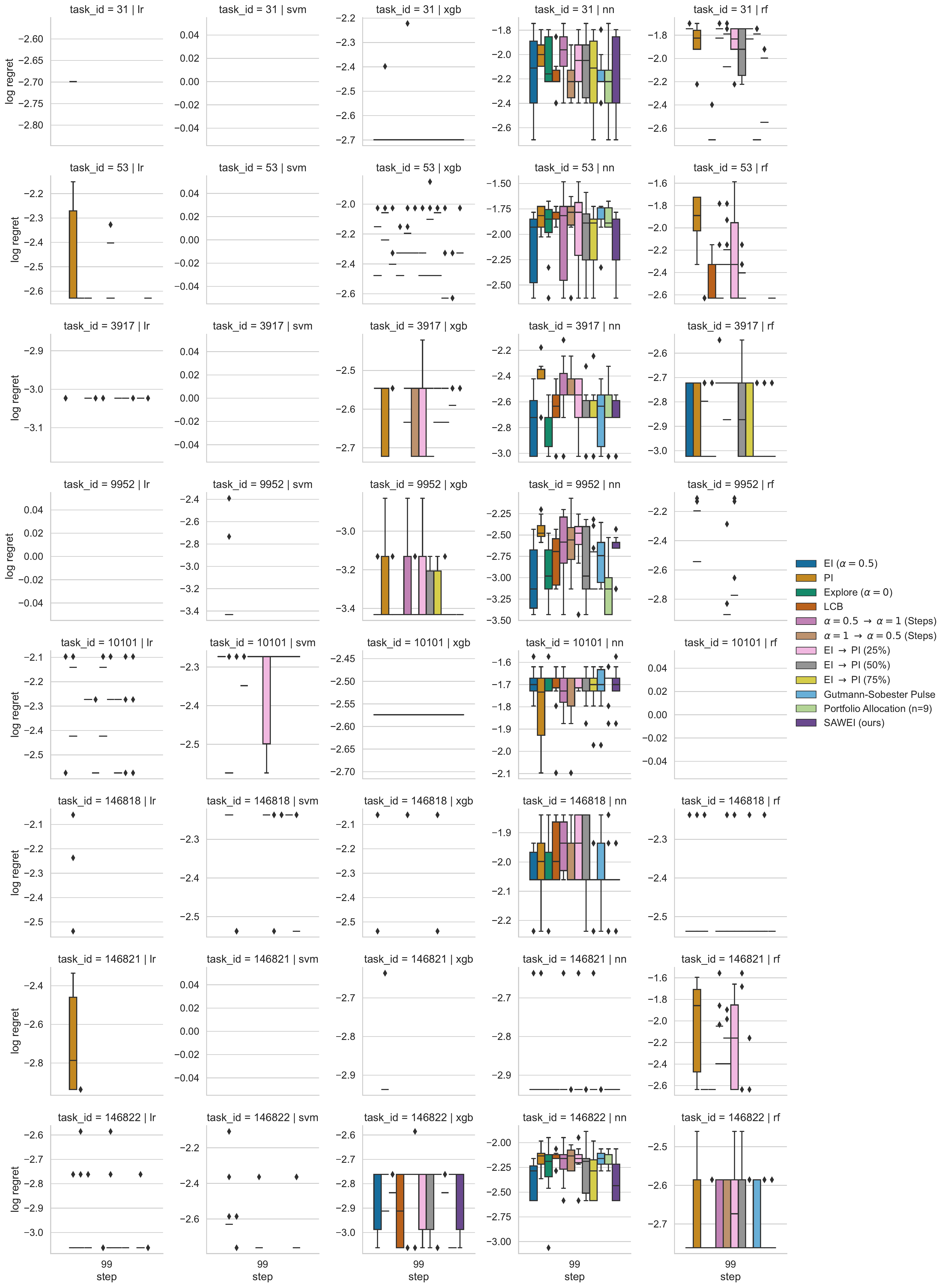}
    \caption{Final log regret on HPOBench (10 seeds). Please note that on this tabular benchmark a log regret of \num{0} can be achieved which is not plotted.}
    \label{fig:hpob_final_log_regret}
\end{figure}

\begin{figure}[h]
    \centering
    \includegraphics[width=0.85\linewidth]{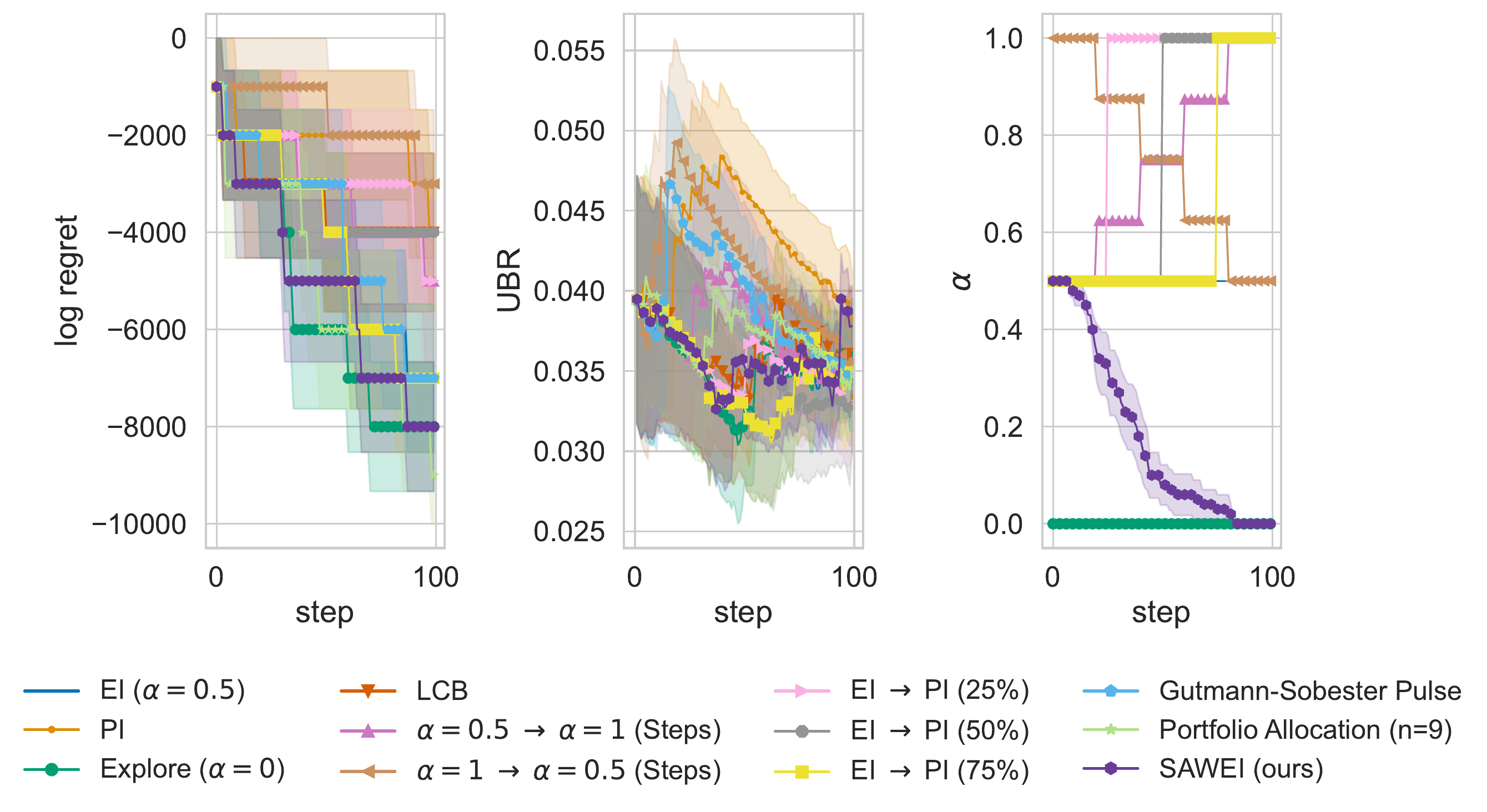}
    \caption{HPOBench ML: (model, task\_id) = (lr, 10101)}
    \label{fig:figures/HPOBench_det/alpha/lr_10101.pdf}
\end{figure}

\begin{figure}[h]
    \centering
    \includegraphics[width=0.85\linewidth]{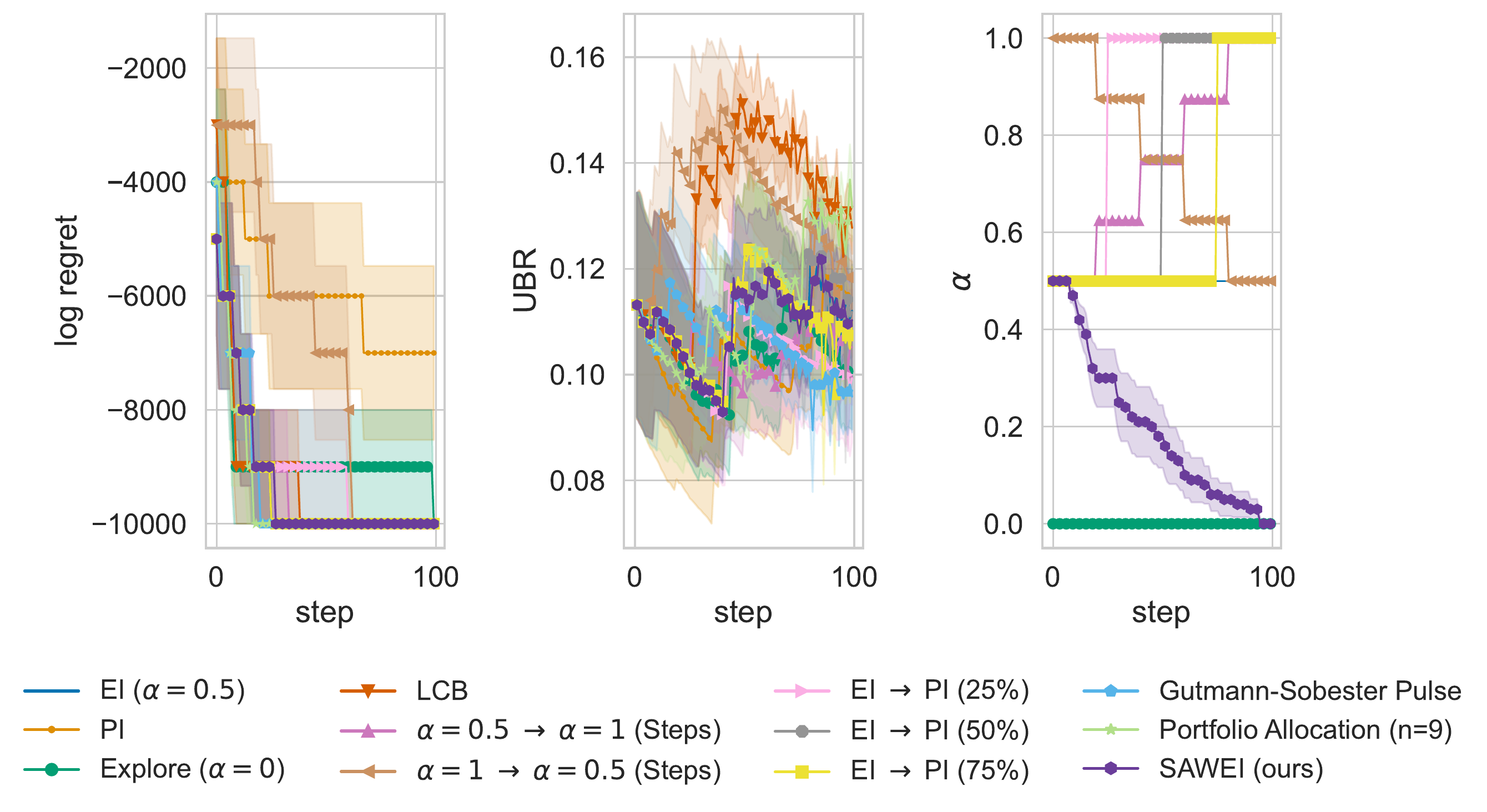}
    \caption{HPOBench ML: (model, task\_id) = (lr, 146818)}
    \label{fig:figures/HPOBench_det/alpha/lr_146818.pdf}
\end{figure}

\begin{figure}[h]
    \centering
    \includegraphics[width=0.85\linewidth]{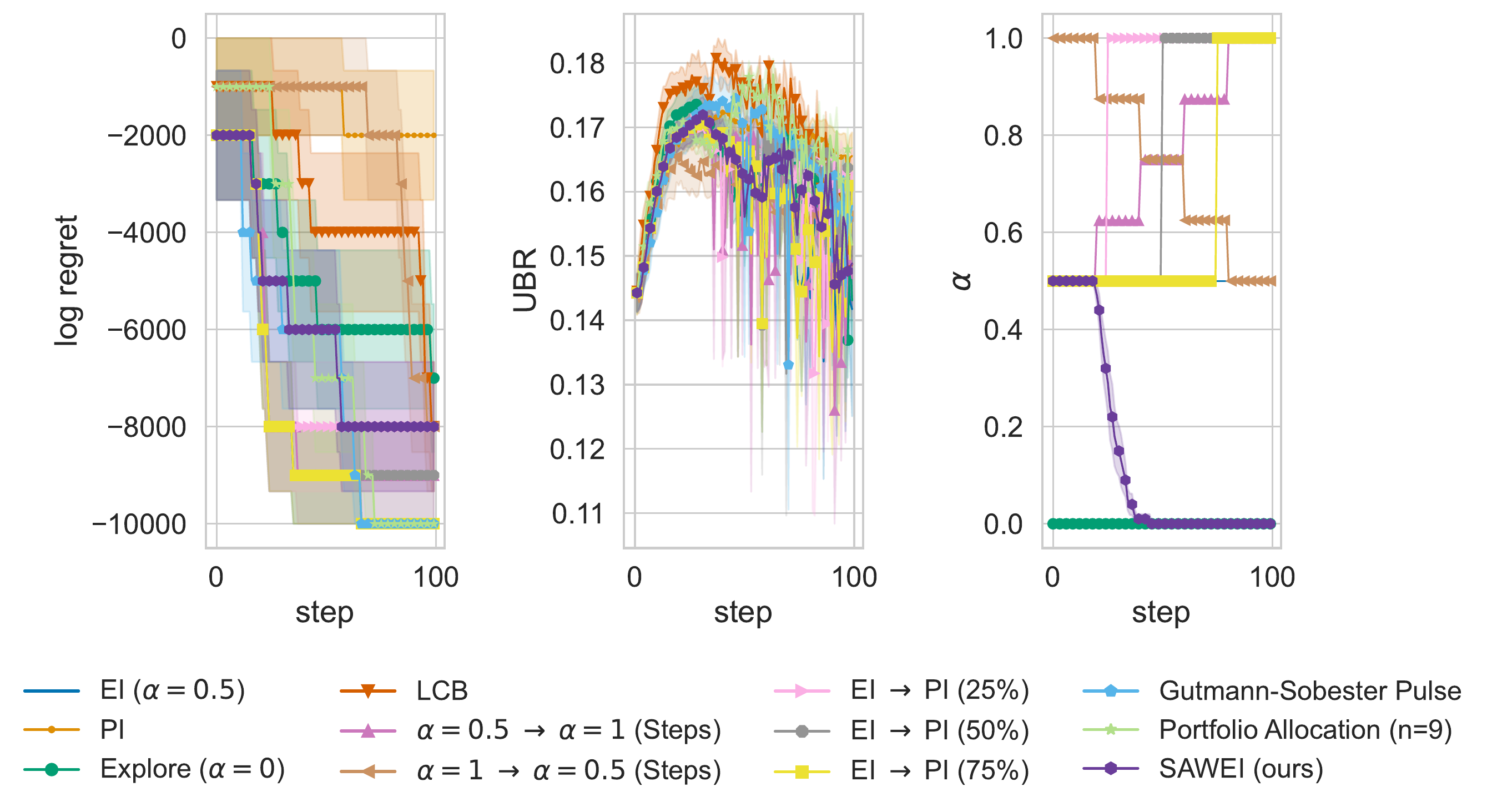}
    \caption{HPOBench ML: (model, task\_id) = (lr, 146821)}
    \label{fig:figures/HPOBench_det/alpha/lr_146821.pdf}
\end{figure}

\begin{figure}[h]
    \centering
    \includegraphics[width=0.85\linewidth]{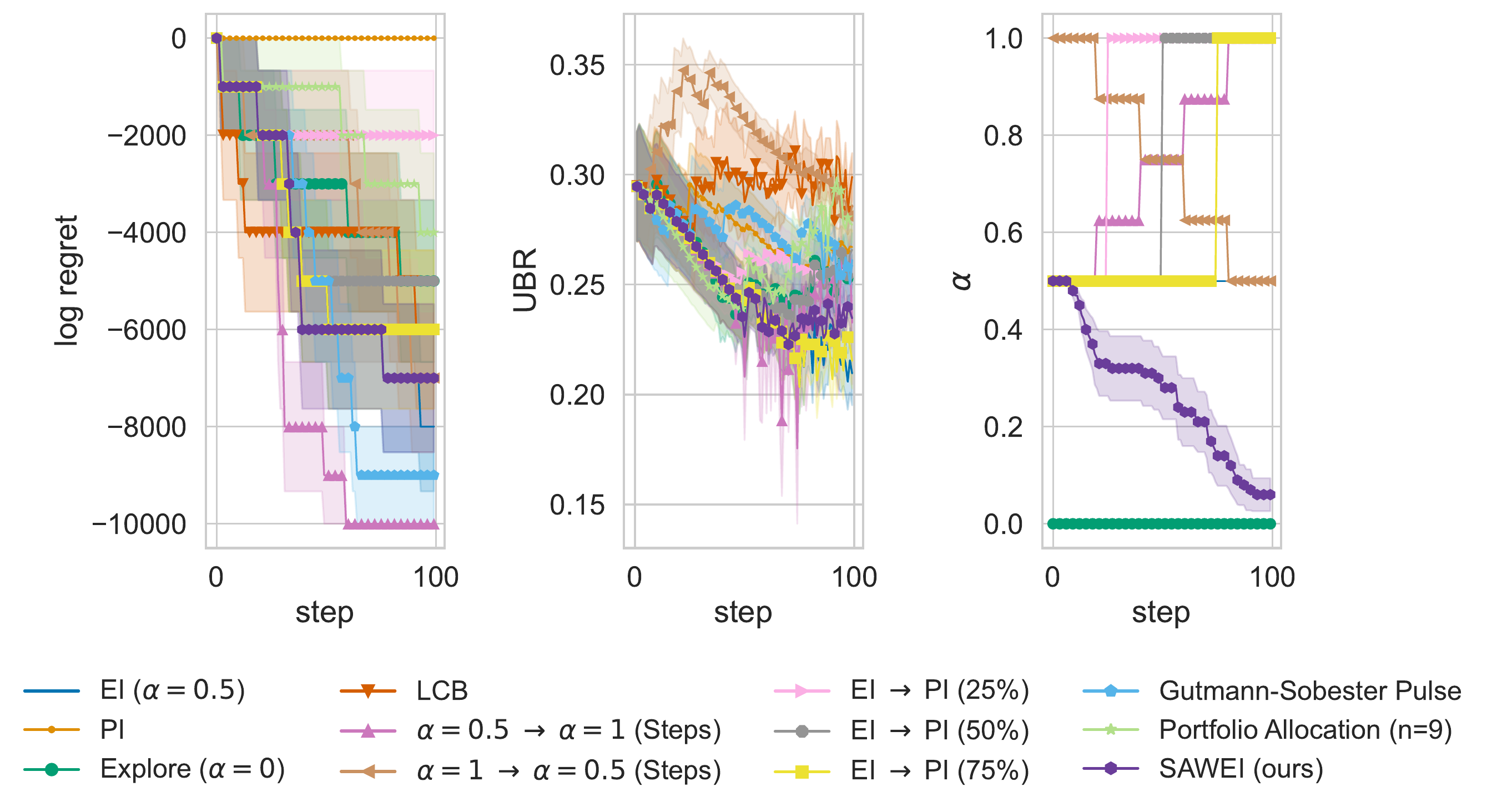}
    \caption{HPOBench ML: (model, task\_id) = (lr, 146822)}
    \label{fig:figures/HPOBench_det/alpha/lr_146822.pdf}
\end{figure}

\begin{figure}[h]
    \centering
    \includegraphics[width=0.85\linewidth]{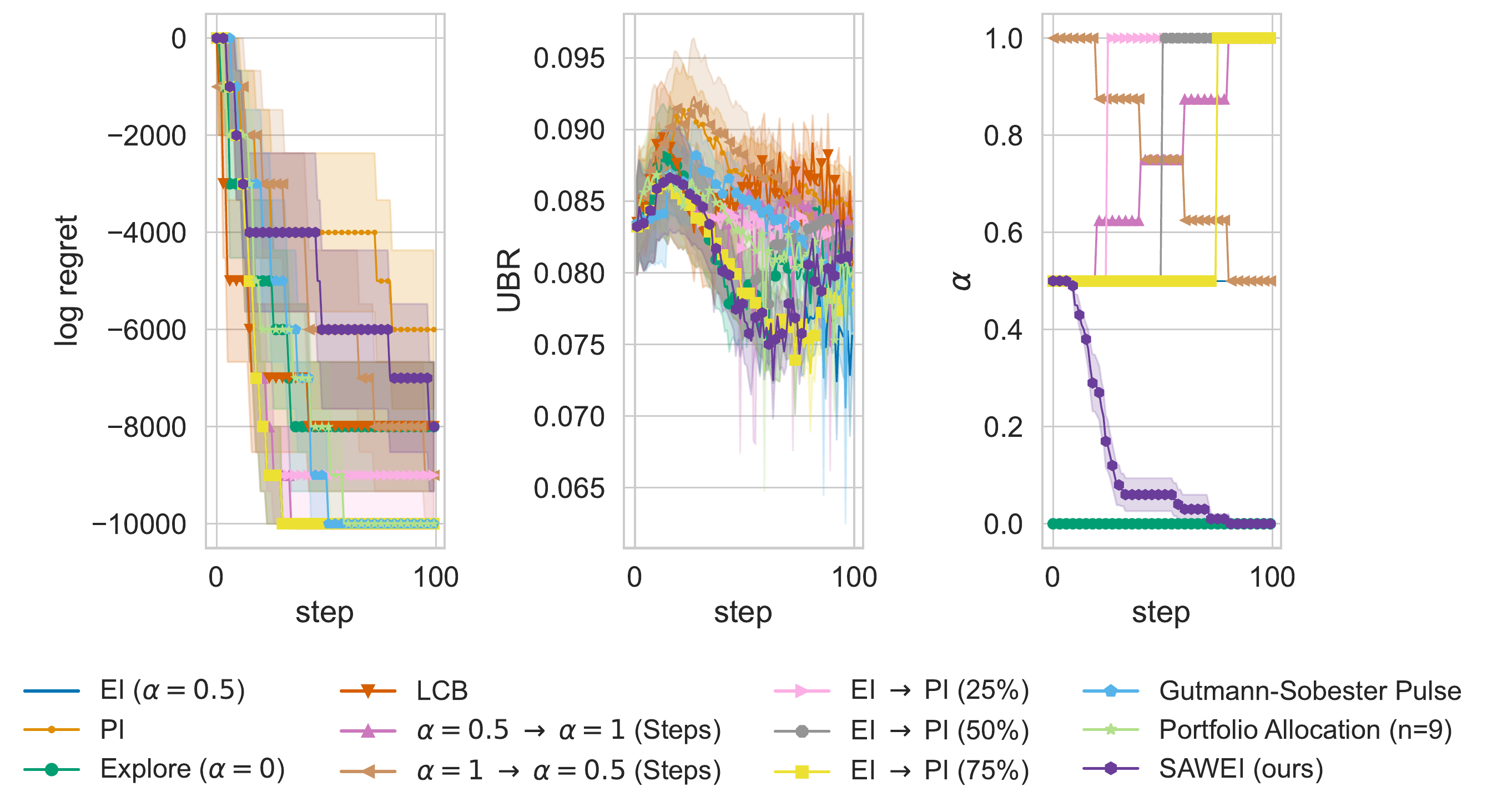}
    \caption{HPOBench ML: (model, task\_id) = (lr, 31)}
    \label{fig:figures/HPOBench_det/alpha/lr_31.pdf}
\end{figure}

\begin{figure}[h]
    \centering
    \includegraphics[width=0.85\linewidth]{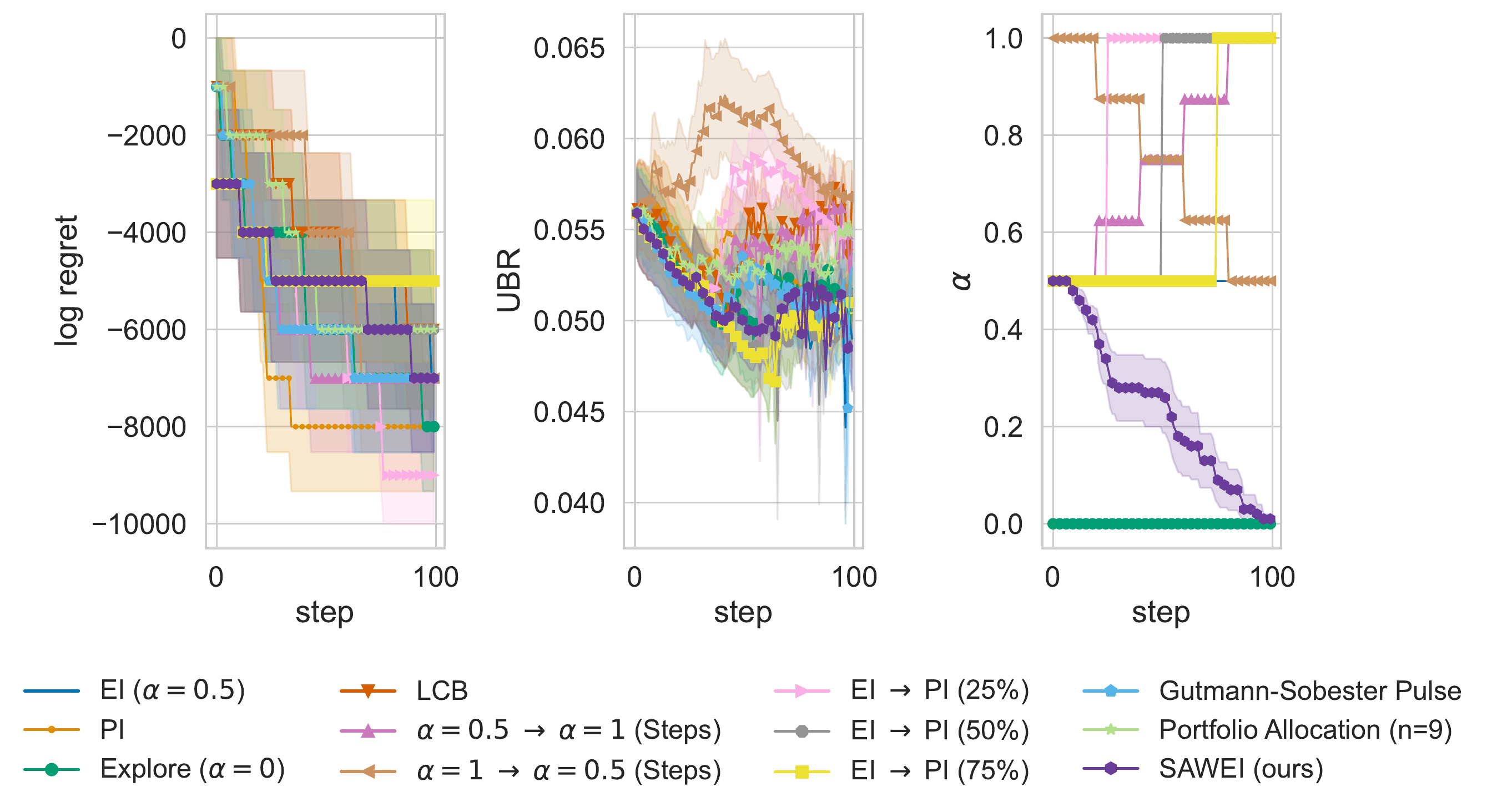}
    \caption{HPOBench ML: (model, task\_id) = (lr, 3917)}
    \label{fig:figures/HPOBench_det/alpha/lr_3917.pdf}
\end{figure}

\begin{figure}[h]
    \centering
    \includegraphics[width=0.85\linewidth]{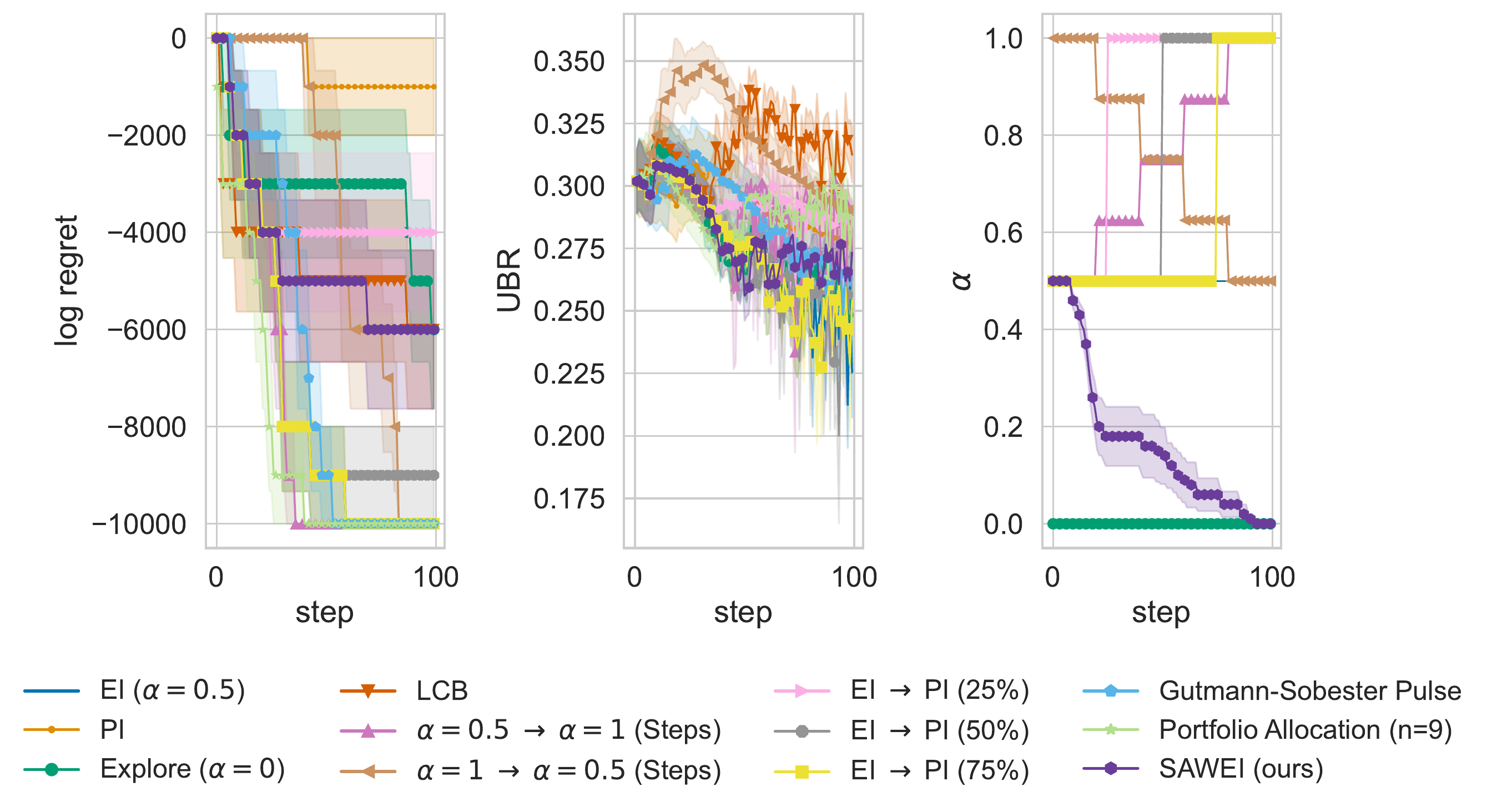}
    \caption{HPOBench ML: (model, task\_id) = (lr, 53)}
    \label{fig:figures/HPOBench_det/alpha/lr_53.pdf}
\end{figure}

\begin{figure}[h]
    \centering
    \includegraphics[width=0.85\linewidth]{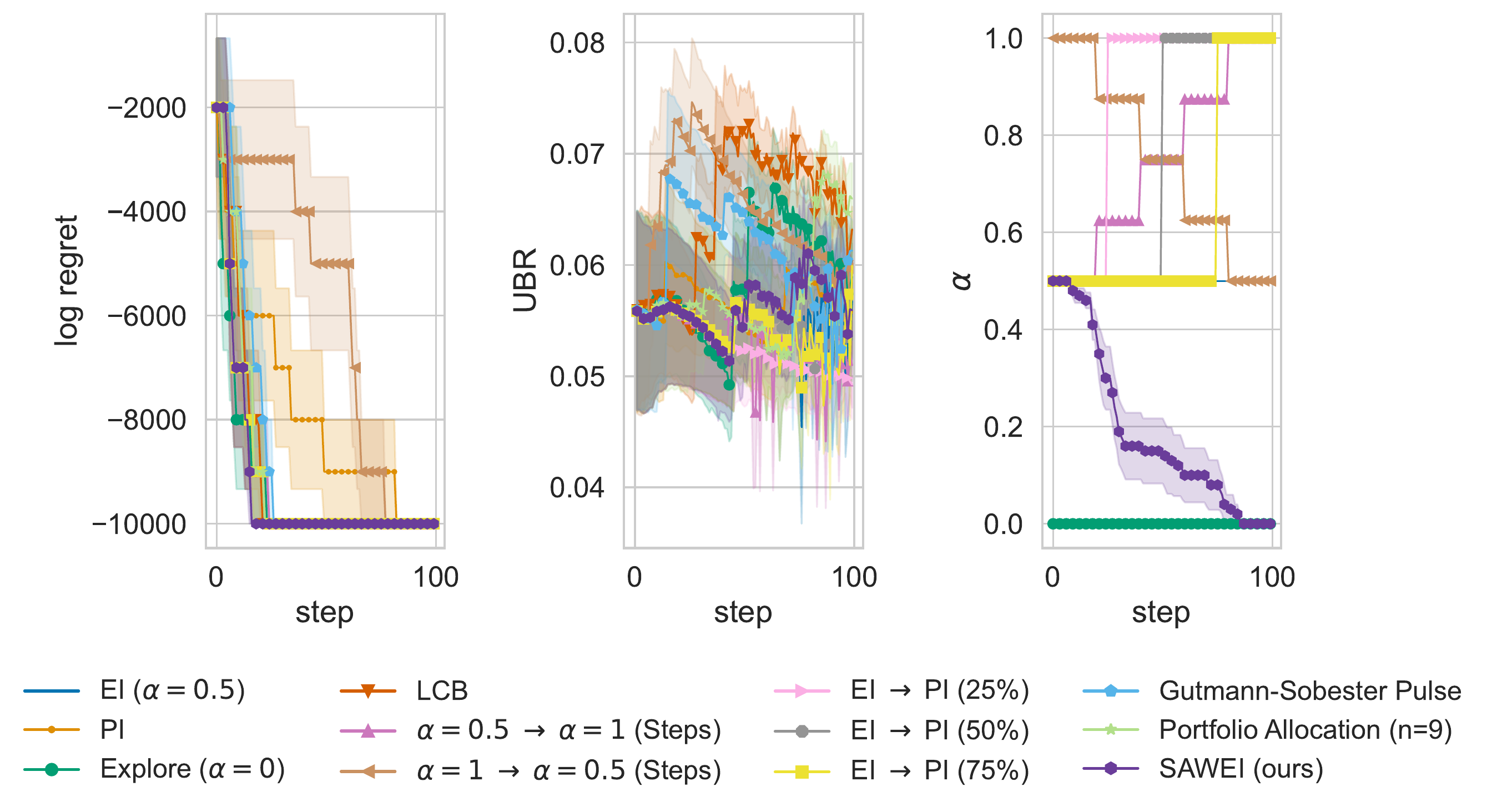}
    \caption{HPOBench ML: (model, task\_id) = (lr, 9952)}
    \label{fig:figures/HPOBench_det/alpha/lr_9952.pdf}
\end{figure}

\begin{figure}[h]
    \centering
    \includegraphics[width=0.85\linewidth]{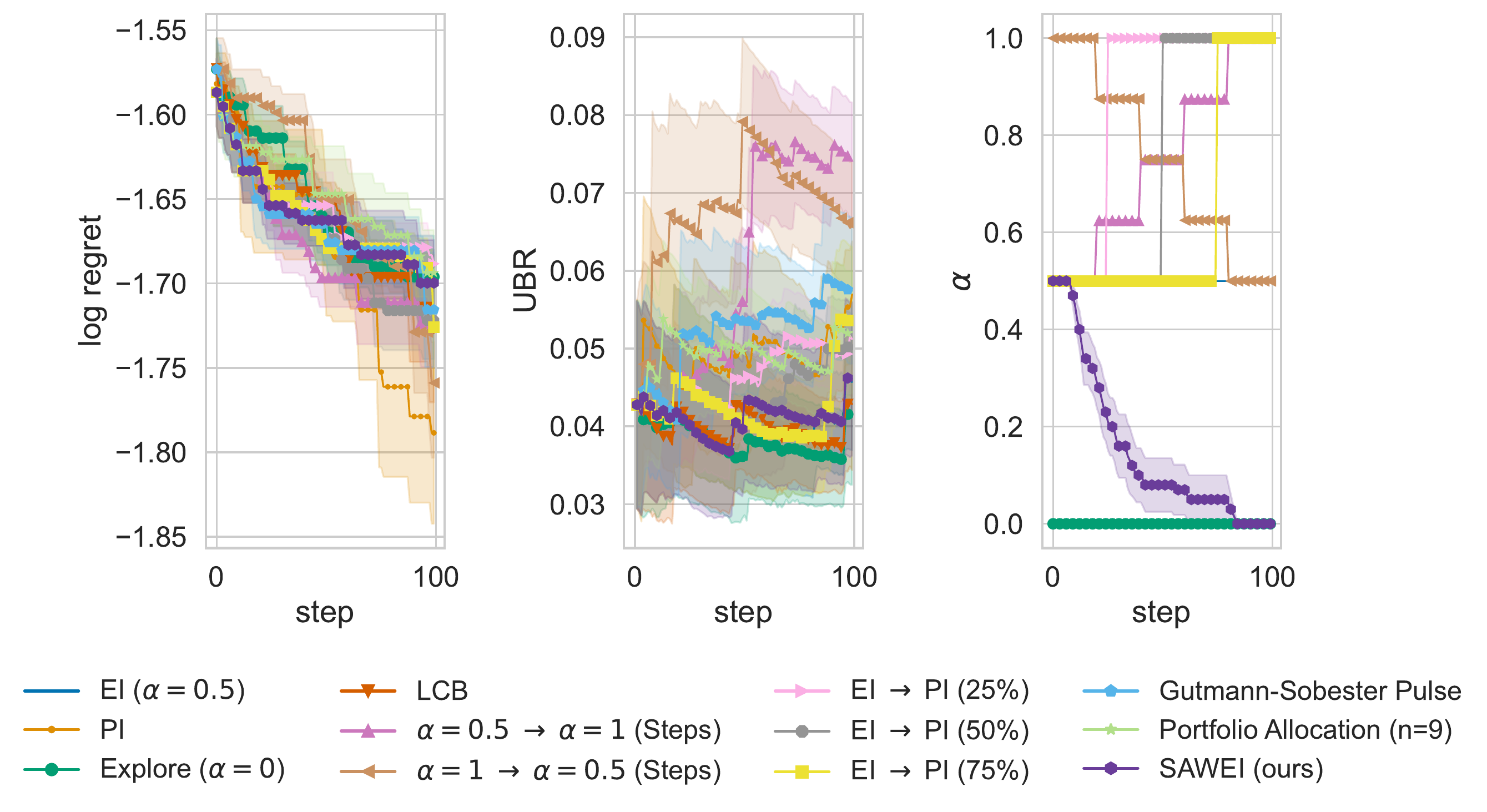}
    \caption{HPOBench ML: (model, task\_id) = (nn, 10101)}
    \label{fig:figures/HPOBench_det/alpha/nn_10101.pdf}
\end{figure}

\begin{figure}[h]
    \centering
    \includegraphics[width=0.85\linewidth]{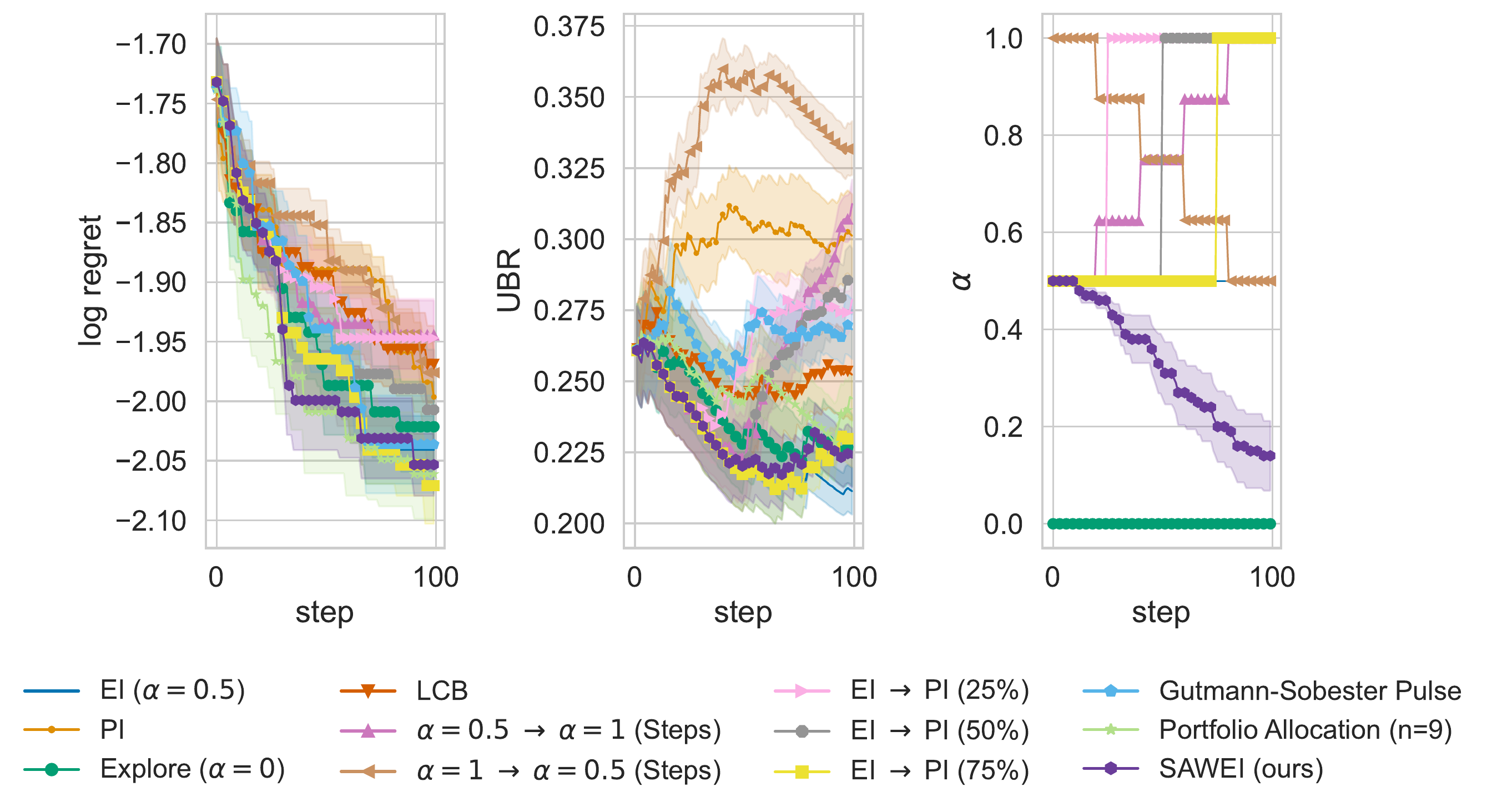}
    \caption{HPOBench ML: (model, task\_id) = (nn, 146818)}
    \label{fig:figures/HPOBench_det/alpha/nn_146818.pdf}
\end{figure}

\begin{figure}[h]
    \centering
    \includegraphics[width=0.85\linewidth]{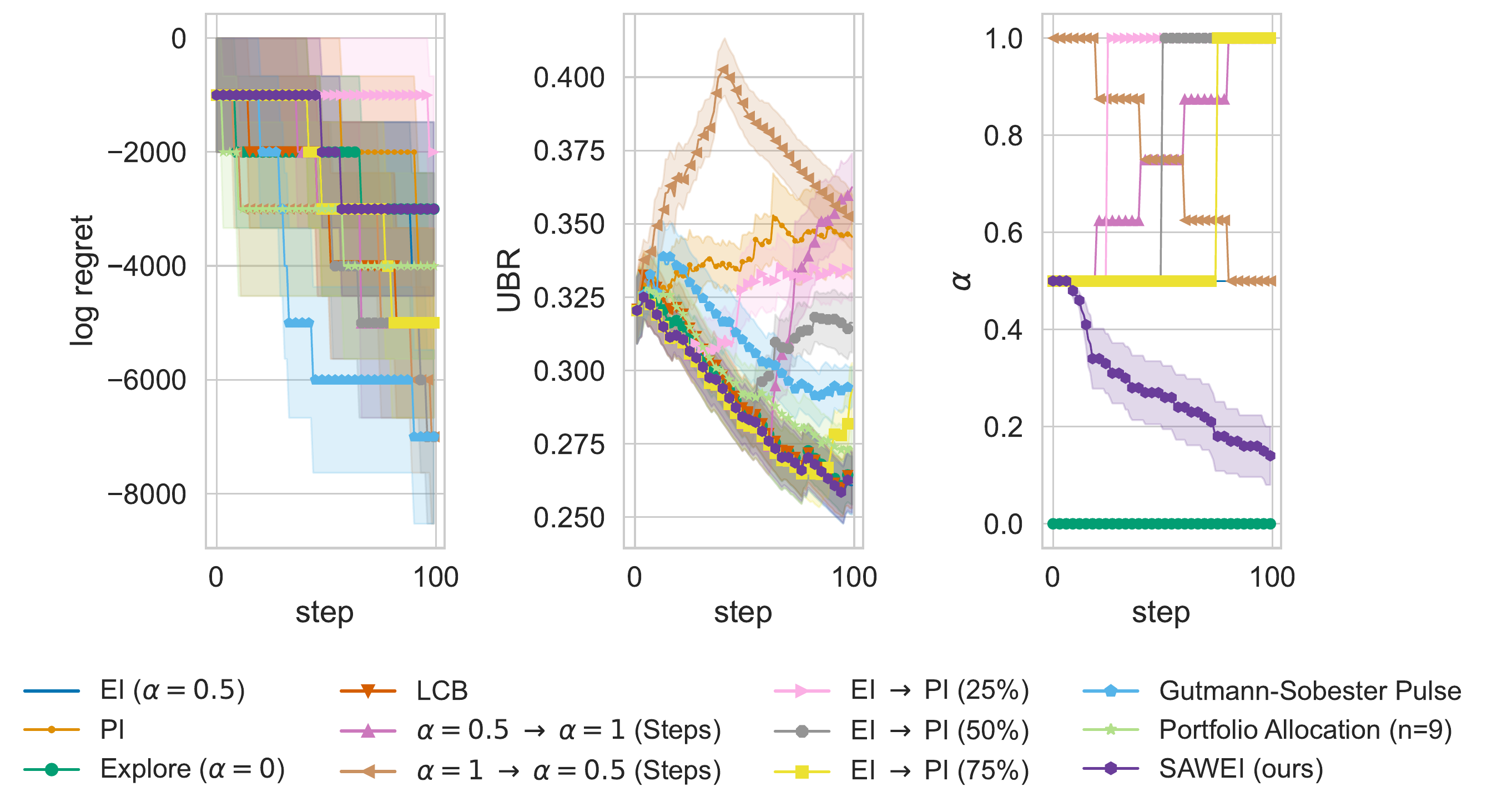}
    \caption{HPOBench ML: (model, task\_id) = (nn, 146821)}
    \label{fig:figures/HPOBench_det/alpha/nn_146821.pdf}
\end{figure}

\begin{figure}[h]
    \centering
    \includegraphics[width=0.85\linewidth]{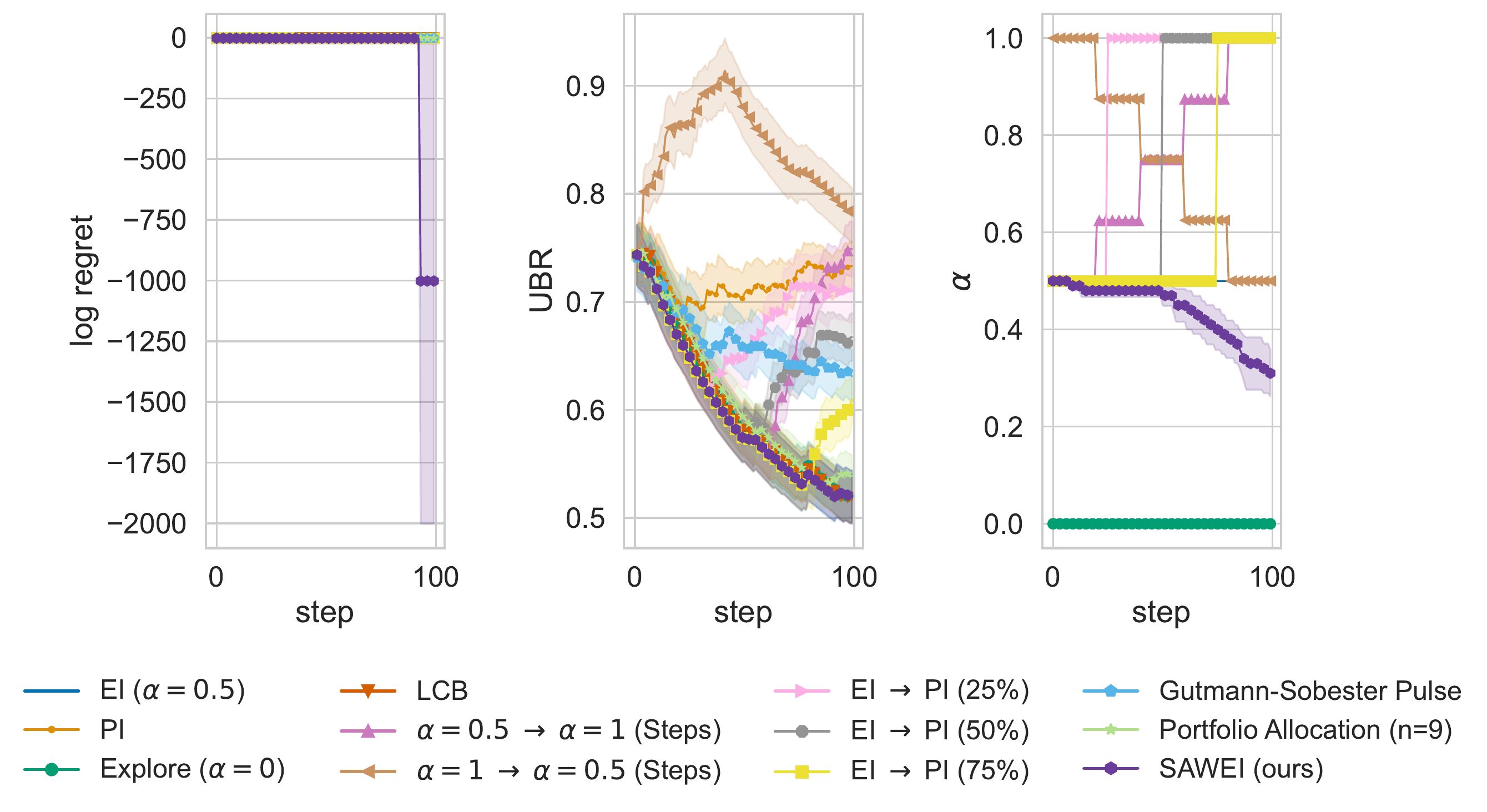}
    \caption{HPOBench ML: (model, task\_id) = (nn, 146822)}
    \label{fig:figures/HPOBench_det/alpha/nn_146822.pdf}
\end{figure}

\begin{figure}[h]
    \centering
    \includegraphics[width=0.85\linewidth]{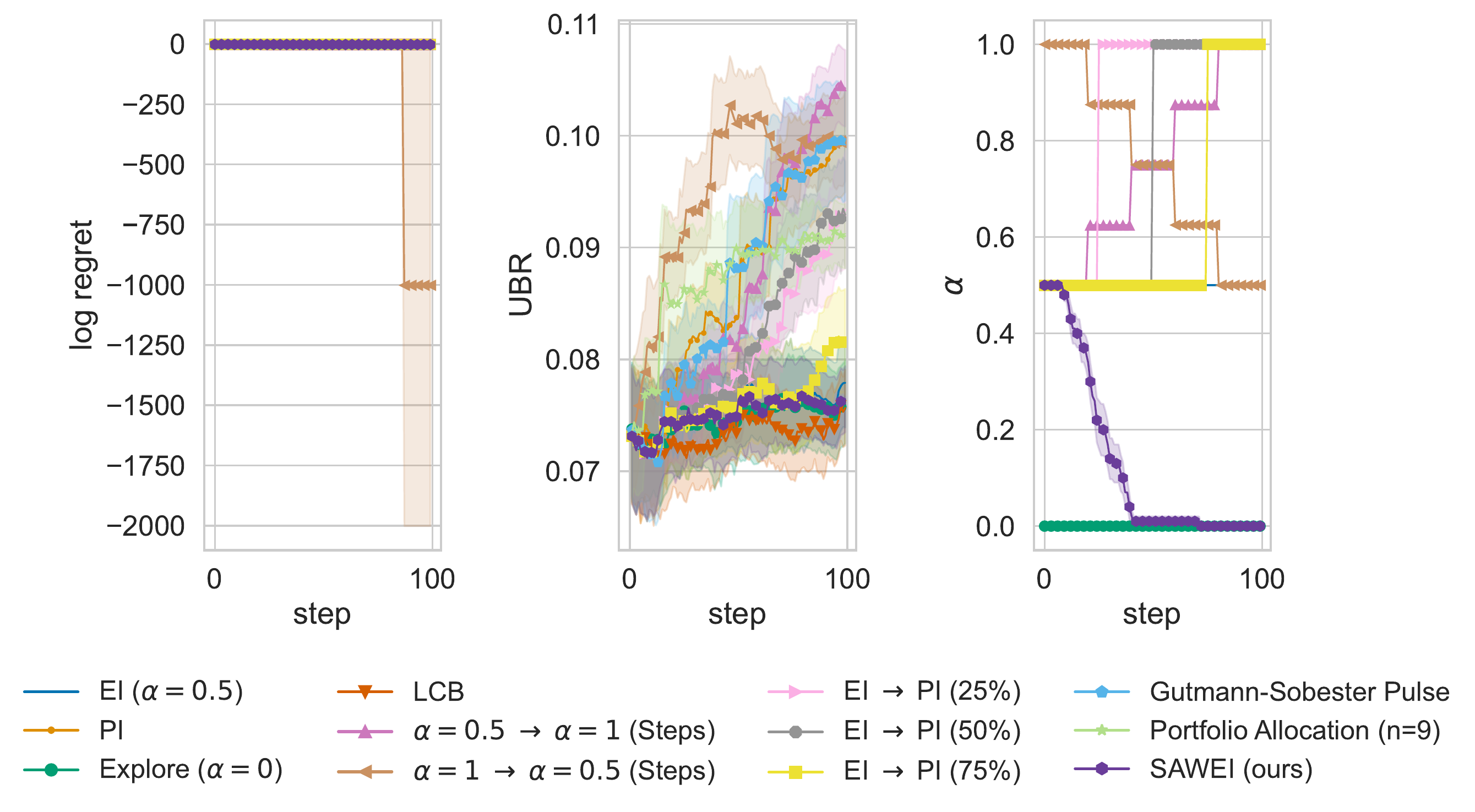}
    \caption{HPOBench ML: (model, task\_id) = (nn, 31)}
    \label{fig:figures/HPOBench_det/alpha/nn_31.pdf}
\end{figure}

\begin{figure}[h]
    \centering
    \includegraphics[width=0.85\linewidth]{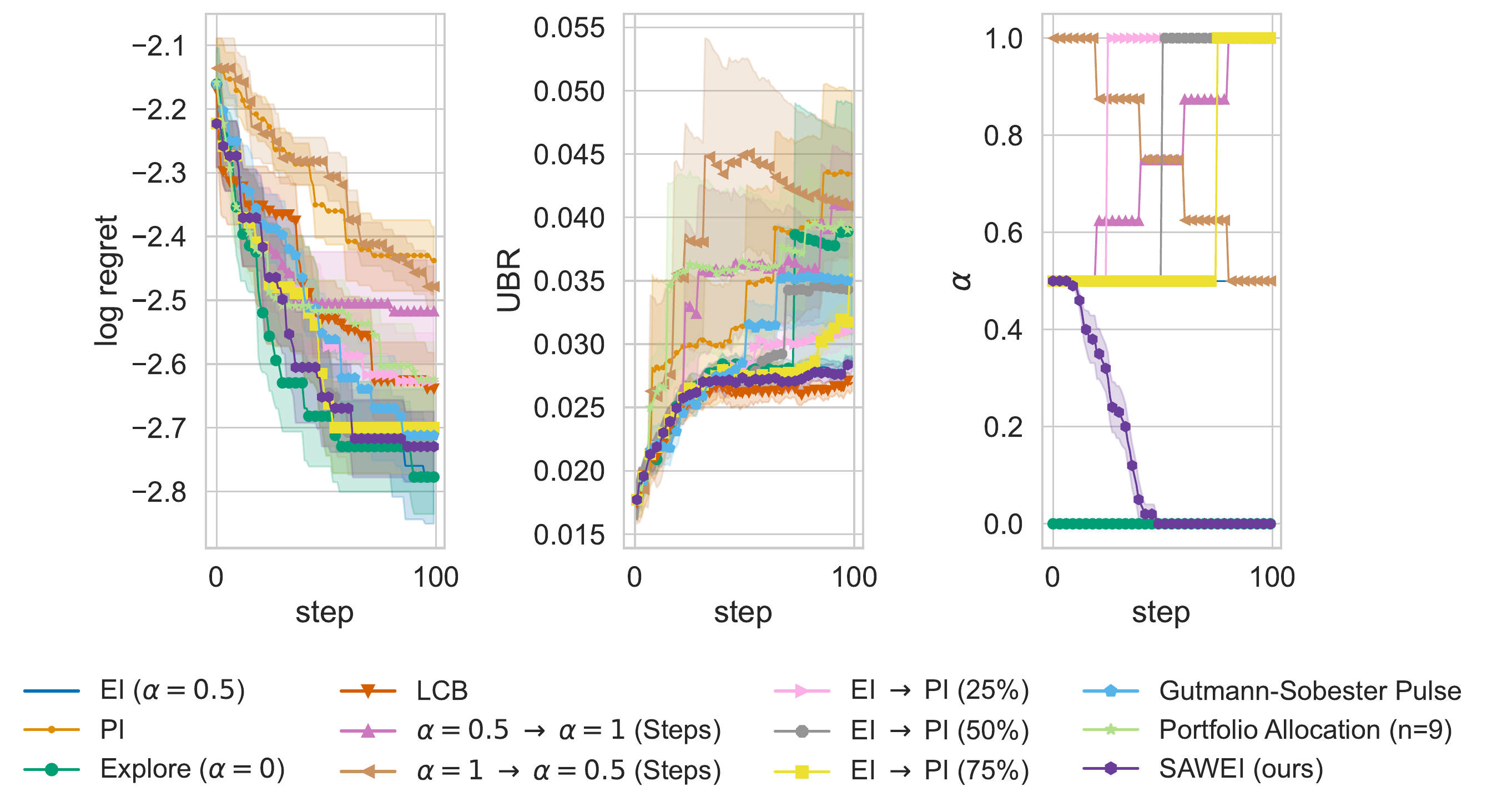}
    \caption{HPOBench ML: (model, task\_id) = (nn, 3917)}
    \label{fig:figures/HPOBench_det/alpha/nn_3917.pdf}
\end{figure}

\begin{figure}[h]
    \centering
    \includegraphics[width=0.85\linewidth]{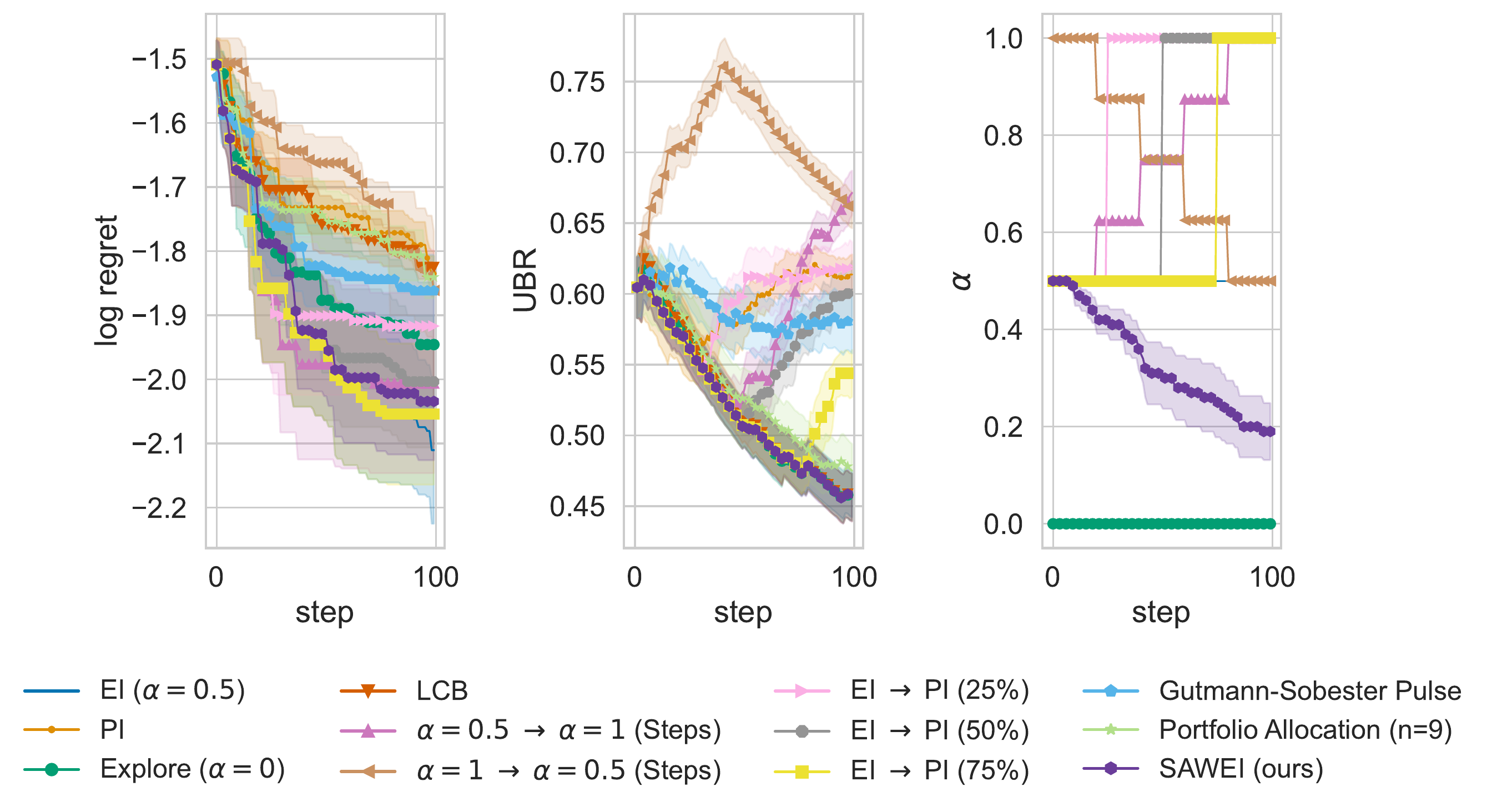}
    \caption{HPOBench ML: (model, task\_id) = (nn, 53)}
    \label{fig:figures/HPOBench_det/alpha/nn_53.pdf}
\end{figure}

\begin{figure}[h]
    \centering
    \includegraphics[width=0.85\linewidth]{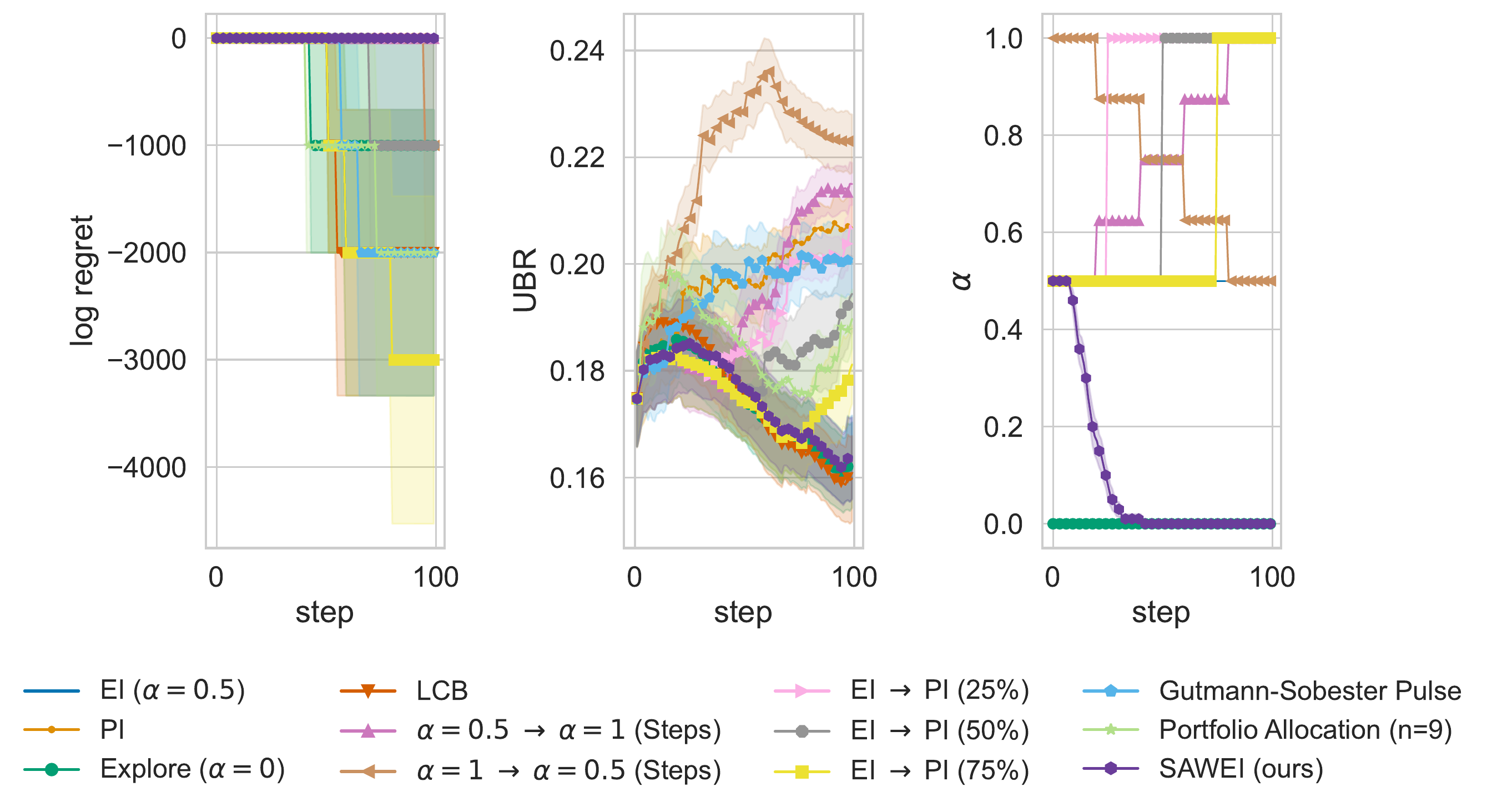}
    \caption{HPOBench ML: (model, task\_id) = (nn, 9952)}
    \label{fig:figures/HPOBench_det/alpha/nn_9952.pdf}
\end{figure}

\begin{figure}[h]
    \centering
    \includegraphics[width=0.85\linewidth]{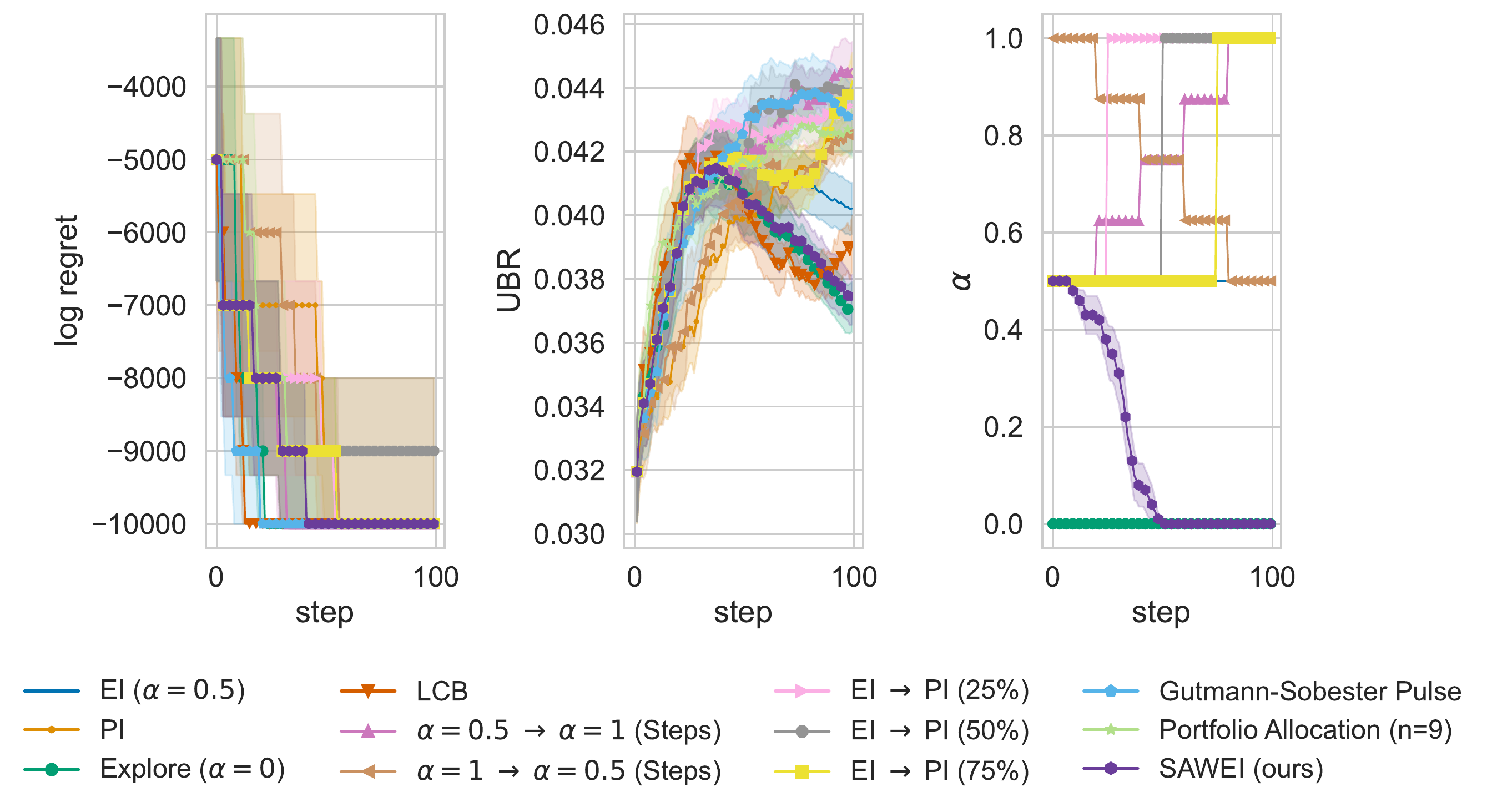}
    \caption{HPOBench ML: (model, task\_id) = (rf, 10101)}
    \label{fig:figures/HPOBench_det/alpha/rf_10101.pdf}
\end{figure}

\begin{figure}[h]
    \centering
    \includegraphics[width=0.85\linewidth]{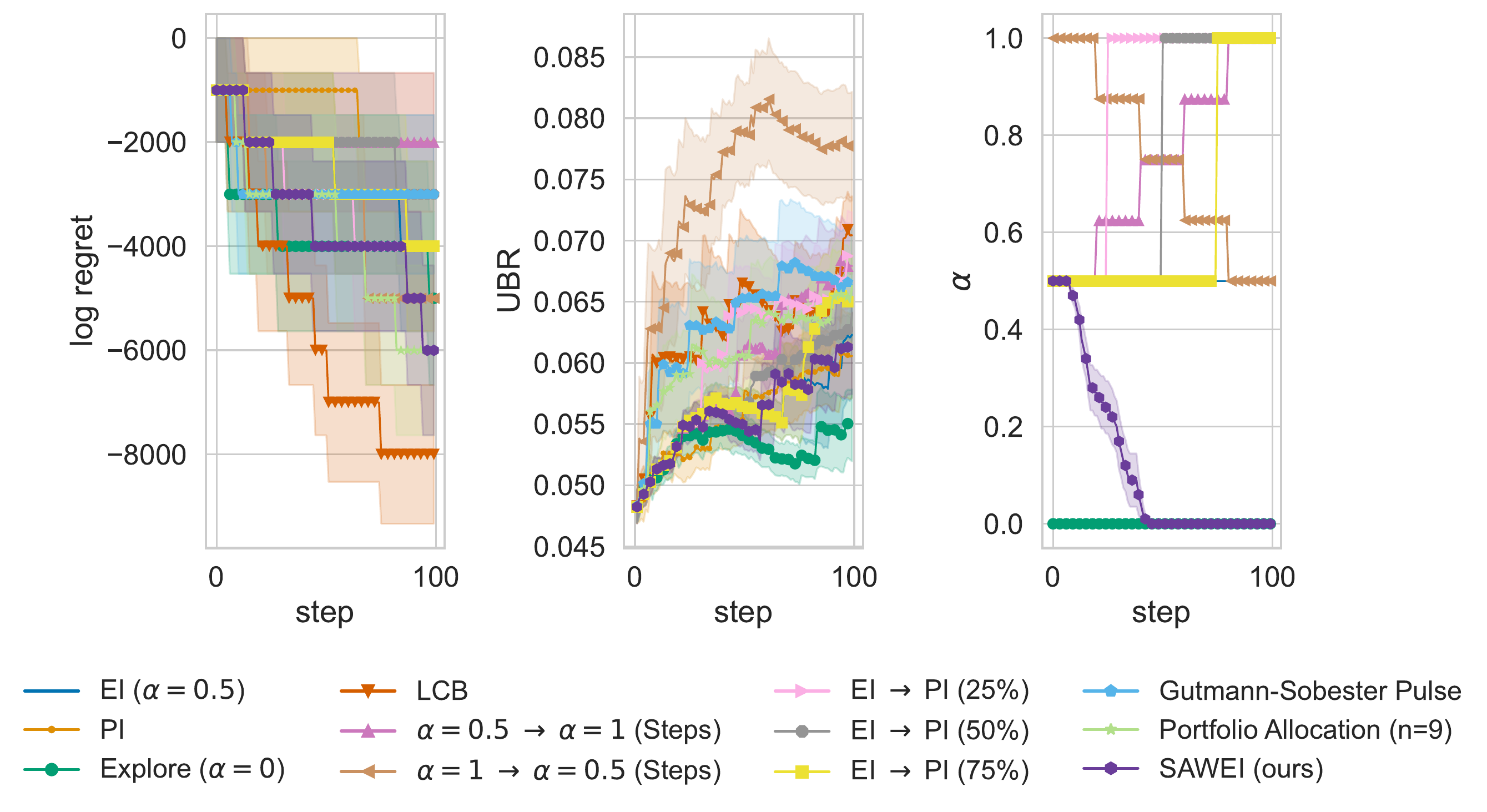}
    \caption{HPOBench ML: (model, task\_id) = (rf, 146818)}
    \label{fig:figures/HPOBench_det/alpha/rf_146818.pdf}
\end{figure}

\begin{figure}[h]
    \centering
    \includegraphics[width=0.85\linewidth]{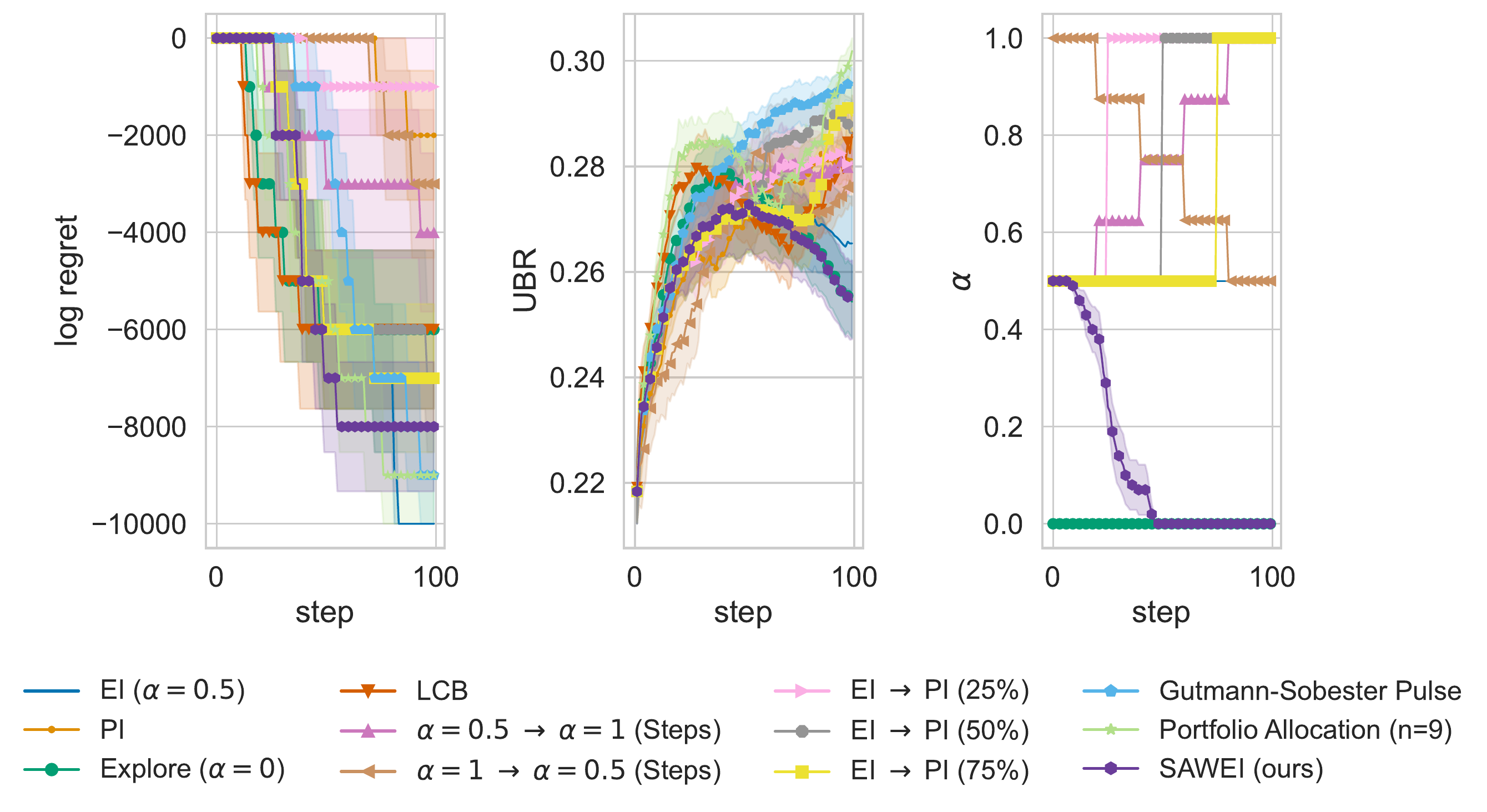}
    \caption{HPOBench ML: (model, task\_id) = (rf, 146821)}
    \label{fig:figures/HPOBench_det/alpha/rf_146821.pdf}
\end{figure}

\begin{figure}[h]
    \centering
    \includegraphics[width=0.85\linewidth]{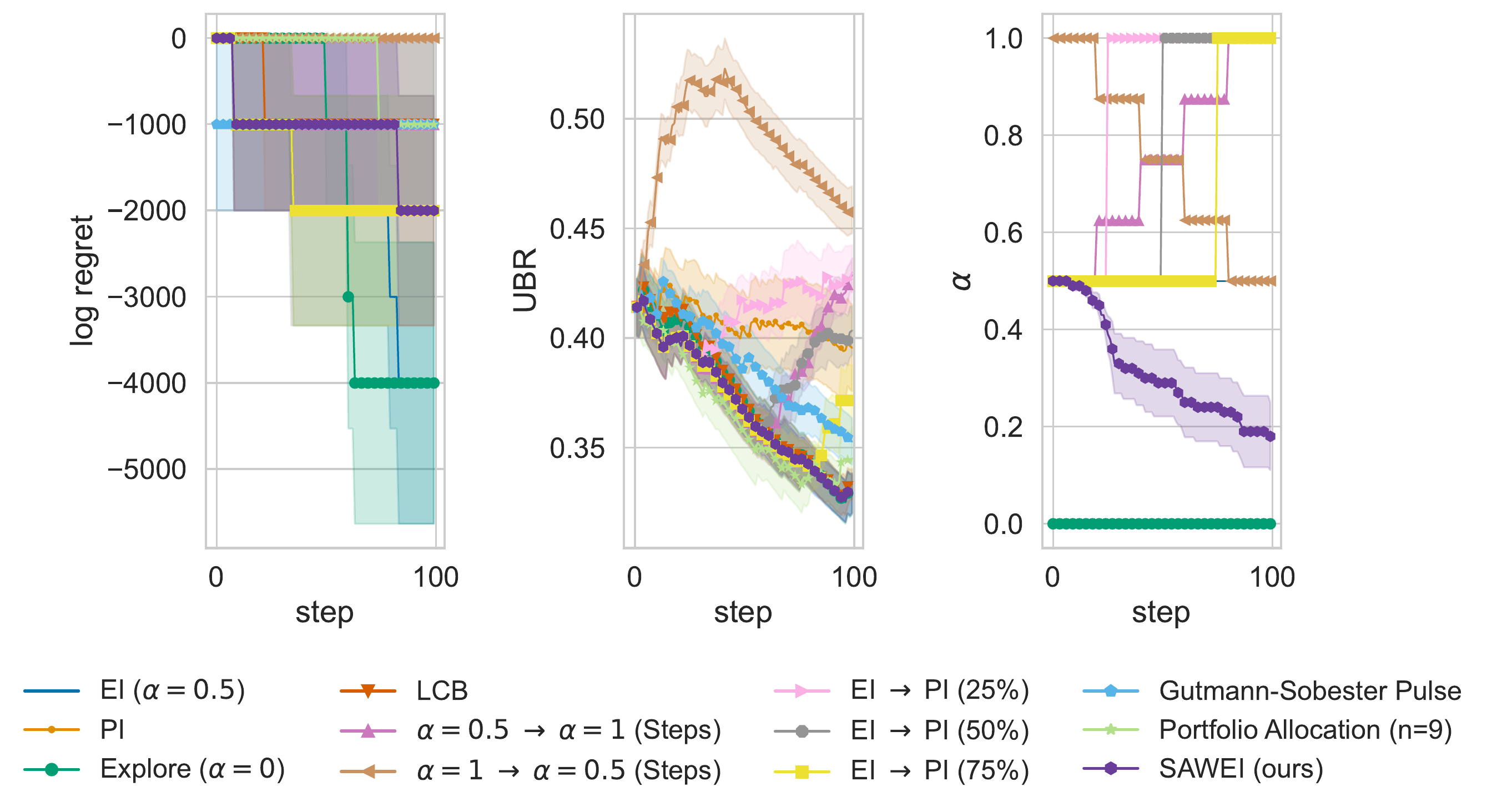}
    \caption{HPOBench ML: (model, task\_id) = (rf, 146822)}
    \label{fig:figures/HPOBench_det/alpha/rf_146822.pdf}
\end{figure}

\begin{figure}[h]
    \centering
    \includegraphics[width=0.85\linewidth]{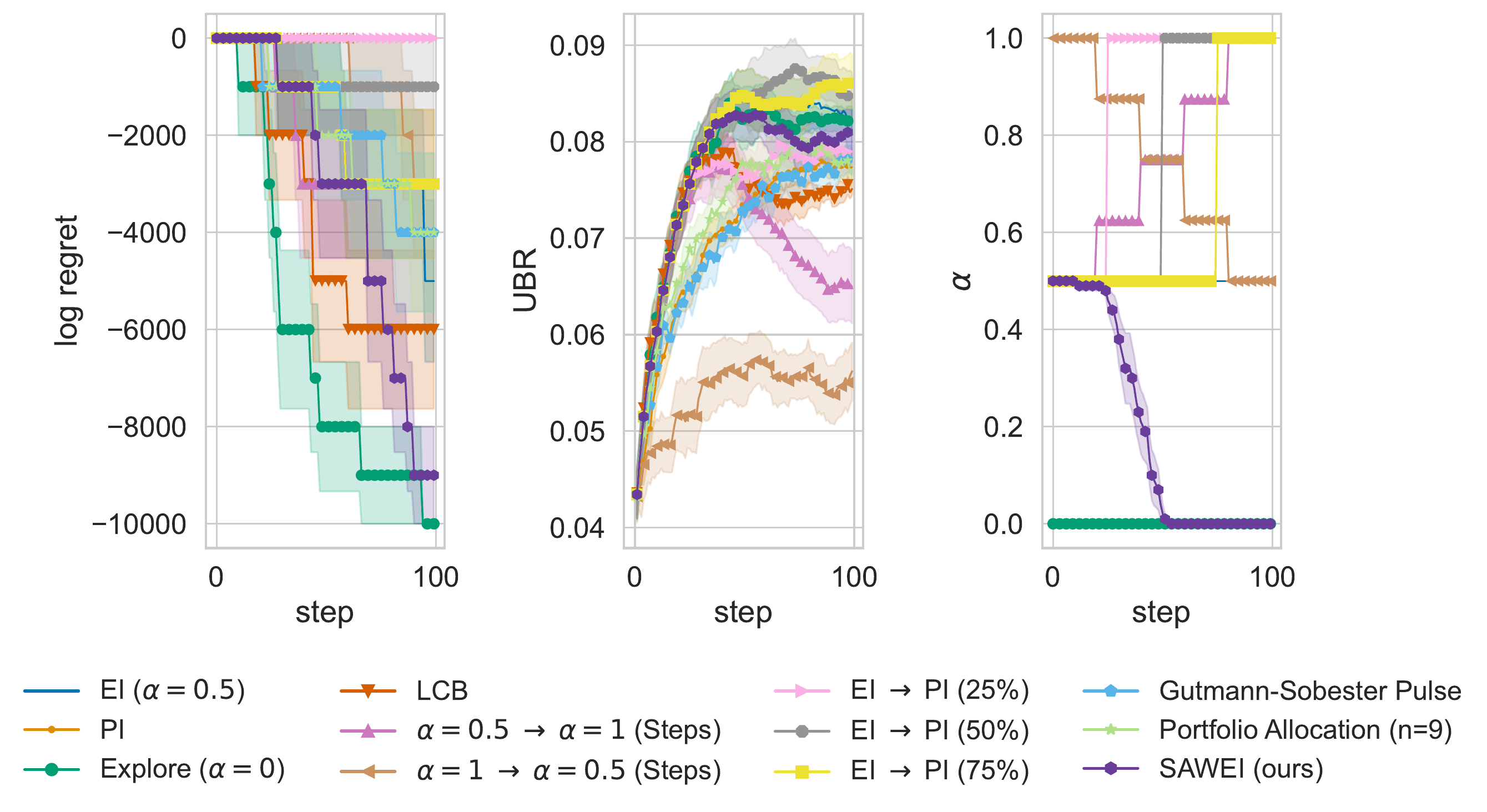}
    \caption{HPOBench ML: (model, task\_id) = (rf, 31)}
    \label{fig:figures/HPOBench_det/alpha/rf_31.pdf}
\end{figure}

\begin{figure}[h]
    \centering
    \includegraphics[width=0.85\linewidth]{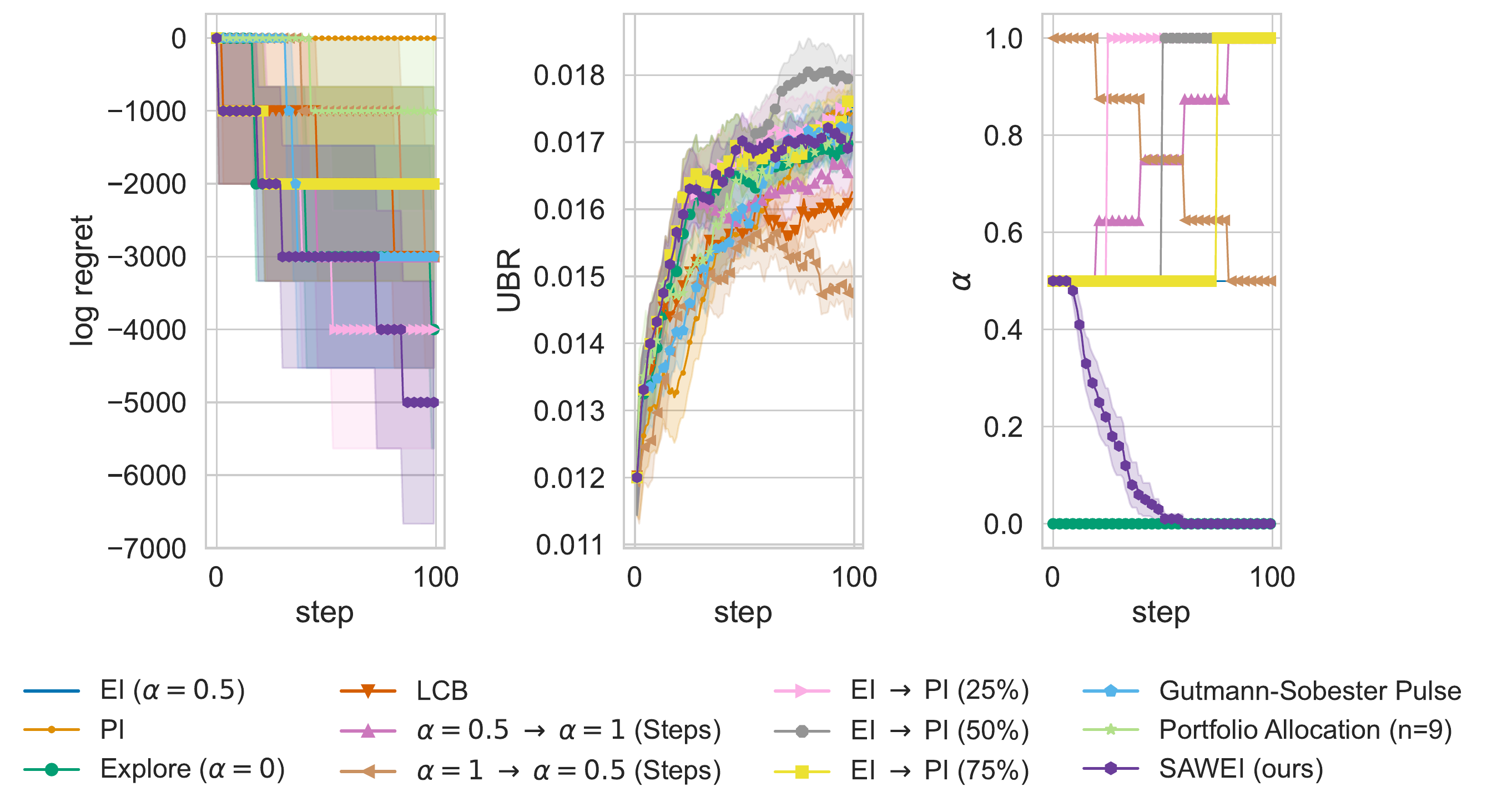}
    \caption{HPOBench ML: (model, task\_id) = (rf, 3917)}
    \label{fig:figures/HPOBench_det/alpha/rf_3917.pdf}
\end{figure}

\begin{figure}[h]
    \centering
    \includegraphics[width=0.85\linewidth]{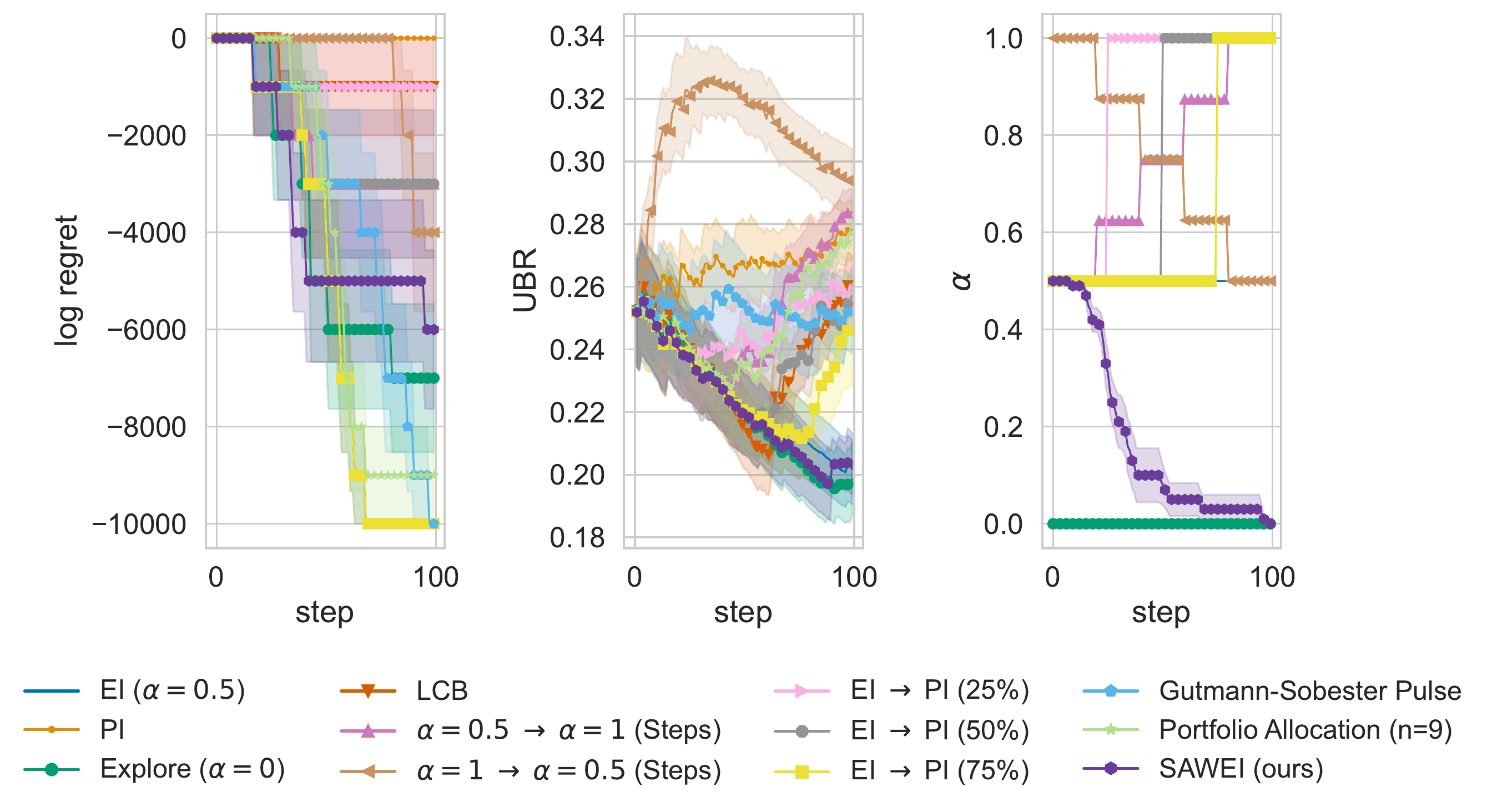}
    \caption{HPOBench ML: (model, task\_id) = (rf, 53)}
    \label{fig:figures/HPOBench_det/alpha/rf_53.pdf}
\end{figure}

\begin{figure}[h]
    \centering
    \includegraphics[width=0.85\linewidth]{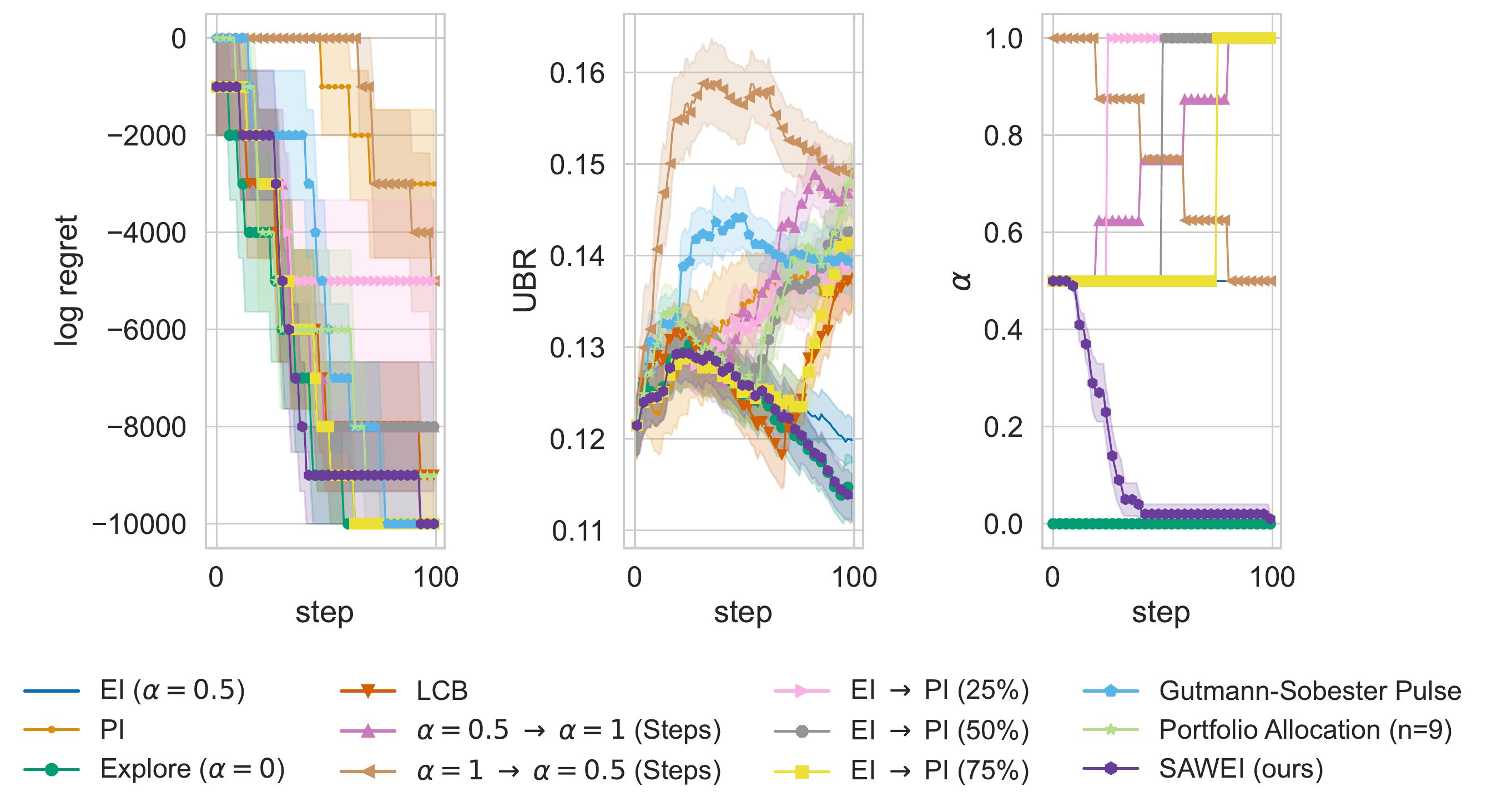}
    \caption{HPOBench ML: (model, task\_id) = (rf, 9952)}
    \label{fig:figures/HPOBench_det/alpha/rf_9952.pdf}
\end{figure}

\begin{figure}[h]
    \centering
    \includegraphics[width=0.85\linewidth]{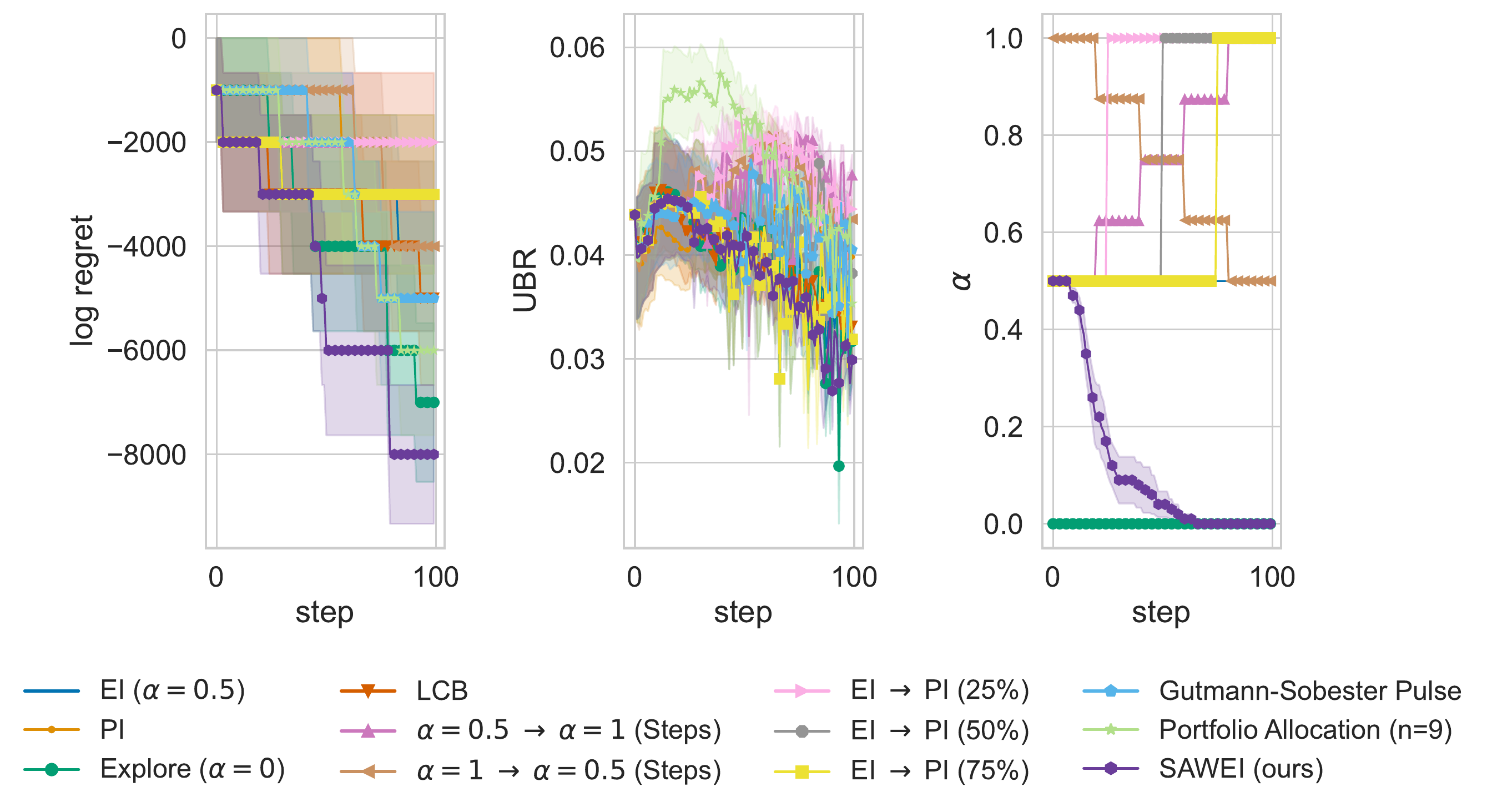}
    \caption{HPOBench ML: (model, task\_id) = (svm, 10101)}
    \label{fig:figures/HPOBench_det/alpha/svm_10101.pdf}
\end{figure}

\begin{figure}[h]
    \centering
    \includegraphics[width=0.85\linewidth]{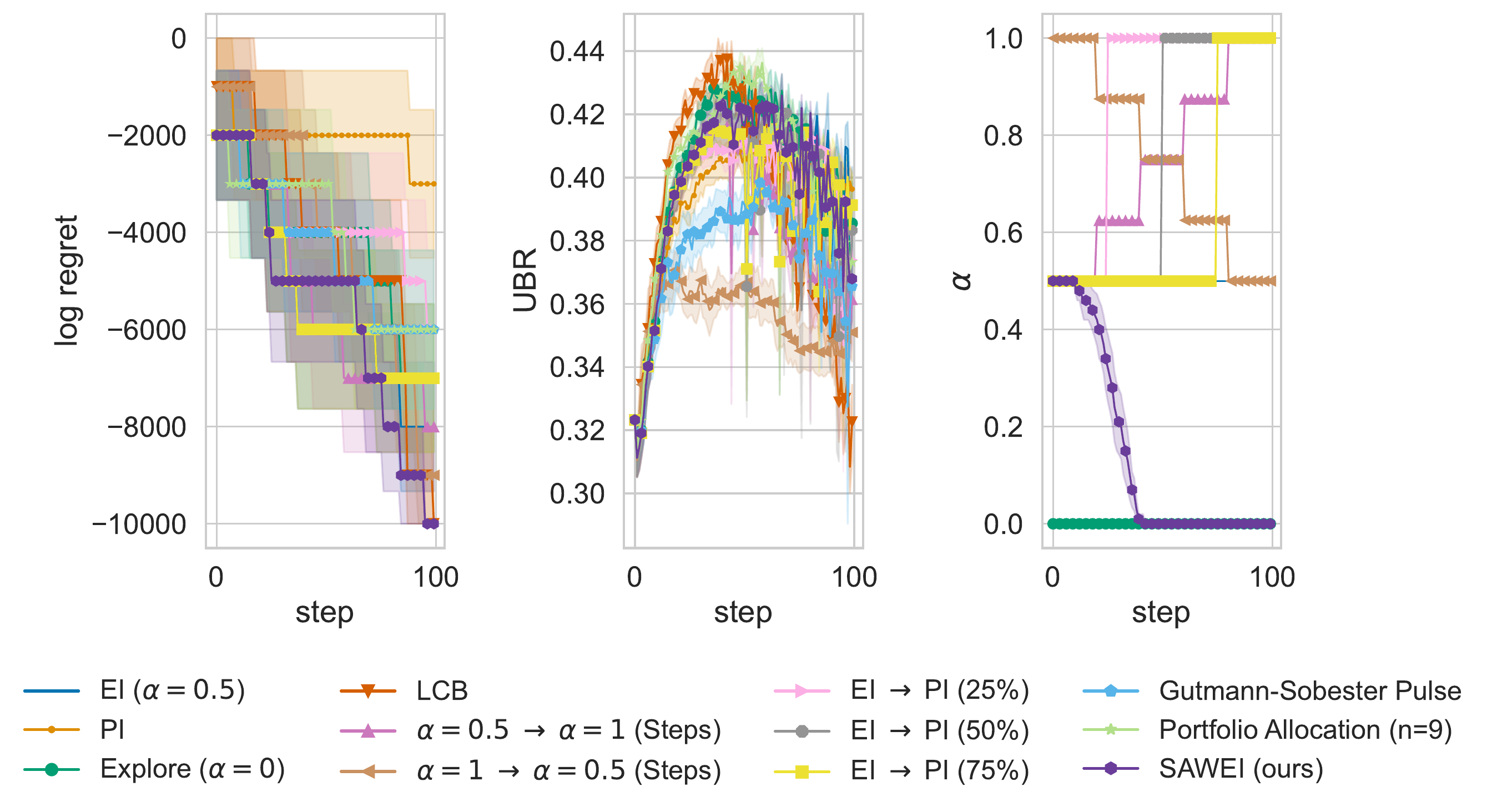}
    \caption{HPOBench ML: (model, task\_id) = (svm, 146818)}
    \label{fig:figures/HPOBench_det/alpha/svm_146818.pdf}
\end{figure}

\begin{figure}[h]
    \centering
    \includegraphics[width=0.85\linewidth]{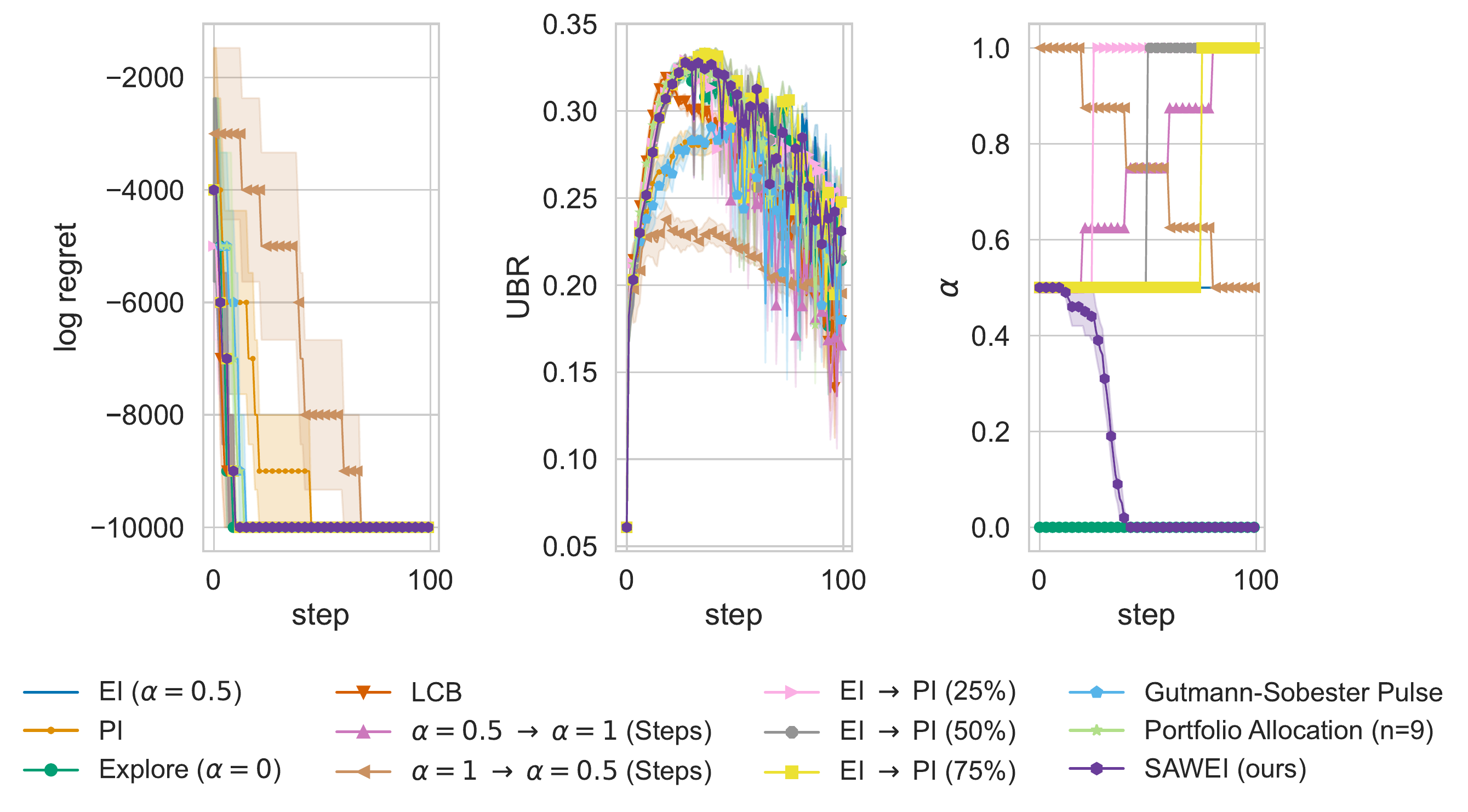}
    \caption{HPOBench ML: (model, task\_id) = (svm, 146821)}
    \label{fig:figures/HPOBench_det/alpha/svm_146821.pdf}
\end{figure}

\begin{figure}[h]
    \centering
    \includegraphics[width=0.85\linewidth]{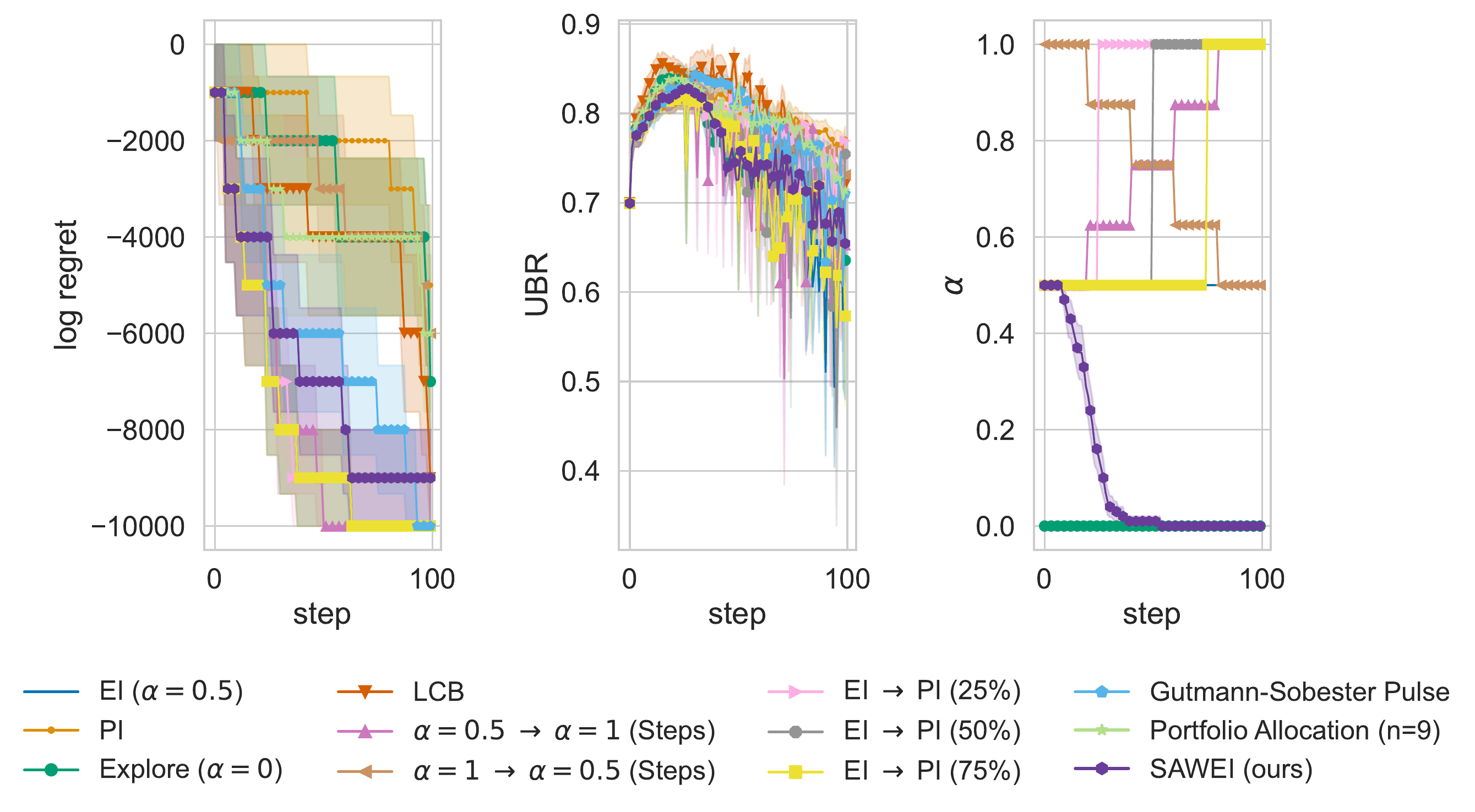}
    \caption{HPOBench ML: (model, task\_id) = (svm, 146822)}
    \label{fig:figures/HPOBench_det/alpha/svm_146822.pdf}
\end{figure}

\begin{figure}[h]
    \centering
    \includegraphics[width=0.85\linewidth]{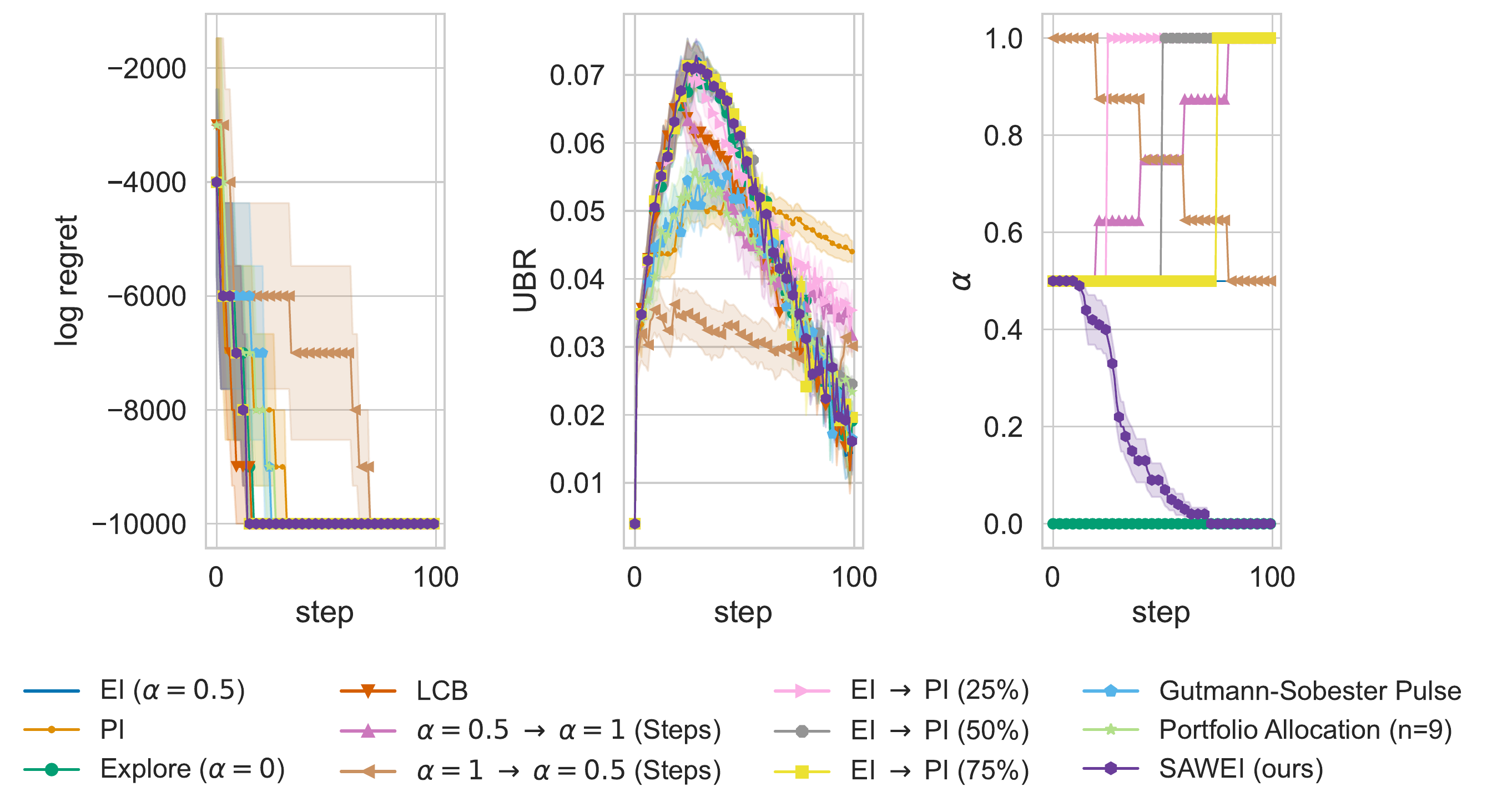}
    \caption{HPOBench ML: (model, task\_id) = (svm, 31)}
    \label{fig:figures/HPOBench_det/alpha/svm_31.pdf}
\end{figure}

\begin{figure}[h]
    \centering
    \includegraphics[width=0.85\linewidth]{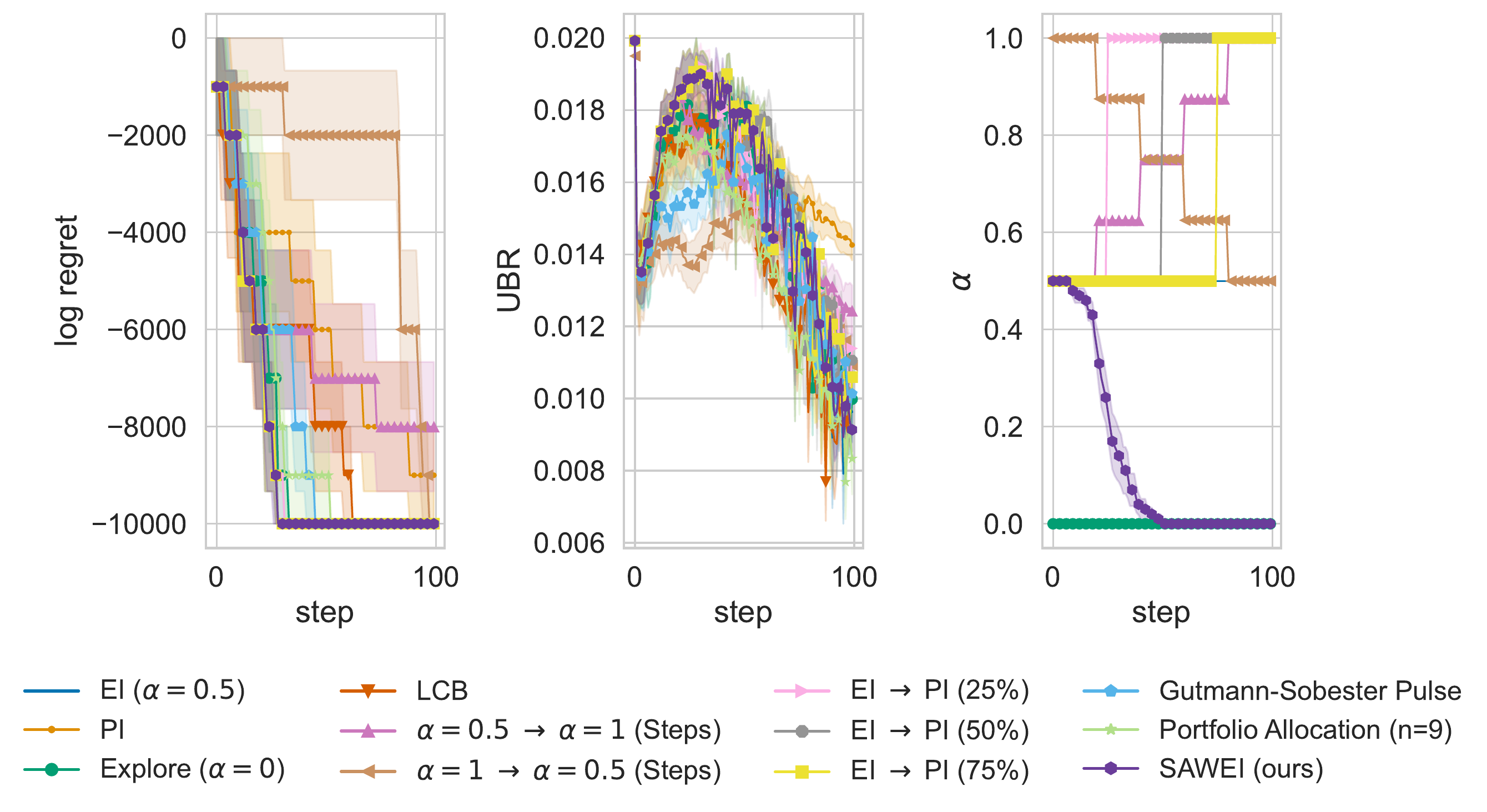}
    \caption{HPOBench ML: (model, task\_id) = (svm, 3917)}
    \label{fig:figures/HPOBench_det/alpha/svm_3917.pdf}
\end{figure}

\begin{figure}[h]
    \centering
    \includegraphics[width=0.85\linewidth]{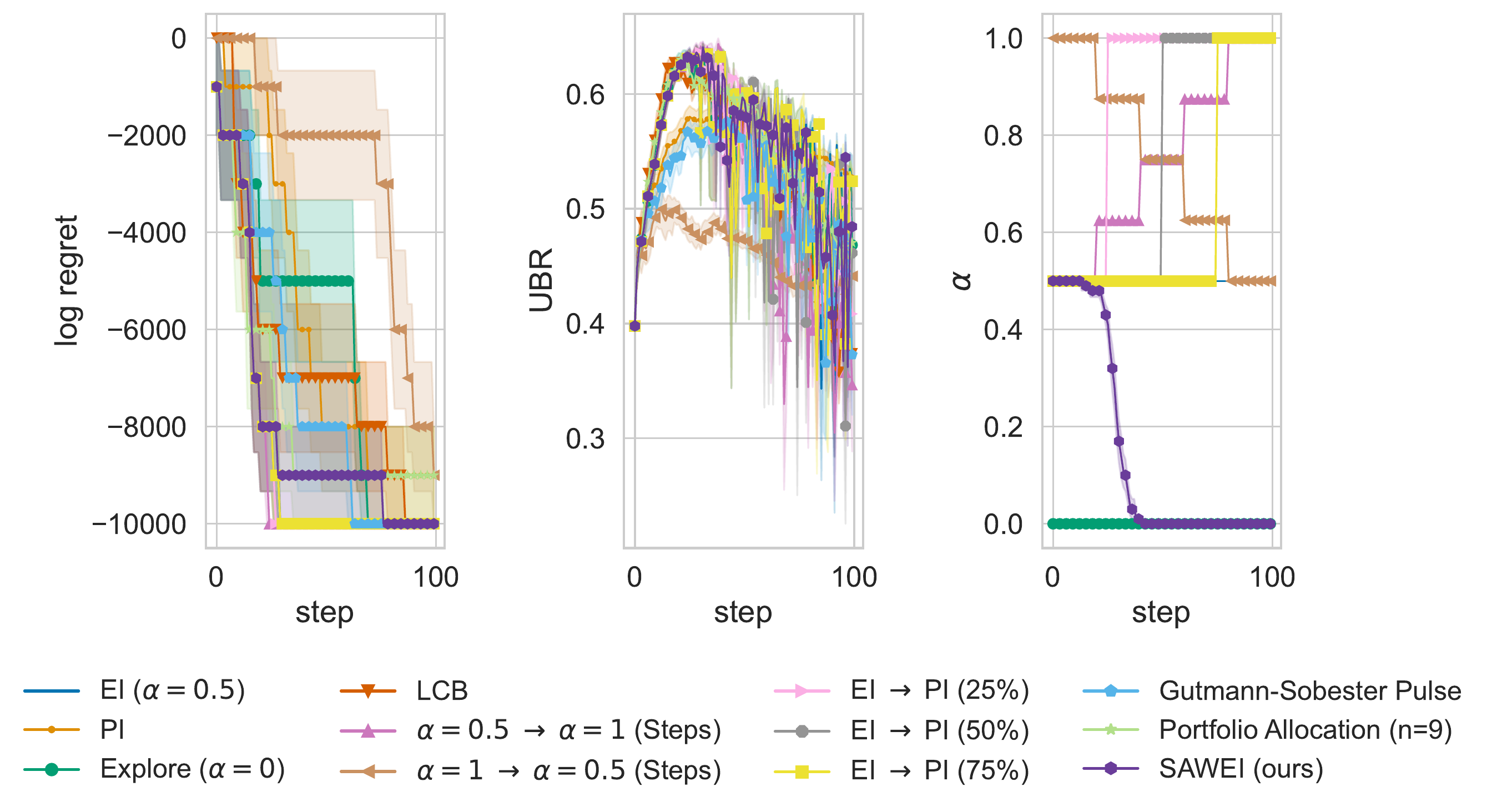}
    \caption{HPOBench ML: (model, task\_id) = (svm, 53)}
    \label{fig:figures/HPOBench_det/alpha/svm_53.pdf}
\end{figure}

\begin{figure}[h]
    \centering
    \includegraphics[width=0.85\linewidth]{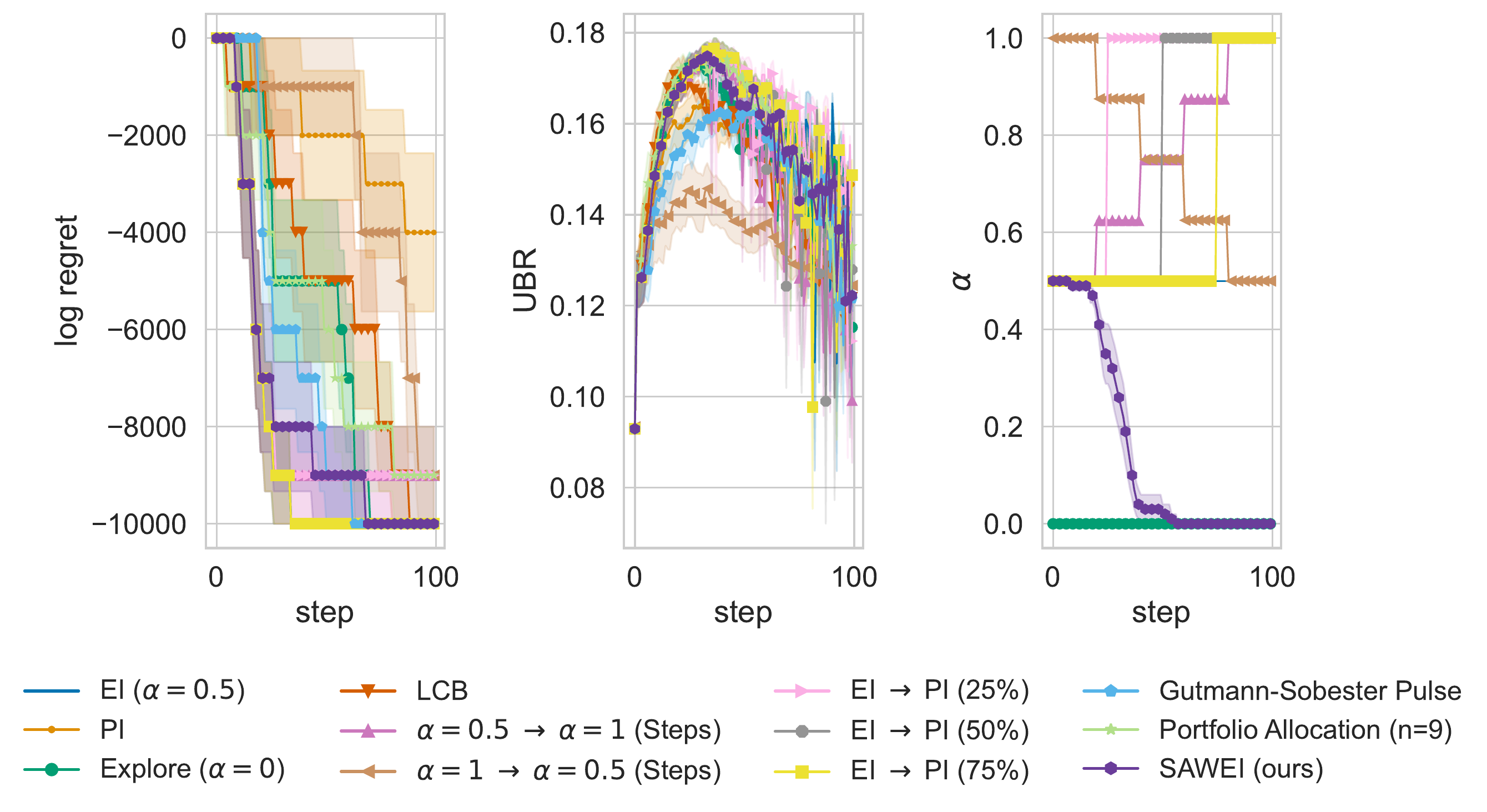}
    \caption{HPOBench ML: (model, task\_id) = (svm, 9952)}
    \label{fig:figures/HPOBench_det/alpha/svm_9952.pdf}
\end{figure}

\begin{figure}[h]
    \centering
    \includegraphics[width=0.85\linewidth]{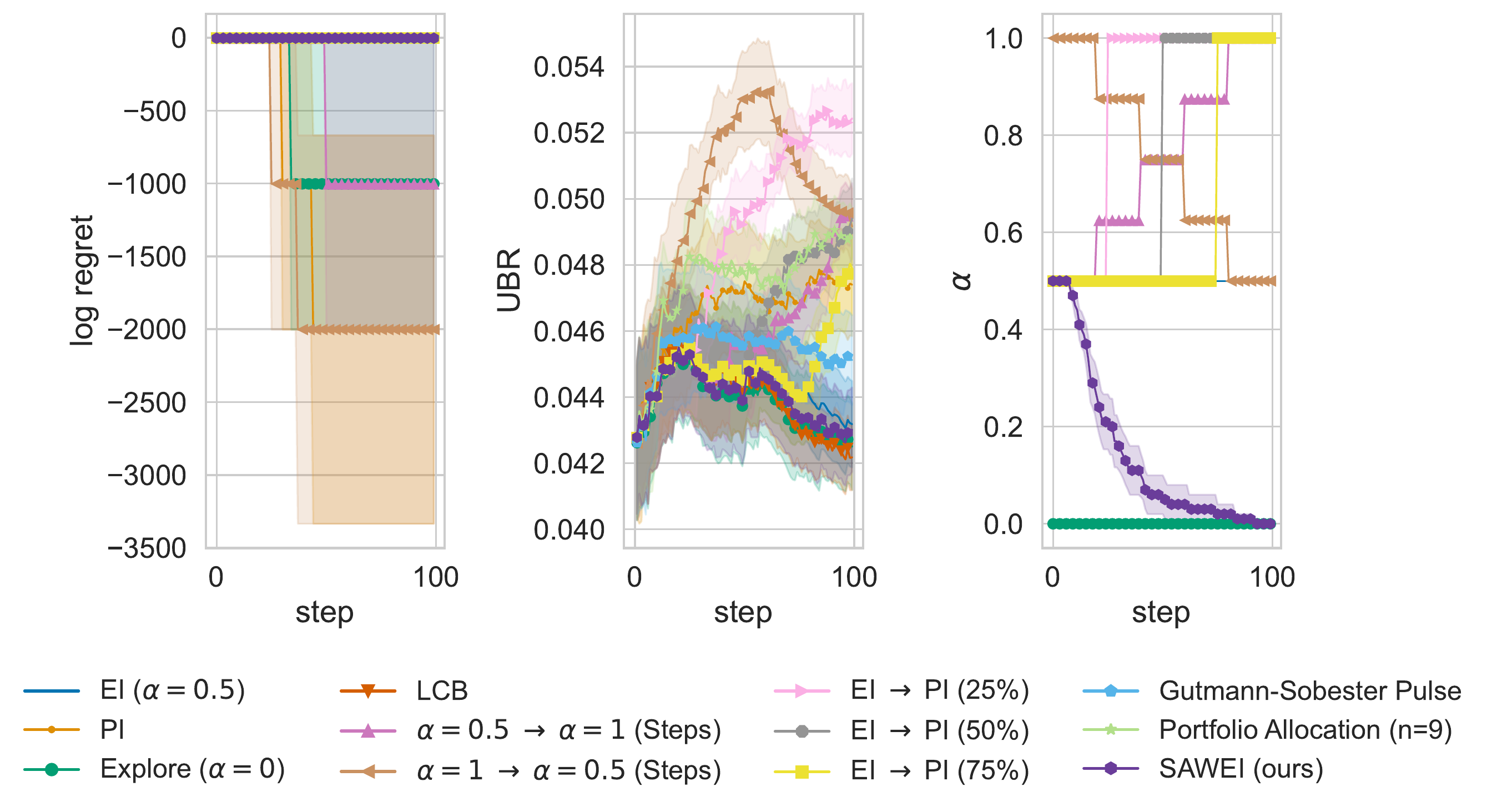}
    \caption{HPOBench ML: (model, task\_id) = (xgb, 10101)}
    \label{fig:figures/HPOBench_det/alpha/xgb_10101.pdf}
\end{figure}

\begin{figure}[h]
    \centering
    \includegraphics[width=0.85\linewidth]{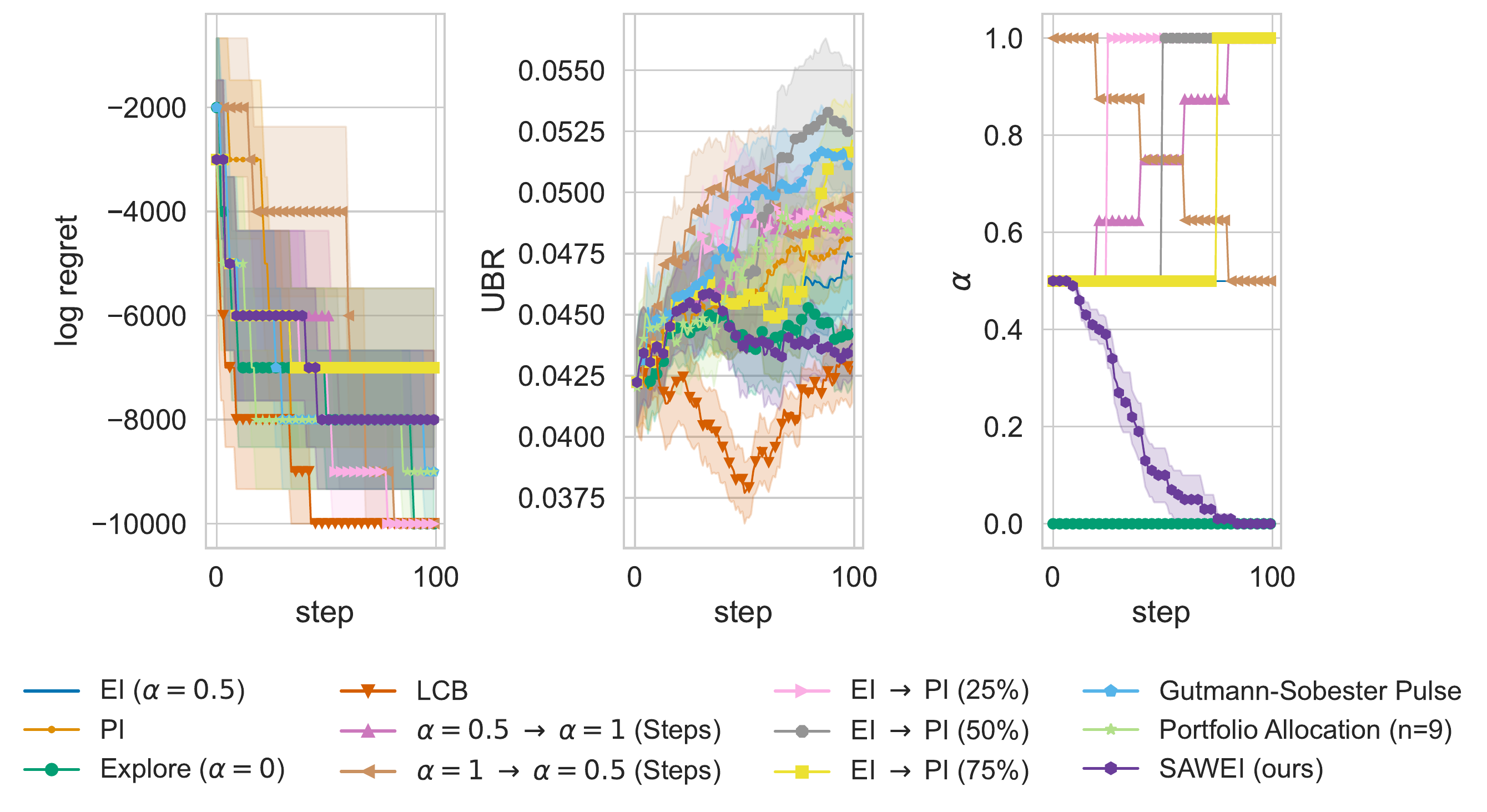}
    \caption{HPOBench ML: (model, task\_id) = (xgb, 146818)}
    \label{fig:figures/HPOBench_det/alpha/xgb_146818.pdf}
\end{figure}

\begin{figure}[h]
    \centering
    \includegraphics[width=0.85\linewidth]{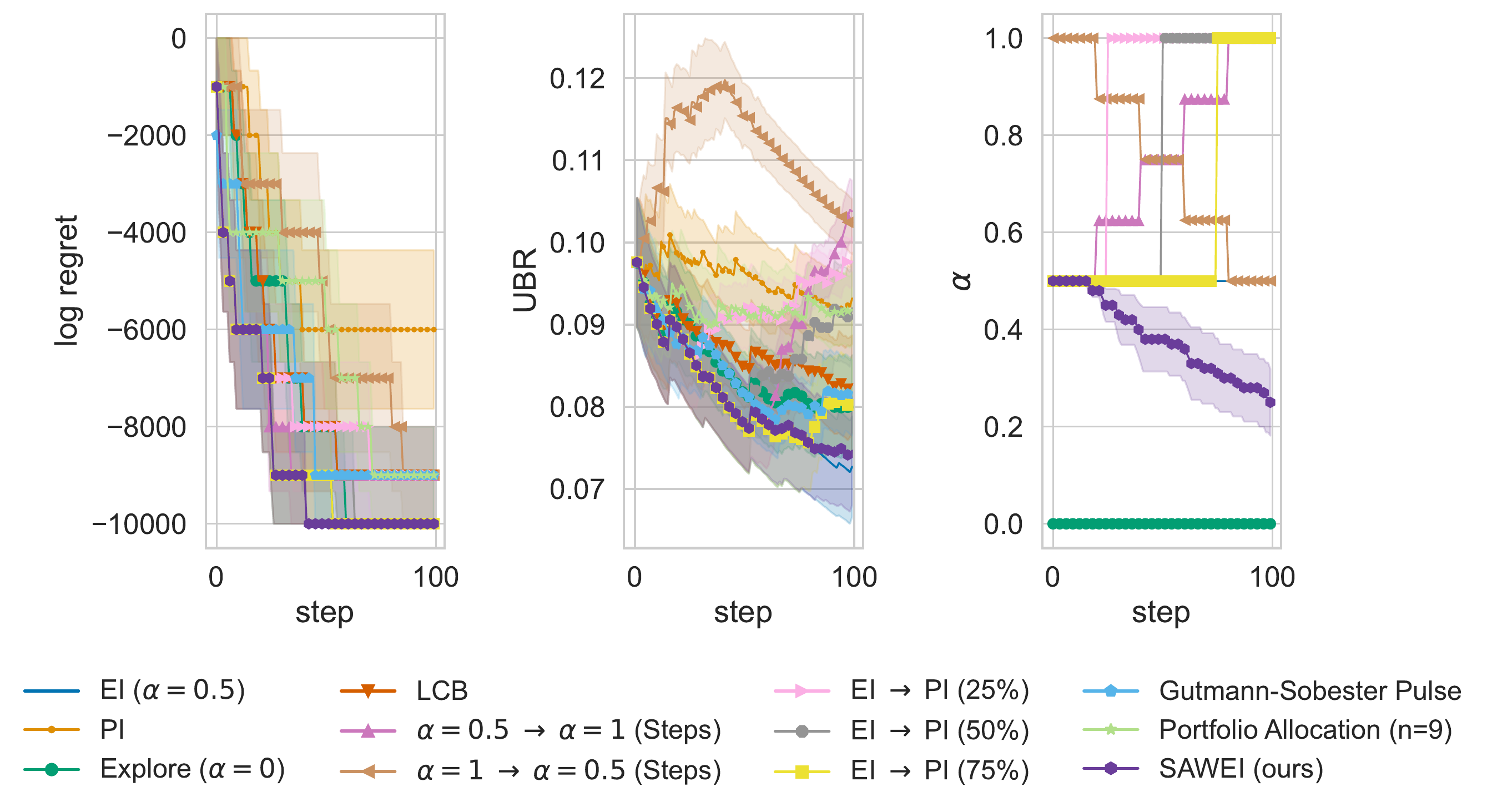}
    \caption{HPOBench ML: (model, task\_id) = (xgb, 146821)}
    \label{fig:figures/HPOBench_det/alpha/xgb_146821.pdf}
\end{figure}

\begin{figure}[h]
    \centering
    \includegraphics[width=0.85\linewidth]{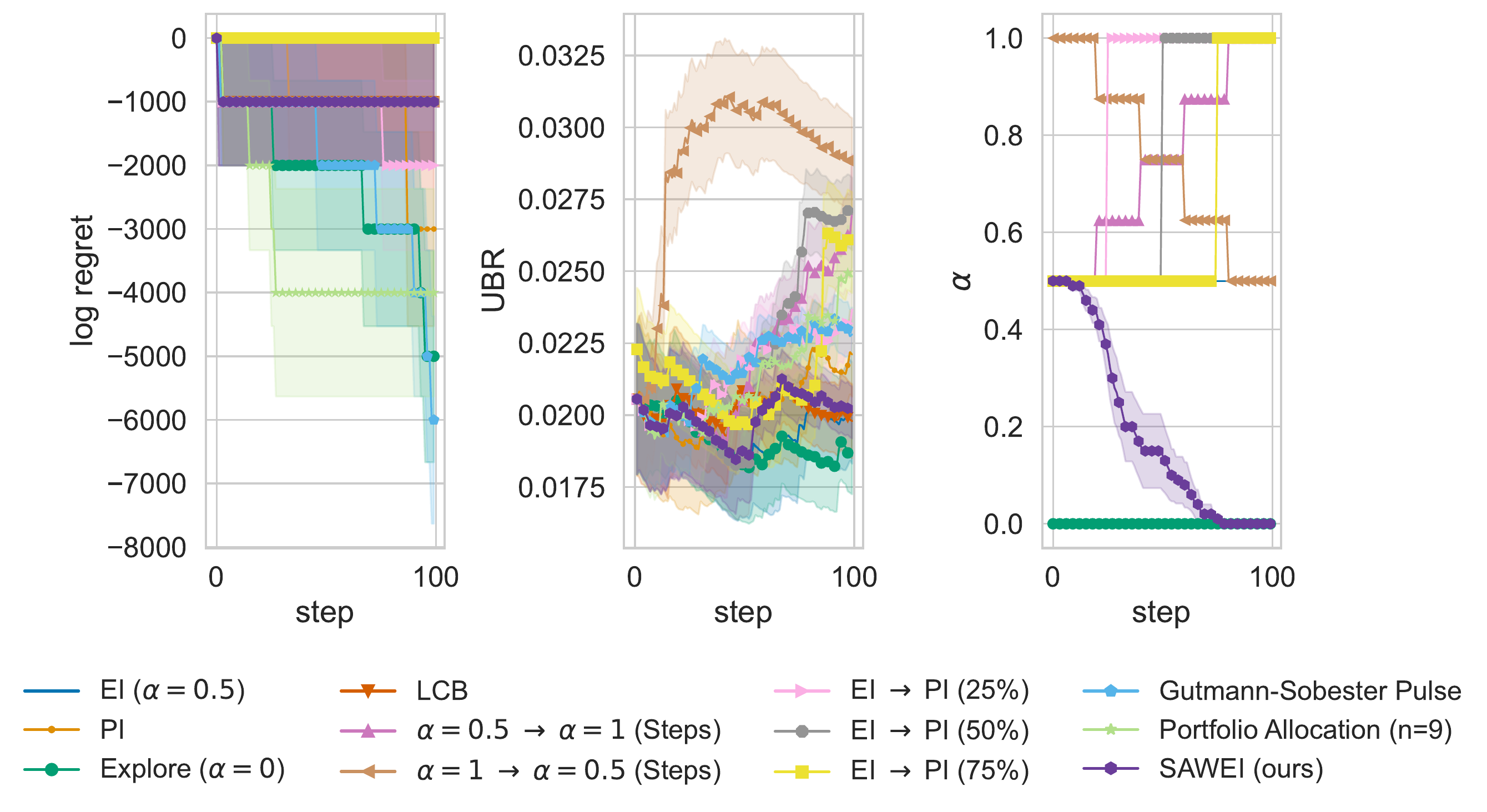}
    \caption{HPOBench ML: (model, task\_id) = (xgb, 146822)}
    \label{fig:figures/HPOBench_det/alpha/xgb_146822.pdf}
\end{figure}

\begin{figure}[h]
    \centering
    \includegraphics[width=0.85\linewidth]{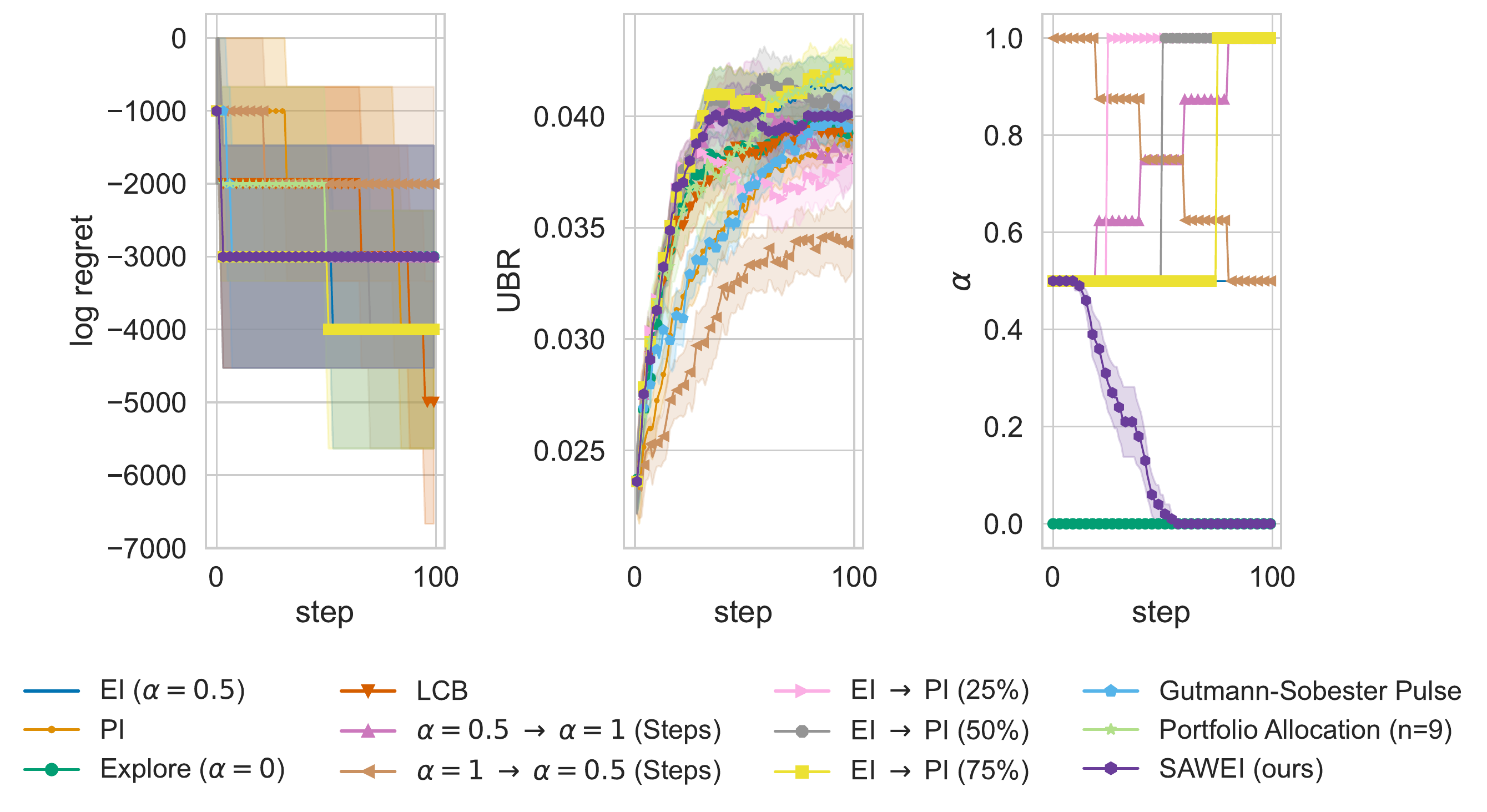}
    \caption{HPOBench ML: (model, task\_id) = (xgb, 31)}
    \label{fig:figures/HPOBench_det/alpha/xgb_31.pdf}
\end{figure}

\begin{figure}[h]
    \centering
    \includegraphics[width=0.85\linewidth]{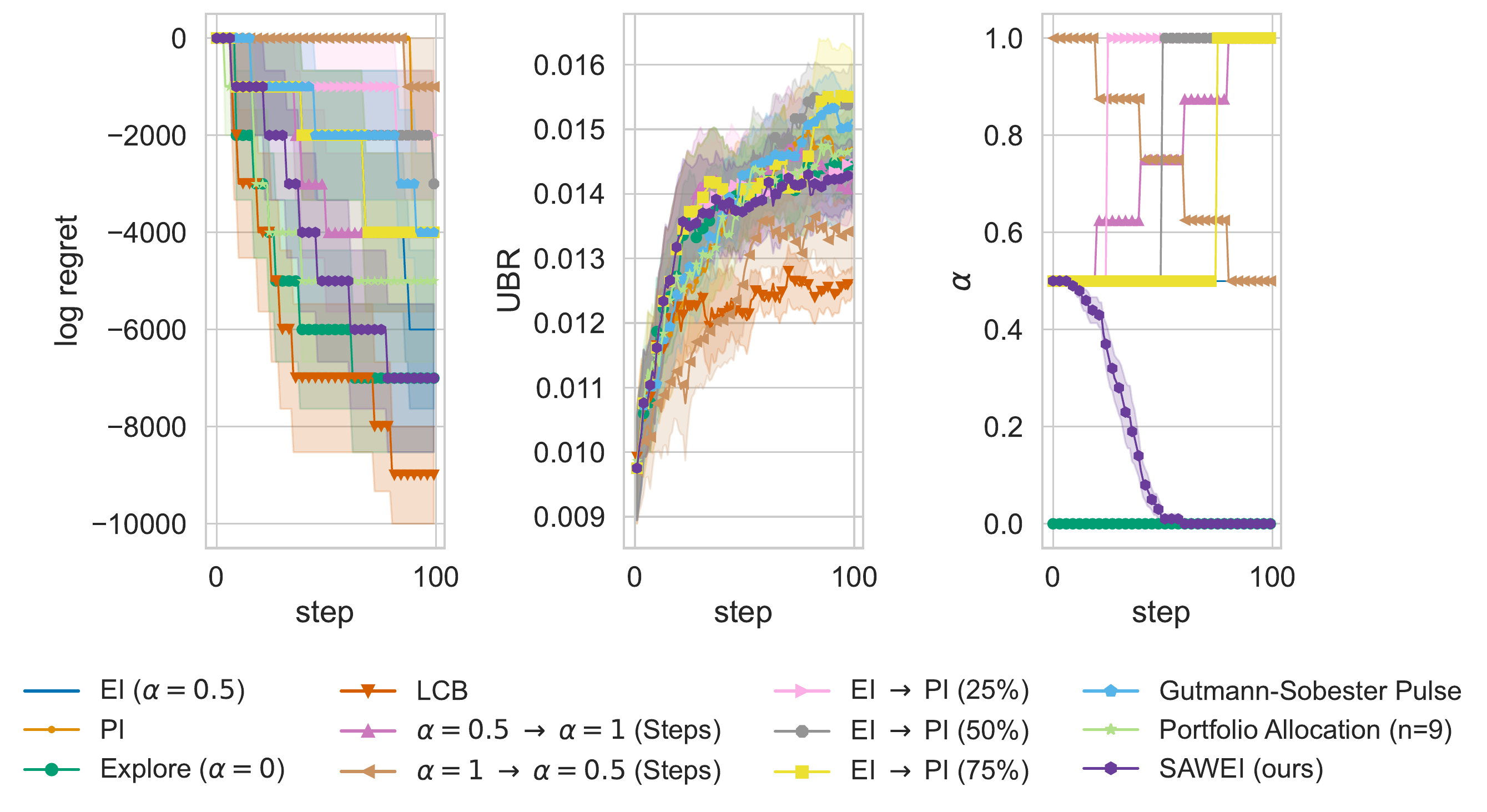}
    \caption{HPOBench ML: (model, task\_id) = (xgb, 3917)}
    \label{fig:figures/HPOBench_det/alpha/xgb_3917.pdf}
\end{figure}

\begin{figure}[h]
    \centering
    \includegraphics[width=0.85\linewidth]{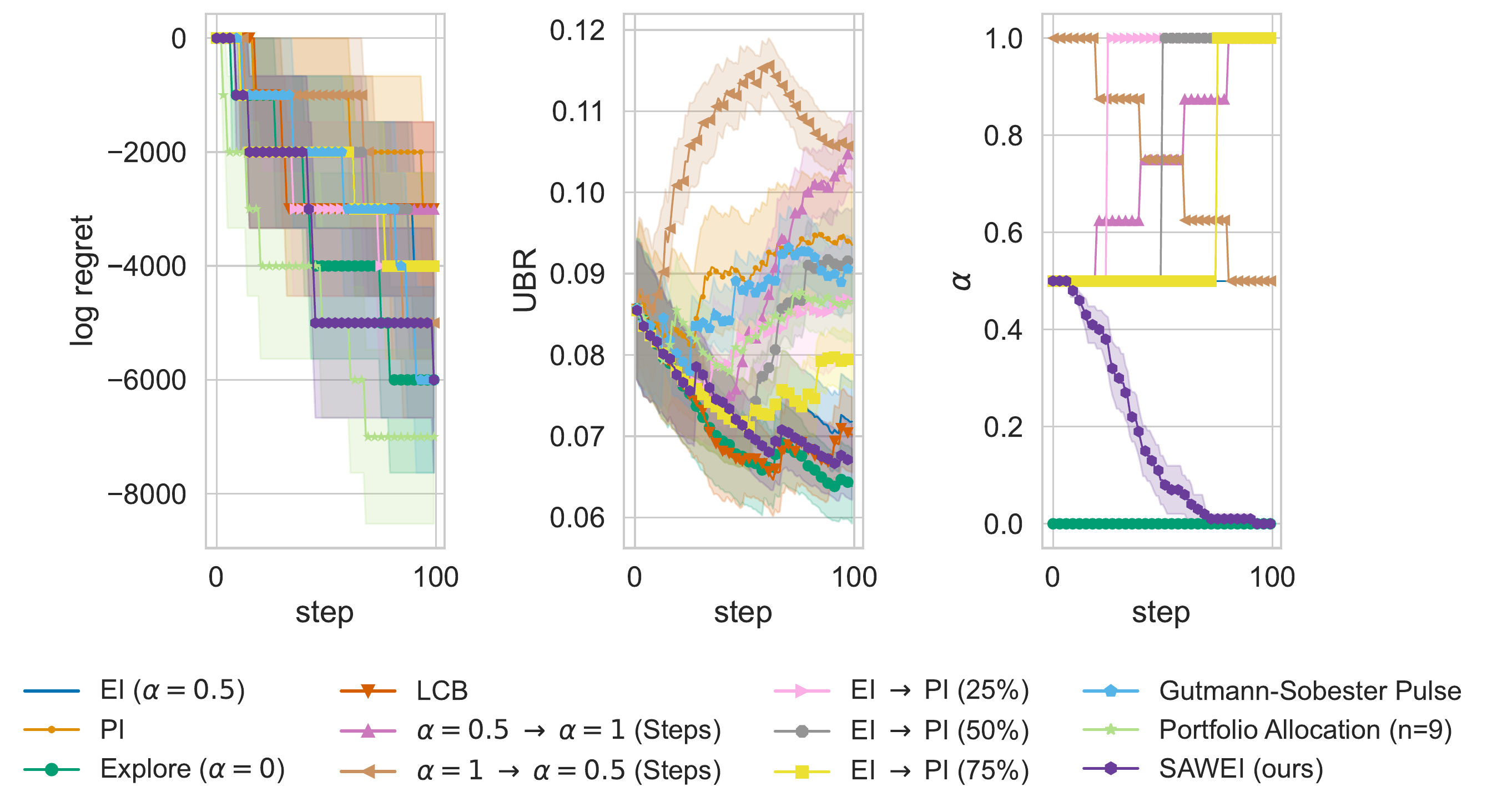}
    \caption{HPOBench ML: (model, task\_id) = (xgb, 53)}
    \label{fig:figures/HPOBench_det/alpha/xgb_53.pdf}
\end{figure}

\begin{figure}[h]
    \centering
    \includegraphics[width=0.85\linewidth]{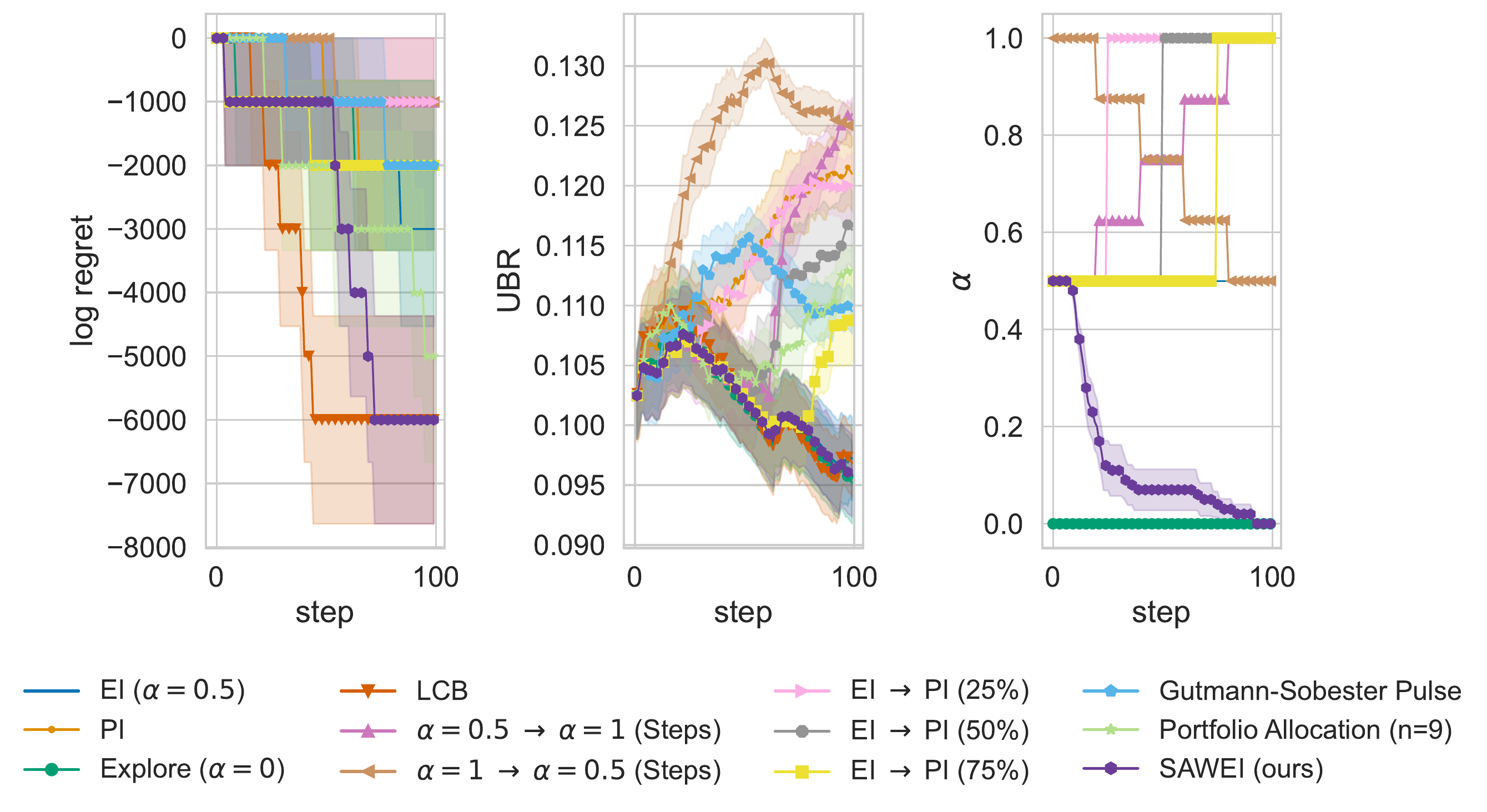}
    \caption{HPOBench ML: (model, task\_id) = (xgb, 9952)}
    \label{fig:figures/HPOBench_det/alpha/xgb_9952.pdf}
\end{figure}

\end{document}